%% file: main.tex
\documentclass{article} 
\usepackage[preprint]{colm2026_conference} 
\usepackage{lineno}  

\input{math_commands.tex}

\usepackage{hyperref}
\usepackage{url}
\usepackage{graphicx}
\graphicspath{{./}{../}{figures/}{../figures/}}
\usepackage{amsmath}
\usepackage{cleveref}
\usepackage{makecell}
\usepackage{caption}
\usepackage{wrapfig}
\usepackage{enumitem}
\usepackage{subcaption}
\usepackage[utf8]{inputenc} 
\usepackage[T1]{fontenc}    
\usepackage{booktabs}       
\usepackage{tabularx}       
\usepackage{amsfonts}       
\usepackage{nicefrac}       
\usepackage{microtype}      
\usepackage{xcolor}         
\usepackage[skins, breakable]{tcolorbox}
\usepackage{listings}

\definecolor{cornflowerblue}{rgb}{0.39, 0.58, 0.93}
\definecolor{mydarkblue}{rgb}{0,0.08,0.45}
\definecolor{kleinred}{HTML}{bc1919}
\hypersetup{
    hidelinks,
    colorlinks=true,
    linkcolor=kleinred,
    citecolor=mydarkblue
}

\tcbuselibrary{listings,skins,breakable}

\usepackage{tabularx}
\newcolumntype{L}{>{\raggedright\arraybackslash}X}
\usepackage{etoc}

\usepackage{graphicx} 
\usepackage{wrapfig}  
\usepackage{amsmath} 
\usepackage{natbib}

\newcommand{\ourdataset}{\texttt{OpenForesight}}
\newcommand{\ourmodel}{\texttt{OpenForecaster}}

\definecolor{sbgreendeep}{HTML}{55a868}
\usepackage{hyperref}
\hypersetup{
  colorlinks,
  urlcolor=kleinred
  }
\usepackage{etoc}
\usepackage[fixed]{fontawesome5}

\definecolor{myblue}{RGB}{0,70,180}

\newcommand{\contribution}[1]{\textcolor{myblue}{\textbf{#1}}}

\title{\vspace{-0.5cm}Scaling Open-Ended Reasoning To Predict the Future}

\author{
Nikhil Chandak\textsuperscript{1,3\thanks{Equal contribution \quad \textsuperscript{\ensuremath{\dagger}}Equal co-supervision}}\quad
Shashwat Goel\textsuperscript{1,2*}\\
\textbf{Ameya Prabhu\textsuperscript{3,4\ensuremath{\dagger}}}\quad
\textbf{Moritz Hardt\textsuperscript{1,3\ensuremath{\dagger}}}\quad
\textbf{Jonas Geiping\textsuperscript{1,2,3\ensuremath{\dagger}}}
\\\\
\textsuperscript{1}Max Planck Institute for Intelligent Systems \quad
\textsuperscript{2}ELLIS Institute T\"ubingen\ \\
\textsuperscript{3}T\"ubingen AI Center\quad
\textsuperscript{4}University of T\"ubingen
}

\ifcolmsubmission
\linenumbers
\fi
\begin{document}

\maketitle

\vspace{-0.5cm}
\begin{center}
    \begin{tabular}{c@{\hskip 19pt}c}
    \hspace*{1.4cm}\raisebox{-1pt}{\faGlobe} \href{https://openforecaster.github.io/}{\texttt{Blog}}
    \hspace*{1.4cm}\raisebox{-1pt}{\faGithub} \href{https://github.com/OpenForecaster/scaling-forecasting-training}{\fontsize{8.8pt}{0pt}\texttt{Code}}
    \hspace*{1.6cm}\raisebox{-1.5pt}{\faDatabase}\href{https://huggingface.co/collections/nikhilchandak/openforecaster}{\fontsize{8.8pt}{0pt} \texttt{Dataset and Models}} 
\end{tabular}
\end{center}

\lhead{} 
\renewcommand{\headrulewidth}{0pt} 
\vspace{0.25cm}
\begin{abstract}
  \input{sections/0_abstract}  
\end{abstract}
\vspace{-0.2cm}
\input{sections/1_intro}
\input{sections/2_setup}

\input{sections/3_data}
\input{sections/4_prediction}
\input{sections/7_results}
\input{sections/9_conclusion}
\input{sections/11_ack}

\bibliography{iclr2026_conference}
\bibliographystyle{colm2026_conference}

\input{sections/16_appendix}

\end{document}

%% file: math_commands.tex
\usepackage{amsmath,amsfonts,bm}
\usepackage{dsfont}

\def\eqref#1{equation~\ref{#1}}

\def\1{\bm{1}}

\DeclareMathAlphabet{\mathsfit}{\encodingdefault}{\sfdefault}{m}{sl}
\SetMathAlphabet{\mathsfit}{bold}{\encodingdefault}{\sfdefault}{bx}{n}



%% file: sections/0_abstract.tex
\looseness -1 High-stakes decision making involves reasoning under uncertainty about the future. In this work, we train language models to make predictions on open-ended forecasting questions. To scale up training data, we synthesize novel forecasting questions from global events reported in daily news, using a fully automated, careful curation recipe. We train the Qwen3 thinking models on our dataset, \ourdataset.  To prevent leakage of future information during training and evaluation, we use an offline news corpus, both for data generation and retrieval in our forecasting system. Guided by a small validation set, we show the benefits of retrieval, and an improved reward function for reinforcement learning (RL). Once we obtain our final forecasting system, we perform held-out testing between May to August 2025. Our specialized model, \ourmodel  8B, matches much larger proprietary models, with our training improving the accuracy, calibration, and consistency of predictions. We find calibration improvements from forecasting training generalize across popular benchmarks. We open-source all our models, code, and data to make research on language model forecasting broadly accessible. 

%% file: sections/1_intro.tex
\section{Introduction}

Every day, people navigate decisions under uncertainty, due to incomplete evidence or competing hypotheses. The highest-stakes choices are inherently forward-looking: governments set policy while anticipating macroeconomic and geopolitical shifts; investors allocate capital amid market and regulatory uncertainty; individuals choose careers as technologies evolve; and scientists pursue research directions in search of the next breakthrough. Decades of work~\citep{tetlock2014forecasting} on human forecasting shows that while prediction is hard and skill varies widely, it is possible to train humans to become better forecasters. In fact, some  ``superforecasters'' consistently outperform peers. While there is a ceiling to predictability in social systems~\citep{franklin1999real}, we do not yet know where that ceiling lies in the real world.

\looseness -1 If trained at scale for forecasting world events, Large Language Models (LLMs) may enjoy structural advantages over humans: they can ingest and synthesize vast, heterogeneous corpora across thousands of topics; and update predictions rapidly as new information arrives. Just like language models now show superhuman reasoning on some exam-style math and coding problems~\citep{icpc_openai_2025}, in the future, language model forecasters may be able to come up with possibilities that humans miss. So in this work, we study:

\begin{quote}
\emph{How can we train language models to better forecast open-ended questions?
}
\end{quote}

\textbf{Scaling training data for forecasting.} As forecasting world events is hard for humans, detailed and correct reasoning traces for forecasting are difficult to obtain. Fortunately, recent success in Reinforcement Learning (RL) for language models enables training with just the eventual outcome of the question~\citep{guo2025deepseek}. Further, the static knowledge cutoff of LLMs enables a unique opportunity: events that resolve after the cutoff are in the future for the model. Even then, sourcing questions at scale for training forecasting abilities has a few key challenges. First, waiting for events to resolve is too slow as a feedback loop for training. Second, prediction markets--the primary source for existing forecasting questions--mostly consist of binary yes or no questions. As there is a 50\% chance of success on these questions even with incorrect reasoning, they make for noisy rewards.

\looseness -1 Instead, we use global news, which covers a large number of salient events every day, to synthesize open-ended forecasting questions like ``Who will be confirmed as the new prime minister of Ukraine on 17 July 2025?''. Our recipe for creating training data is entirely automated and scalable, with one language model extracting events from news articles to generate questions, and a different model filtering and rewriting questions to avoid leaking future information. For this work, we use this recipe with 250,000 articles up till April 2025, to create \ourdataset, a dataset of $\sim$ 50,000 open-ended forecasting questions for training. To grade responses for open-ended questions, we use model-based \emph{answer matching}~\citep{chandak2025answer} consistent with frontier benchmarks like Humanity's Last Exam \citep{phan2025humanity}. \vspace{0.15cm}

\textbf{Ensuring we truly improve forecasting.} We take extensive measures to avoid the leakage of future information during training and evaluation. First, we do not use online search engines for sourcing news, as they have unreliable date cutoffs due to dynamic updates to documents and search ranking~\citep{paleka2025pitfalls}. Instead, we use the CommonCrawl News corpus, which provides static, monthly snapshots of global news. Second, we only train on events until April 2025, which is when the Qwen3 model weights we train were released. Finally, we do not observe performance on the test set until the very end. Our test set is composed of diverse news sources, different from the ones used in training and validation, to ensure we are not just learning distributional biases of the training data.\vspace{0.15cm}

\begin{figure}[t]
    \centering
    \vspace{-0.9cm}
    \includegraphics[width=0.9\linewidth]{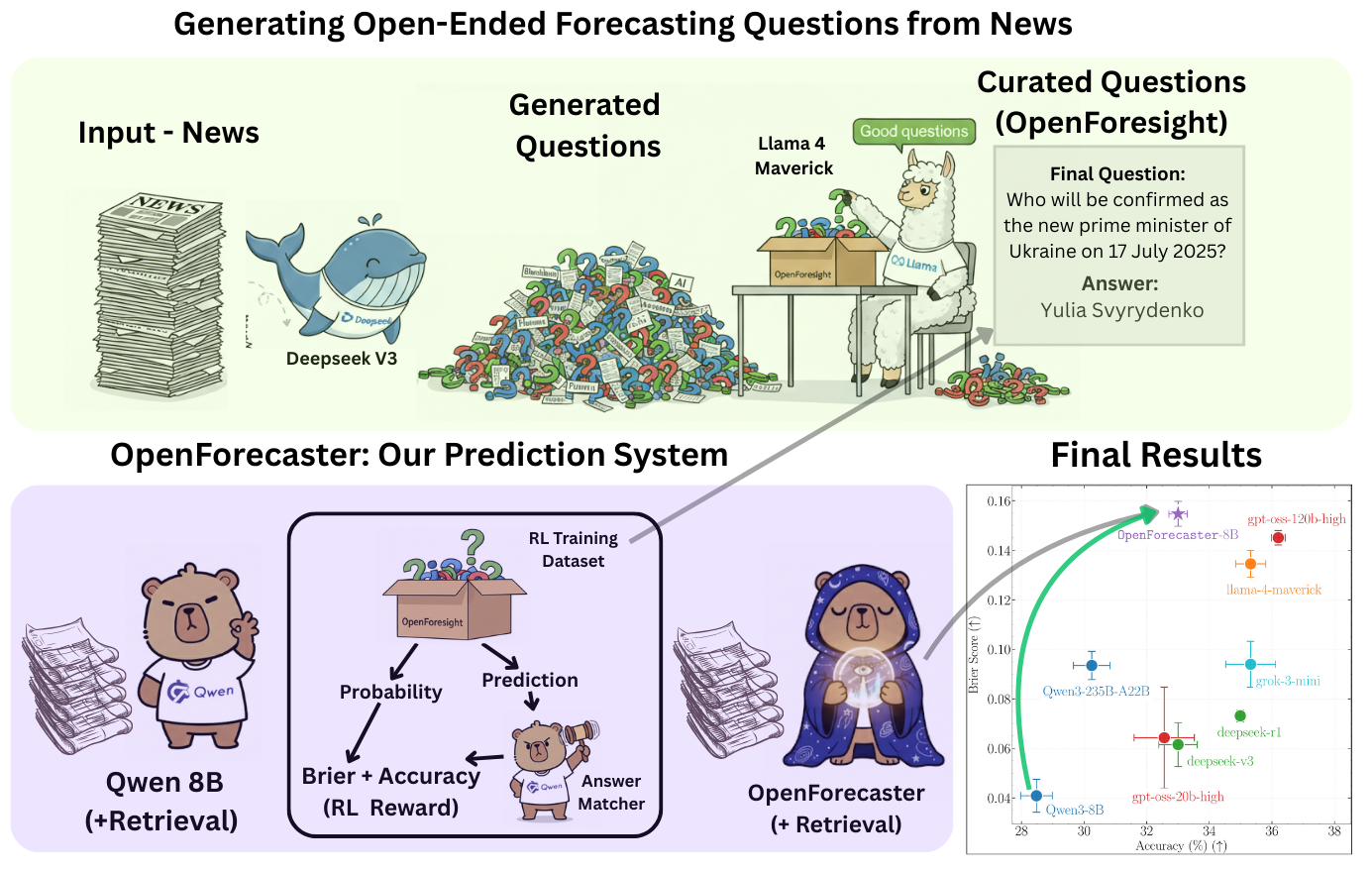}
    \vspace{-0.2cm}
    \caption{A summary of our methodology for training language model forecasters.}
    \label{fig:fig1}
    \vspace{-0.5cm}
\end{figure}

\textbf{Validating design choices for LLM Forecasting Systems.} We start from Qwen3~\citep{yang2025qwen3technicalreport} 4B and 8B models with thinking enabled. We perform all ablations on a small validation set. We use dense retrieval with the Qwen3-8B Embedding model to provide forecasters relevant chunks from our offline news corpus. Despite a cautious approach of only retrieving articles until \emph{one month} before the question resolution date to avoid leakage, the retrieved information leads to large improvements. Then, we train language models using RL with GRPO. For the reward function, we propose combining accuracy, and an adaptation of the brier score for open-ended responses~\citep{damani2025beyond}. Ablations show rewarding accuracy alone hurts calibration, while optimizing only the brier score hurts exploration on hard questions. Our final methodology is illustrated in Figure~\ref{fig:fig1}. \vspace{0.15cm}

\looseness -1 \textbf{Final results}. In \Cref{sec:results}, we report results on our held-out test set of open-ended forecasting questions from May to August 2025, and FutureX~\citep{zeng2025futurex}, an external forecasting benchmark. RL training on \ourdataset \ makes the predictions of our specialized 8B model competitive with much larger proprietary models in both accuracy and calibration. We also observe large improvements on consistency evaluations for long-term predictions~\citep{paleka2025consistency}. Finally, we find calibration from our forecasting training generalizes to multiple out of distribution benchmarks. \vspace{0.15cm}

By providing rigorous probabilistic predictions, open-ended forecasting systems could transform policy making, corporate planning, and financial risk management~\citep{tetlock2017expert}. To promote forecasting research, we open-source our dataset, code, and models.

%% file: sections/2_setup.tex
\section{Related Work}
\label{sec:related}

\looseness -1 \textbf{Forecasting World Events.} Much prior work in Machine Learning and Statistics has focused on forecasting numeric or time-series data~\citep{BoxJenkins1976} in diverse domains like weather~\citep{Richardson1922}, econometrics~\citep{Tinbergen1939} or finance~\citep{Cowles1933}. Instead, our work focuses on the prediction of discrete world events, with both questions and answers described in natural language, also called \textit{judgemental forecasting}~\citep{tetlock2016superforecasting}. In the rest of our paper, we refer to this as \textit{forecasting} for brevity. In prior work on evaluating language models for forecasting~\citep{zou2022forecasting, karger2024forecastbench}, questions are primarily sourced from prediction markets like Metaculus, Manifold, and Polymarket. Prediction markets provide a platform for online participants to register predictions on questions like ``Will Donald Trump win the US Presidential Election in 2024?", which mostly have binary, yes or no, outcomes and have rapidly grown in popularity over the last few years.  \vspace{0.15cm}

\looseness -1 \textbf{Evaluating LLMs for Forecasting.} New information (before the event resolves) benefits forecasting. Thus, LLM forecasting work~\citep{zou2022forecasting, halawi2024approaching} provides relevant retrieved articles to models~\citep{lewis2020retrieval} often obtained via web-search APIs. \citet{paleka2025pitfalls} discuss pitfalls of LLM forecasting evaluations, including leakage of outcomes from online search in backtests, and distributional biases of prediction market questions. To avoid these issues, we use static, monthly snapshots of global news for retrieval and creating questions. \citet{jin-etal-2021-forecastqa} ask humans to create forecasting questions, while \citet{dai2024llms} try to automate this process with LLMs. However, their questions pre-define a few outcomes to choose from. ~\citet{guan2024openep, wang2025openforecast} evaluate models on open-ended forecasts, but we go a step further by showing how to train models for this task. \vspace{0.15cm} 

\textbf{Reinforcement Learning for LLMs.} \citet{shao2024deepseekmath} proposed \textit{Group Relative Policy Optimization} (GRPO), an RL algorithm that only uses outcome rewards. This approach has been highly successful in training LLMs to \textit{reason} about well-specified coding~\citep{jain2024livecodebench} and exam-style questions across domains~\citep{phan2025humanity}. Instead, forecasting requires LLMs to reason about uncertainty. \citet{halawi2024approaching} proposed training language models for forecasting by Supervised Finetuning (SFT) on chain of thought traces that lead to brier scores better than the prediction market aggregate. In the same setting of binary forecasting questions, \citet{turtel2025llms} optimize brier scores using GRPO, while ~\citet{damani2025beyond} extend it to short answer questions in other domains. We depart from these works in showing how to synthesise large-scale forecasting training data from daily news, to train models at open-ended reasoning about the future.

\section{Open-Ended Forecasting}
\label{sec:open-ended}

\textbf{Motivation.} The forecasting task we study is \textit{open-ended} in two key ways: 
\begin{enumerate}
    \item It allows expressing arbitrary natural language forecasting questions
    \item It may not have a structured outcome set, unlike numeric or categorical predictions. This differentiates it from both time-series forecasting, and prediction markets.
\end{enumerate}
For example, prediction markets are dominated by binary (yes/no) or multiple choice questions. While this design is easy to score, the most foresight often lies in predicting the unexpected, or when a large number of possibilities could occur. The most important questions to forecast---such as scientific breakthroughs, geopolitical shocks, or technological disruptions---often emerge as \emph{unknown unknowns}: possibilities not anticipated, and hard to enumerate. Thus, in this work, we focus on training models to make open-ended predictions like "Which company will the US Government buy a >5\% stake in by September 2025?". Such questions require exploration and imagination, rewarding novel hypotheses that turn out to be correct, rather than just distributing probabilities over a known set of outcomes.\vspace{0.15cm} 

\textbf{Background.} LLM weights are frozen after training, especially for open-weight models. Any event that happens after the last date in the training corpus is in the future for the LLM. This provides a time window to collect questions for training models to reason about future events. Similarly, their evaluation involves testing on questions resolving after the cutoff date of the training data, called \textit{backtesting}~\citep{Tashman2000}. While prior work has relied on prediction market questions as training data, this has three key problems:
\begin{enumerate}
    \item The questions are created by humans, which makes them low in number~\citep{paleka2025pitfalls}. This becomes a bottleneck for scaling training data, which has been an essential component in the success of LLMs~\citep{kaplan2020scalinglawsneurallanguage, lu2025internet}.
    \item Most questions have binary outcomes, which creates a 50\% baseline success rate. This leads to noisy rewards in outcome-based RL, which means even incorrect reasoning has a high chance of being reinforced.
    \item Each platform has a skewed distribution of events. All overrepresent US political news, along with their specific focus such as crypto-currency price movements in Polymarket, technology in Metaculus, personal life of users in Manifold, and sports events on Kalshi~\citep{paleka2025pitfalls}. 
\end{enumerate}   
These limitations motivate us to explore alternate ways to create forecasting questions.\vspace{0.15cm} 

\looseness -1 \textbf{Setup.} Let $\mathcal{X}$ be the set of open-ended forecasting questions; and $\mathcal{Y}$ the set of short textual answers. We provide a language model $\pi_\theta$ a question $x \in \mathcal{X}$, for which we already know the ground-truth outcome $y^\star$ as it has resolved in the real-world. We ask the model to report its best guess prediction $y$, and the probability $q$ of it being the true outcome. \vspace{0.15cm} 

\textbf{Measuring Accuracy.} We measure accuracy by checking if the model's attempted answer $y$ matches with the ground truth outcome $y^\star$, using another language model to test for semantic equivalence (for example ``Geoffrey Hinton'' = ``Geoffrey Everest Hinton'') consistent with recent frontier benchmarks~\citep{wei2024measuringshortformfactualitylarge, phan2025humanity}. For evaluations, we use Llama-4-Scout~\citep{meta_llama4_2025}, as in a recent study~\citep{chandak2025answer}, it at matching answers, it has inter-human levels of alignment with human judgments. During training, we use Qwen3-4B in non-thinking mode, as it achieves high alignment levels for its size in the same study. We find the two models agree on $\sim 97\%$ grading responses, and manual validation ensures they are accurate in $\ge 95\%$ cases.\vspace{0.15cm} 

\textbf{Measuring Calibration.} We adapt the multi-class Brier scoring rule \citep{mucsanyi2023trustworthy} for free-form responses as follows (details in \Cref{sec:freeform_brier}):
$$
S^\prime(q, y, y^*) = \begin{cases}
    1 -(q - 1)^2, & \text{if } y \equiv y^*\\
     - q^2, & \text{if } y \neq y^*
    \end{cases} \\
    \label{eq:rlcr}
$$
\looseness -1 \textbf{Interpretation.} Predicting an event with a probability $q=0$ returns a baseline score of $0$ regardless of the prediction $y$. Correct predictions receive positive scores while incorrect predictions negative. For brevity, we call $S^\prime(q, y, y^*)$ \emph{Brier score} throughout this paper. Our Brier score is equivalent to the reward metric used by~\citet{damani2025beyond}. They show this is a proper scoring rule, incentivizing both high accuracy and truthful reporting of probability on the answer that seems most likely. For completeness, we discuss this further in \Cref{sec:freeform_brier}.\vspace{0.15cm} 

\textbf{Training Algorithm: GRPO~\citep{shao2024deepseekmath}.} 
We train LLMs using outcome-based reinforcement learning on our dataset. For each prompt $x$, we draw $K$ completions $\{(y_i,p_i)\}_{i=1}^K \sim \pi_\theta(\cdot\mid x)$ and compute rewards $r_i=R(y_i,p_i;y^\star)$. However, following prior work \citep{damani2025beyond, turtel2025outcome}, we do not divide by \emph{remove} the group standard deviation when computing advantages, as this stabilizes updates in settings like ours where reward variance can be small. \vspace{0.15cm} 

\textbf{Initial Policy: Qwen3 Thinking~\citep{yang2025qwen3technicalreport}.} We start with the 8B thinking model. For Qwen3 models, no official knowledge–cutoff date is reported. When queried directly, the models return inconsistent cutoff dates (most often \emph{October 2023} or \emph{June 2024}). Usually, they treat questions about 2024 as being in the future. Since the model weights were released and frozen in April 2025, we train up to this date, and use the period between May to August 2025 for testing. In the Appendix, we show large improvements from our training on even the Llama and Gemma models in~\Cref{sec:extra_forecasting_results}.

%% file: sections/3_data.tex
\section{Generating Open-Ended Forecasting Questions from News}
\label{sec:questions_data}

\looseness -1 We now discuss our methodology to convert daily news articles into forecasting questions using language models. Any fixed forecasting dataset loses value as newer base models with training cutoffs after the dataset was created are adopted. Thus, we first describe the general methodology which can be used in the future, and then describe the specific instantiations we used to create our training data \ourdataset which has questions until April 2025. We conclude by demonstrating forecasting improvements due to our data filtering steps.

\subsection{Methodology for Generating Forecasting Questions}
\label{sec:question_creation_pipeline}

\begin{figure}[t]
\vspace{-0.6cm}
    \centering
    \includegraphics[width=0.885\textwidth]{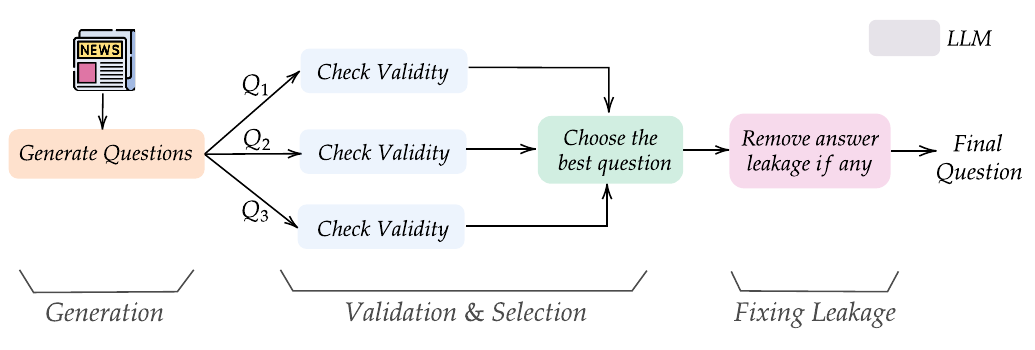}
    \vspace{-0.3cm} 
    \caption{\textbf{Our question generation methodology.} We use DeepSeek-v3 to generate multiple forecasting questions per news article. Then, we use Llama-4-Maverick to check if questions follow all guidelines, choose the best question, and remove any hints revealing the answer.}
    \label{fig:qgen_pipeline}
    \vspace{-0.4cm} 
\end{figure}

We generate short-answer, open-ended forecasting questions from individual news articles as illustrated in \Cref{fig:qgen_pipeline}. We describe each step in detail below:

\textbf{Sourcing Event Information.}
News outlets establish global infrastructure for reporting salient events as they occur. Unfortunately, \citet{paleka2025pitfalls} show that sourcing news via online search engines is unreliable. While search engines provide date cutoffs, future information can leak through updates to articles after the publish date, and even search engine ranking. This compromises the reliablity of backtests, and leaks future information in training, which can hurt Deep Learning models that easily overfit to spurious correlations. Fortunately, the CommonCrawl News (CCNews) Corpus~\citep{nagel2016-ccnews} provides static monthly snapshots of global news with accurate dates. This makes it free and easy to obtain news articles for creating forecasting questions. \vspace{0.15cm}

\textbf{Generating samples from documents.} Based on each news article, we ask a \textit{sample creator} model to generate up to three diverse forecasting samples. Each sample consists of: 
\begin{enumerate}
    \item \textbf{Question:} Asks about the prediction of an event. 
    \item \textbf{Background:} Provides brief context, and defines uncommon terms.
    \item \textbf{Resolution criteria:} Fixes a source of truth, proposes a resolution date for the question, and the expected answer format.
    \item \textbf{Answer:} Drawn verbatim from the article, unique, short (usually 1–3 words), and non-numeric (usually a name or location).
    \item \textbf{Source article link:} Obtained from article metadata for future reference.
\end{enumerate} \vspace{0.15cm}

\begin{tcolorbox}[title=\textbf{Sample Generated Forecasting Question}, colframe=black!70, colback=black!2, boxrule=0.5pt, breakable]
\textbf{Question.} Who will be confirmed as the new prime minister of Ukraine by 17 July 2025?

\medskip
\textbf{Background.}
Ukraine's parliament is scheduled to vote to appoint a new prime minister.

\medskip
\textbf{Resolution Criteria.}
\begin{itemize}[leftmargin=1.1em, itemsep=0.2em]
  \item \textbf{Source of Truth:} Official announcement from the Verkhovna Rada (Ukraine's parliament) confirming the appointment, via parliamentary records or government press release.
  \item \textbf{Resolution Date:} 17 July 2025, the date on which the parliamentary vote occurs and results are published.
  \item \textbf{Accepted Answer Format:} Full name of the individual exactly as given in the parliamentary announcement.
\end{itemize}

\medskip
\textbf{Answer Type.} String (Name)

\medskip
\textbf{Ground-Truth Answer.} Yulia Svyrydenko

\medskip
\textbf{Source.} The Guardian (live blog): \href{https://www.theguardian.com/world/live/2025/jul/17/ukraine-russia-war-patriot-nato-latest-europe-news-live-updates}{Ukraine live updates --- 17 July 2025}
\end{tcolorbox} \vspace{0.15cm}

\looseness -1 A challenging issue we face is that sometimes news articles talk about past events, or report an event late. This is why we ask the sample creator to propose a resolution date, and set the final resolution date as \texttt{min}\big(\texttt{model\_generated\_date}, \texttt{article\_publish\_date}\big). We perform additional steps, including manual review, to address this issue for evaluation questions, as described later in \Cref{sec:results}. For training data, we do not add more complex steps to fix resolution dates due to cost constraints. \vspace{0.15cm}

\looseness -1 \textbf{Filtering samples.} For each question, we use another LLM, \textit{the sample selector}, to verify:
\begin{enumerate}
    \item The question-answer pair is fully based on information in the source article.
    \item The question is forward-looking, for e.g. it is in future tense
    \item The answer is definite, unambiguous, and resolvable by the publication date.
\end{enumerate}
We mark a question as valid only if it passes these checks. If multiple questions from a single article remain, we ask the sample selector to pick the best one, favoring questions with clear, unique answers and high relevance. This is to ensure data diversity, and enhance quality. \vspace{0.15cm}

\looseness -1 \textbf{Editing to fix leakage.} At this stage, we find that even the filtered samples sometimes leak information about the answer. This can create shortcuts during training. To fix this, we do a final editing stage where we ask the sample selector to scan the title, background, and resolution criteria to check if they reveal the answer. When it finds leakage, we ask it to rewrite only the offending spans, replacing specifics with generic placeholders. Finally, we re-scan using exact string matching any remaining mentions of the answer, and discard those samples.

Overall, this pipeline can continually ingest news articles and generate open-ended forecasting questions. We use the same methodology to create train, validation and test splits, but use \textit{different news sources} to check if our model learns generalizable forecasting skills. 

\subsection{\ourdataset: An Open, Large-Scale Forecasting Training Dataset}

We now describe the specific composition of our training dataset. We use DeepSeek v3 as the sample creator and Llama-4-Maverick as the sample selector, with prompts in \Cref{sec:prompt_templates}.

\textbf{Generating samples.} One practical issue we face is that many top news sources, such as The Reuters and Associated Press (AP), have disallowed scraping even for CommonCrawl, due to the rise of commercial use in language model training~\citep{grynbaum2023times,longpre_consent_2025}. Still, we are able to collect articles from popular outlets spanning diverse geographies and topics. 

Particularly, for our training set, we start with $\sim248,000$ deduplicated English-language articles between June 2023 to April 2025 from \emph{Forbes}, \emph{CNN}, \emph{Hindustan Times}, \emph{Deutsche Welle}, and \emph{Irish Times}. The distribution is described in \Cref{fig:databreakdown}. From each article, the sample creator produces three forecasting samples, yielding $\sim745,000$ samples. \vspace{0.15cm}

\textbf{Filtering samples.} ~\Cref{tab:new_combined} (right) contains a breakdown of questions remaining after each filtering stage. 60\% of question-answer candidates are marked invalid --- most commonly because the article does not unambiguously resolve the question to the given answer. At this stage, zero questions remain from 40\% articles, and 21\% articles yield exactly one valid question, which we keep as is. For the 39\% with multiple valid questions, the sample selector picks the best one. Finally, we remove samples with numeric answers.\vspace{0.15cm}

\looseness -1 \textbf{Editing to fix leakage.} Despite explicit prompts to avoid it, over $40\%$ of selected questions directly contain the answer string. The sample selector's rewriting and rejection removes $\sim$90\% of such cases. We apply a string matching filter to remove the remaining questions with such direct leakage and finally keep only those questions which resolve after January 1, 2024. 

\begin{table}[t]
\vspace{-0.5cm}
\begin{minipage}{0.51\textwidth}
    \includegraphics[width=0.95\linewidth]{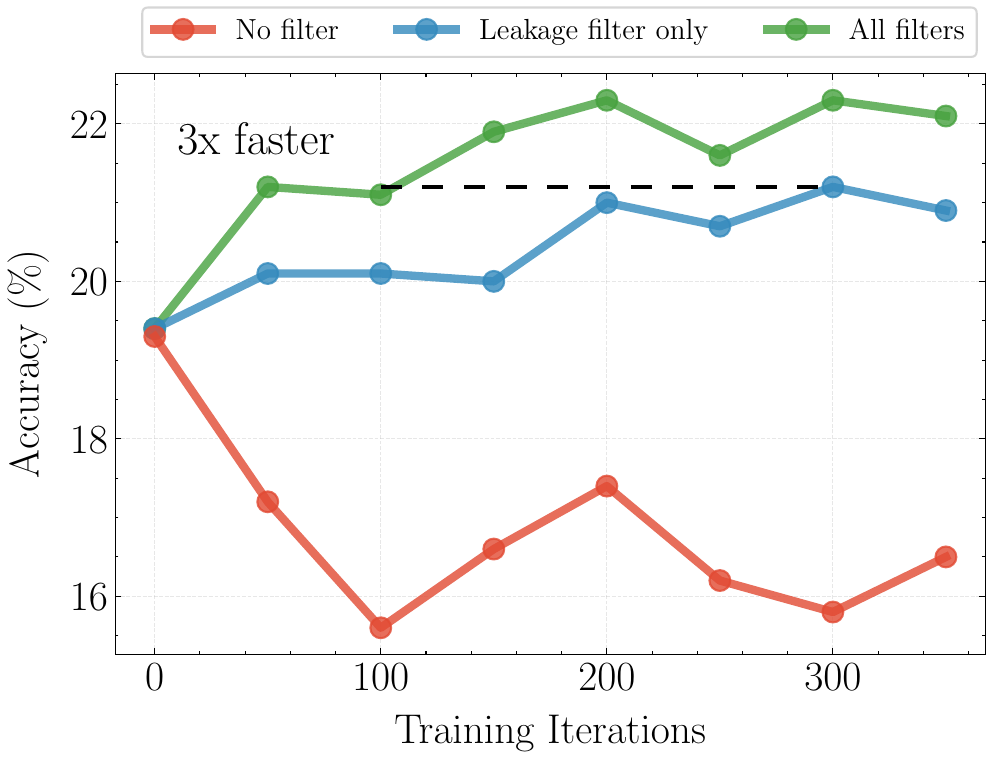}
\end{minipage}
\hfill
\begin{minipage}{0.48\textwidth}
    \resizebox{\linewidth}{!}{
\begin{tabular}{lc}
\toprule
\textbf{Stage} & \textbf{Number (\%Total)} \\ \midrule
Source Articles & 248,321 \\ \midrule
Question Generation & 744,963 \  (100\%) \\ 
Validation & 295,274 \   (40\%) \\ 
Best Question Selection & 157,260 \   (21\%) \\ 
Fixing Leakage & 150,500 \   (20\%) \\ 
Answer Type Filtering & \ 62,279  \ \   (8\%) \\
Resolving after 2024-01-01 & \ 52,183  \ \   (7\%) \\ \midrule
\textbf{Final Set} & \ 52,183  \   (7\%)
\\ \bottomrule 
\end{tabular}}
\end{minipage}
\vspace{-0.2cm}
\caption{(Left) \textbf{Benefits of our filtering stage.} Without leakage removal (red), the model worsens at forecasting, possibly learning shortcuts. With only the leakage removal step (blue), we find that achieving the same performance requires $3\times$ more compute and data. Applying all filtering steps (green) leads to higher accuracy. (Right) Number of questions after each filtering stage.}
\vspace{-0.4cm}
\label{tab:new_combined}
\end{table}

\looseness -1 \textbf{Validation Set.} We use the same recipe to create a validation set of 207 questions using 500 randomly sampled articles from TheGuardian in the month of July 2025. To ensure high-quality, we used \texttt{o4-mini-high}, a much more capable model than DeepSeek-v3, to generate the seed questions. It is $\sim10\times$ costlier, but followed the generation guidelines better, leading to $\sim$ 40\% retention rate (207 final questions out of 500 starting articles).\vspace{0.15cm}

\looseness -1 \contribution{Result 1: Filtering Improves Performance.}  ~\Cref{tab:new_combined} (left) shows the effect of our filtering steps. We train Qwen3-8B using RL with identical hyperparameters on three data variants: (red) 30,000 original generated questions, without any filtering; (blue) 30,000 samples obtained after the question editing step to remove leakage; and (green) 10,000 samples sourced from Forbes and included in \ourdataset. First, we observe the drastic impact of leakage in training. Training without leakage removal (red) worsens the model, perhaps due to shortcut learning. After the leakage removal steps, training improves the model (blue line). Yet, using all filtering stages (green line) leads to both higher accuracy and Brier score, in one-third the data and training steps. \vspace{0.15cm}

In \Cref{sec:binary_ablations}, we also ablate the effect of training on binary-only, free-form-only, and combined binary and free-form data for \texttt{Qwen3-8B}. We find that free-form data is crucial for improving open-ended forecasting, however, training solely on freeform data does not improve performance on binary Metaculus questions. Training with both kind of questions achieves the best trade-off, so we use this for our final training runs presented in \Cref{sec:results}. 

\textbf{Final training dataset.} Across stages, we remove $\sim90\%$ of questions, yielding a training set of 52K samples, each drawn from a unique article. We evaluated Qwen3-32B on this corpus while providing it the source article alongside the generated questions. The model achieves $95\%$ accuracy confirming high validity. We release this training dataset, as {\ourdataset}.

\begin{figure}[t]
\vspace{-0.25cm}
    \centering

    \begin{subfigure}[t]{0.49\textwidth}
        \centering
        \includegraphics[width=0.9\linewidth]{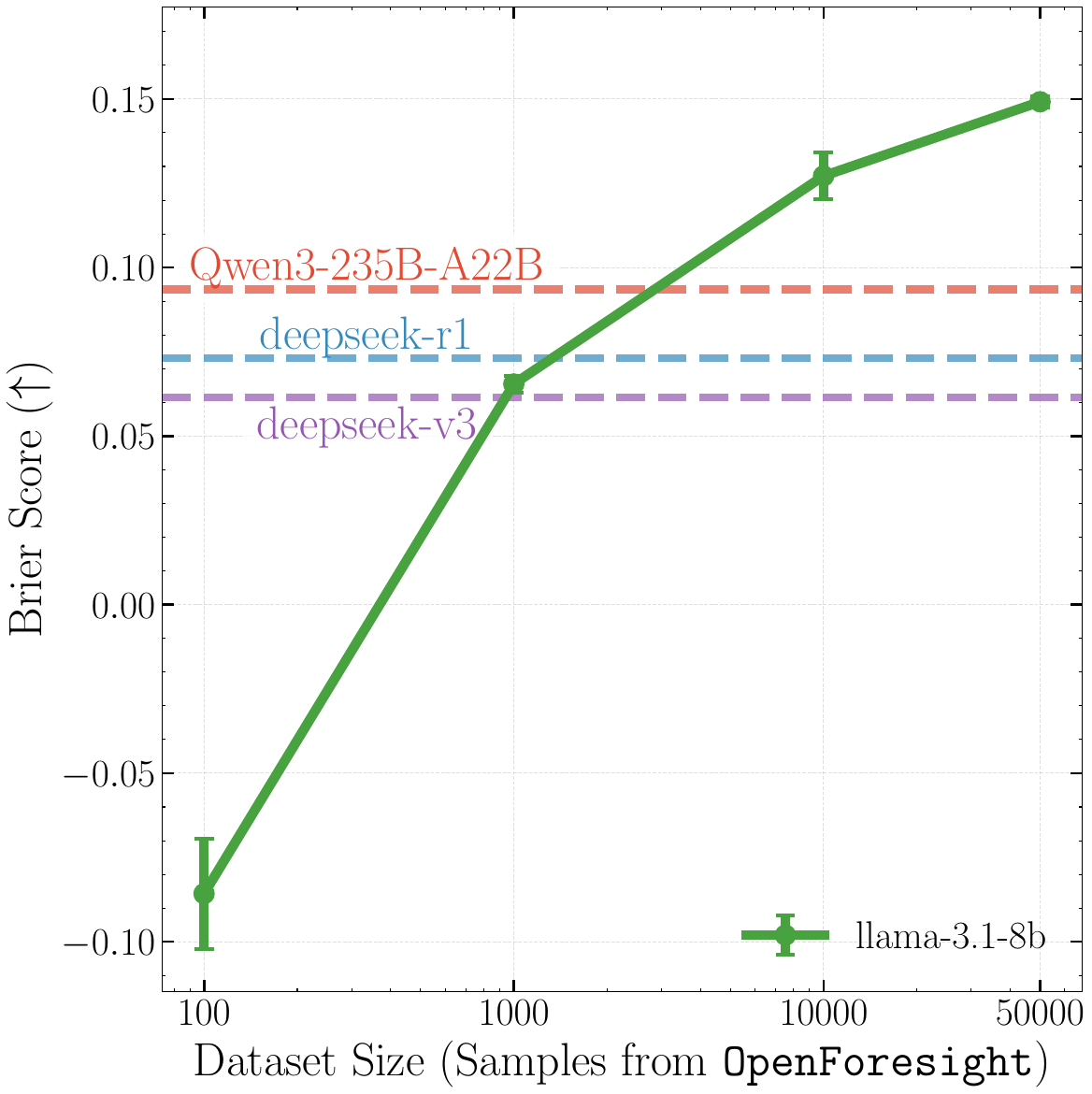}
        \label{fig:training_data_brier}
    \end{subfigure}
    \hfill
    \begin{subfigure}[t]{0.49\textwidth}
        \centering
        \includegraphics[width=0.9\linewidth]{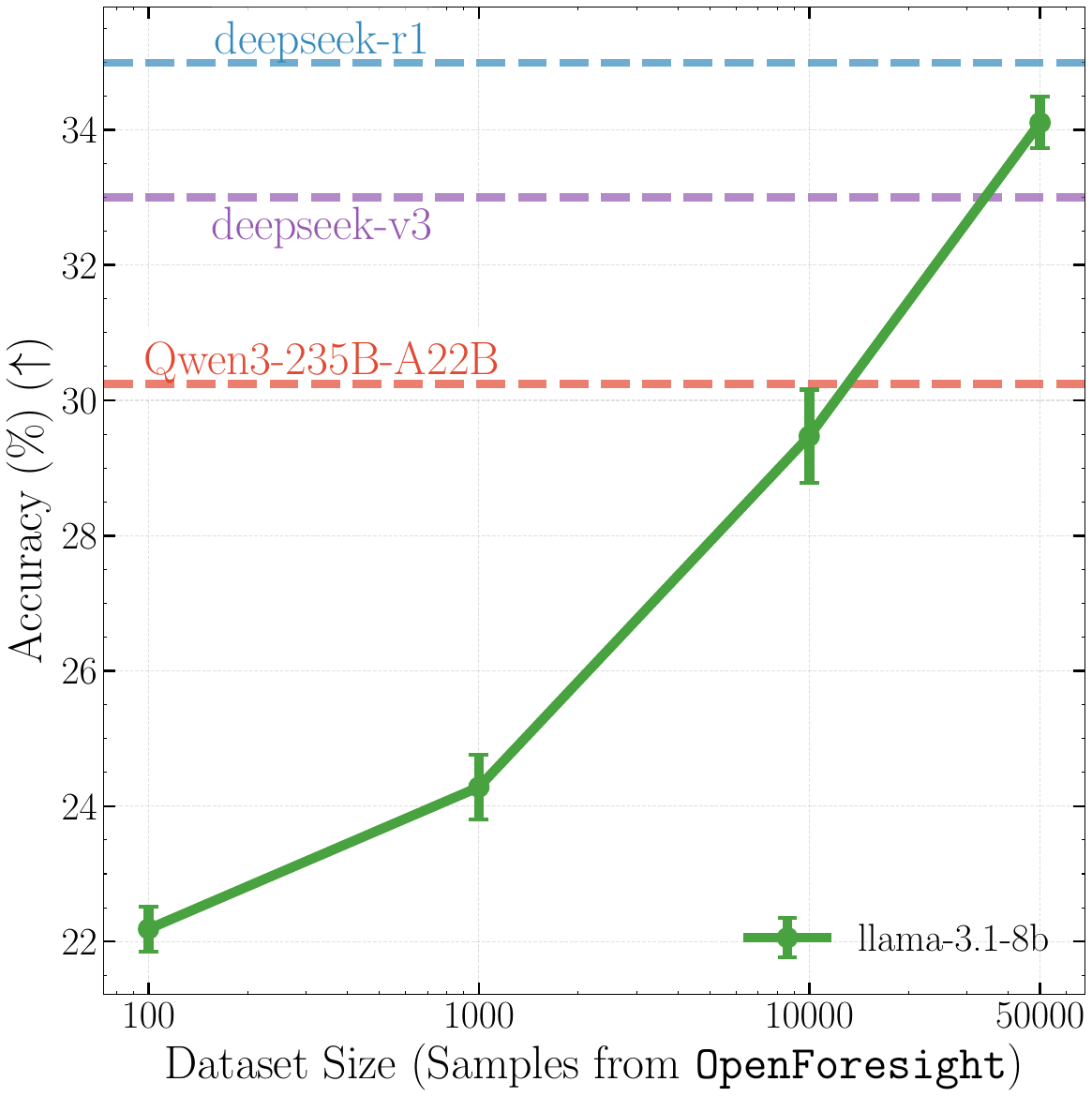}
        \label{fig:training_data_acc}
    \end{subfigure}

    \caption{
        \textbf{Benefits of scaling training data.} We take the best scores from training Llama-3.1-8B-Instruct on different sized subsets of \ourdataset. We see continued improvements in both accuracy and brier as the data size increases, eventually making Llama-3.1-8B-Instruct match or surpass much larger, more recent models.  
    }
    \vspace{-.4cm}
    \label{fig:training_data_performance}
\end{figure}

\looseness -1 \contribution{Result 2: Continued Improvements from Scaling Training Data.}  ~\Cref{fig:training_data_performance} shows the effect of scaling training dataset size. For each dataset size, we report test set (described later) results from the best validation checkpoint over prolonged training runs. We use Llama-3.1-8B-Instruct as it starts without any RL post-training. We see continued, and large gains in both accuracy and Brier score as we continue to scale up the training data, demonstrating the importance of our contribution in releasing \ourdataset.

%% file: sections/4_prediction.tex
\section{Prediction System}
\label{sec:predict}

\looseness -1 We now present intermediate results that guided the design decisions for our prediction system. We did not measure performance on the held-out test set throughout development, making decisions solely based on our validation set. \emph{We did not any find any notable difference between training in temporal order (sorted by resolution date), compared to training in a randomly shuffled order}. Below we present results which show the benefits of our reward design for RL training and retrieval system.  \vspace{0.15cm}

\textbf{Reward Design.} For training with RL, we compare three reward functions:
\begin{enumerate}
  \item \textbf{Only Accuracy} \textit{(Baseline):} \( R = \mathds{1}_{y \equiv y^*} \). Vanilla success rewards are commonly used in literature on LLM RL with verifiable rewards~\citep{guo2025deepseek}.
  \item \textbf{Only Brier score} \textit{(\citet{damani2025beyond}):} \( R = S'(q, y, y^*) = -q^2 + \mathds{1}_{y \equiv y^*}\cdot 2q \). This incentivizes both correct predictions and calibrated confidence estimates.
  \item \textbf{Accuracy + Brier score} \textit{(Ours):} \( R = \mathds{1}_{y \equiv y^*} + S'(q, y, y^*) \). We hypothesise that optimizing the Brier score alone might hurt exploration as when the model assigns a low confidence to its prediction, the correctness has a small impact on the Brier score. To fix this, we propose adding the accuracy term as well. In this case, even on hard questions, which merit low confidence, models get a large boost for correct predictions.
\end{enumerate}

\begin{figure}[t]

    \centering
    
    \includegraphics[width=0.99\textwidth]{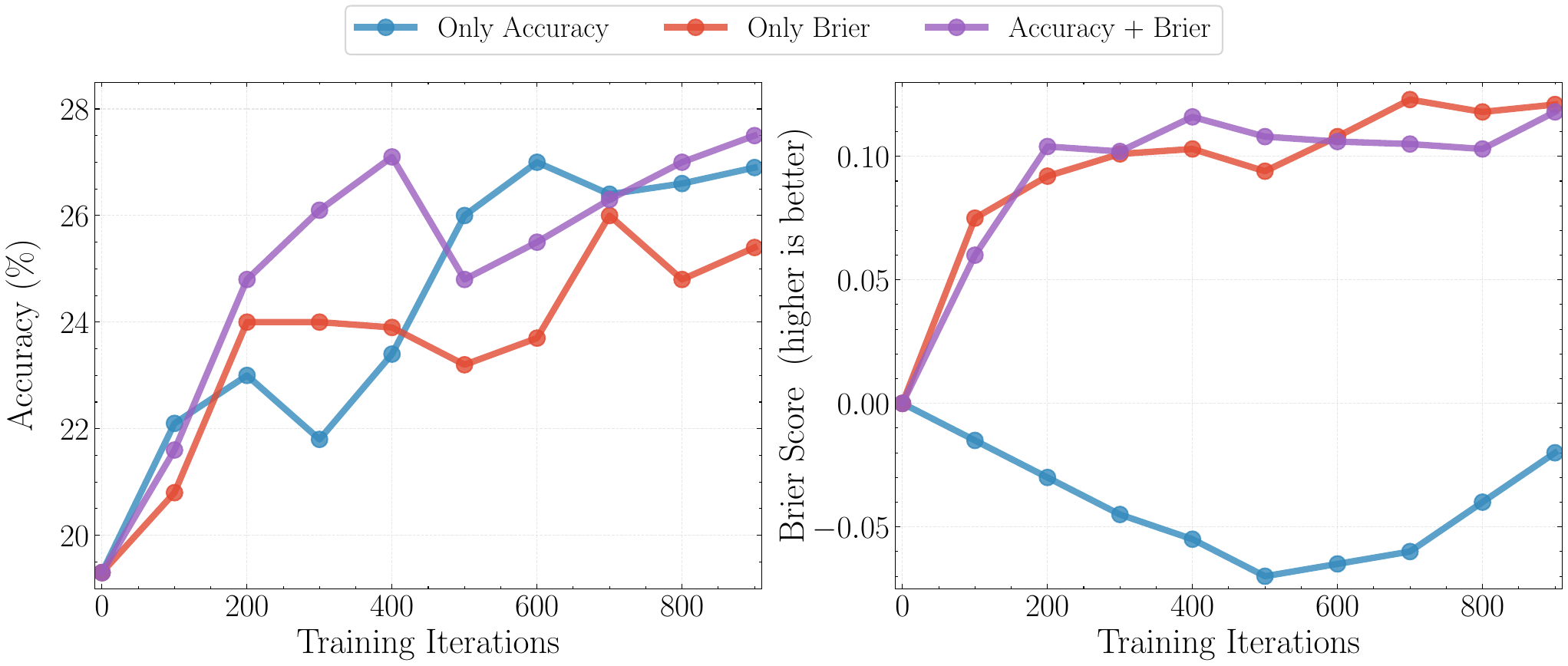}
     \vspace{-0.25cm} 
    \caption{\textbf{Accuracy + Brier score reward performs the best.} Accuracy alone leads to poor calibration. While brier score incentivizes both correct predictions and calibration, on hard questions with low confidence, it provides little signal on correctness. Adding the accuracy term boosts exploration in this situation.}
    \label{fig:reward_ablations}
    \vspace{-.5cm} 
\end{figure}
\looseness -1 \contribution{Result 3: Accuracy + Brier score as the reward improves performance.} \Cref{fig:reward_ablations} shows the validation set results of training with all three reward functions on the full \ourdataset \ dataset. We observe that optimizing accuracy alone (blue line) leads to negative brier scores, worse than a constant (0) baseline. In contrast, optimizing the Brier score alone (red line) also improves the accuracy, but to a lesser extent. Our proposed reward, accuracy + Brier (purple line), leads to the best performance on both metrics. It improves accuracy beyond the Brier alone while maintaining equal brier score on the validation set. Analyzing output distributions, we find that the brier-only trained model predicts ``Unknown'' with near-0 confidence in $\sim 40\%$ of samples, due to low reward for correct yet low-confidence guesses, which hurts exploration. In contrast, our proposed reward yields ``Unknown'' in only $\sim 4\%$ of samples, making low-confidence guesses on hard cases---improving both accuracy and Brier score. \vspace{0.15cm}

\looseness -1 \textbf{Retrieval.} Like prior work~\citep{zou2022forecasting, halawi2024approaching}, we provide the same relevant recent information across forecasting models. This provides them new evidence, or competing viewpoints to weigh, that was available before the resolution date, but potentially after the model's training cutoff. To prevent leakage issues~\citep{paleka2025pitfalls}, we use our offline CCNews corpus, and only use articles up to \emph{one month} before the question's resolution date. Our overall pool consists of 1 million articles across 60 different sources. We de-duplicate the articles and split each into fixed-size chunks (512 tokens) and embed each chunk with the Qwen3-embedding 8B model. During evaluation, we retrieve the top-$k$ most relevant chunks and append them to the model prompt in order as context. \vspace{0.15cm}

\begin{figure}[t]
    \centering
    \vspace{-0.25cm}
        \includegraphics[width=0.8\linewidth]{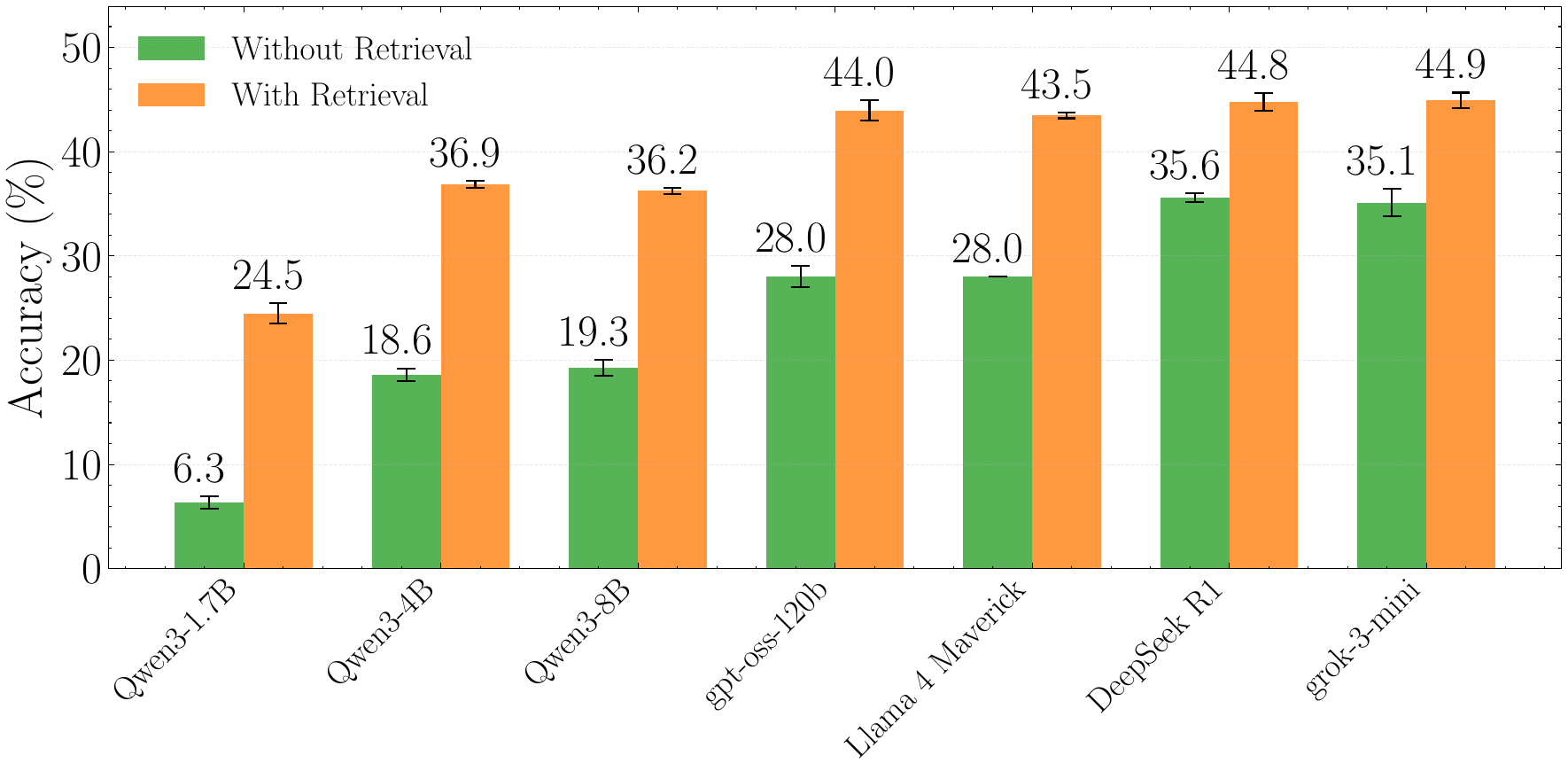}
       
        \label{fig:theguardian_retrieval_acc}
        \vspace{-.35cm} 
    
     \caption{\textbf{Retrieval improves accuracy across models.} We use the specialized Qwen3 8B embedding model to retrieve the 5 most relevant chunks (512 tokens) for each question. We take a cautious approach, using articles only until a month before the resolution date.}
     \vspace{-0.25cm}
\end{figure}

\looseness -1 \textbf{Retrieval leads to large improvements in accuracy.} As shown in \Cref{fig:theguardian_retrieval_acc}, providing retrieved information improves accuracy by $9-18\%$ across model families and sizes. In Appendix~\Cref{fig:articles_retrived}, we vary the number of retrieved chunks and find that improvements plateau after five chunks. Thus, we fix $k=5$ chunks for all evaluations henceforth. 

\textbf{Training the final forecasting system.} Based on the above design decisions guided by validation set performance, we now describe our final training methodology: We use the Qwen3-8B embedding model to retrieve the 5 most relevant chunks from news articles until a month before each question's resolution date. During training, we add a random number of such article chunks (between 0 to 5) in the prompt to make our forecaster generalizable to variable number of articles. We train the Qwen3-8B thinking model with GRPO on \ourdataset \ which has $\sim$ 50,000 samples and also include 2000 binary resolved questions from Metaculus (from 2024), both with retrieval. For the reward, we use our Accuracy + Brier score for free-form questions and only brier score for binary questions. We provide configuration details for training and evaluation in \Cref{sec:setupdetails}. 


%% file: sections/7_results.tex
\vspace{-0.2cm}
\section{Final Results}
\label{sec:results}

\looseness -1 We now present evaluations of our model, \ourmodel 8B. To avoid making decisions based on future information, we evaluate on test sets that were not observed until the end.\vspace{0.15cm}

\textbf{Open-ended Test Set.} Given the lack of \emph{open-ended} forecasting benchmarks that are still ``in the future'' for our models, we create our own test set with additional steps to ensure high quality. We first use our data creation recipe to generate an initial set of 1,000 questions between May to August 2025 using a stronger model, \texttt{o4-mini-high}. We draw from five diverse news sources: Al Jazeera English (global news, based out of Qatar), Time (global news, based out of USA) The Independent (UK focused), Fox News (USA focused), NDTV (India focused), with 200 questions selected from each. The choice of sources was made under the constraint of many established news sources have disallowed crawling of their articles starting 2025.
We deliberately use distinct sources from the training set to ensure that our model is learning generalizable forecasting skills, and not source distribution specific biases. Beginning from this initial set of 1000 questions, we perform additional filtering steps to prepare a high-quality test set: 
\begin{enumerate}
    \item We remove any potentially unanswerable questions (noise) by keeping only those which \texttt{grok-4.1-fast} could successfully answer with search tool access ($85\%$).
    \item To address the issue of late reporting in news outlets, we again use \texttt{grok-4.1-fast} with search tool to find the \textbf{earliest resolution date} for a given question. This is important to prevent leakage from retrieving articles with the true answer. We retain only those questions with resolution date after May 2025 ($64\%$).
    \item Finally, we manually filter the remaining questions to meet our quality checks, resulting in a final test set of $302$ questions. We provide more details like news source specific statistics in \Cref{sec:test_set_details}.
\end{enumerate} 

\textbf{External Datasets.} We use the FutureX benchmark~\citep{zeng2025futurex}, filtering to non-numeric, English, resolved forecasting questions and evaluating all models with our retrieval. This leaves 86 binary or multiple choice questions, between July to August 2025. For evaluating long-term predictions (without retrieval), we measure consistency metrics on binary questions up to 2028 as proposed by ~\citet{paleka2025consistency}, who show they correlate strongly with forecasting performance. Finally, to measure whether our forecasting training generalizes to calibration on standard benchmarks of LLM capabilities, we evaluate without retrieval on a challenging factuality benchmark, SimpleQA \citep{wei2024measuringshortformfactualitylarge}, and popular cross-domain reasoning benchmarks, MMLU-Pro and GPQA-Diamond. \vspace{0.15cm}

\begin{figure}[t]
\vspace{-0.25cm}
    \centering
    \includegraphics[width=0.9\linewidth]{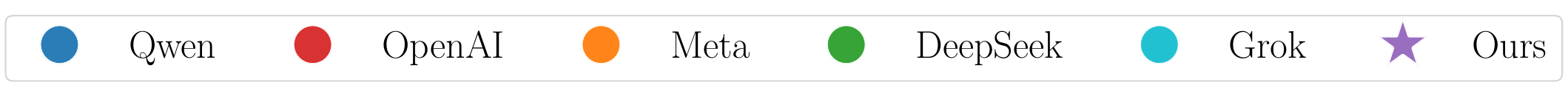}
    \hfill

    \begin{subfigure}[t]{0.49\textwidth}
        \centering
        \includegraphics[width=0.9\linewidth]{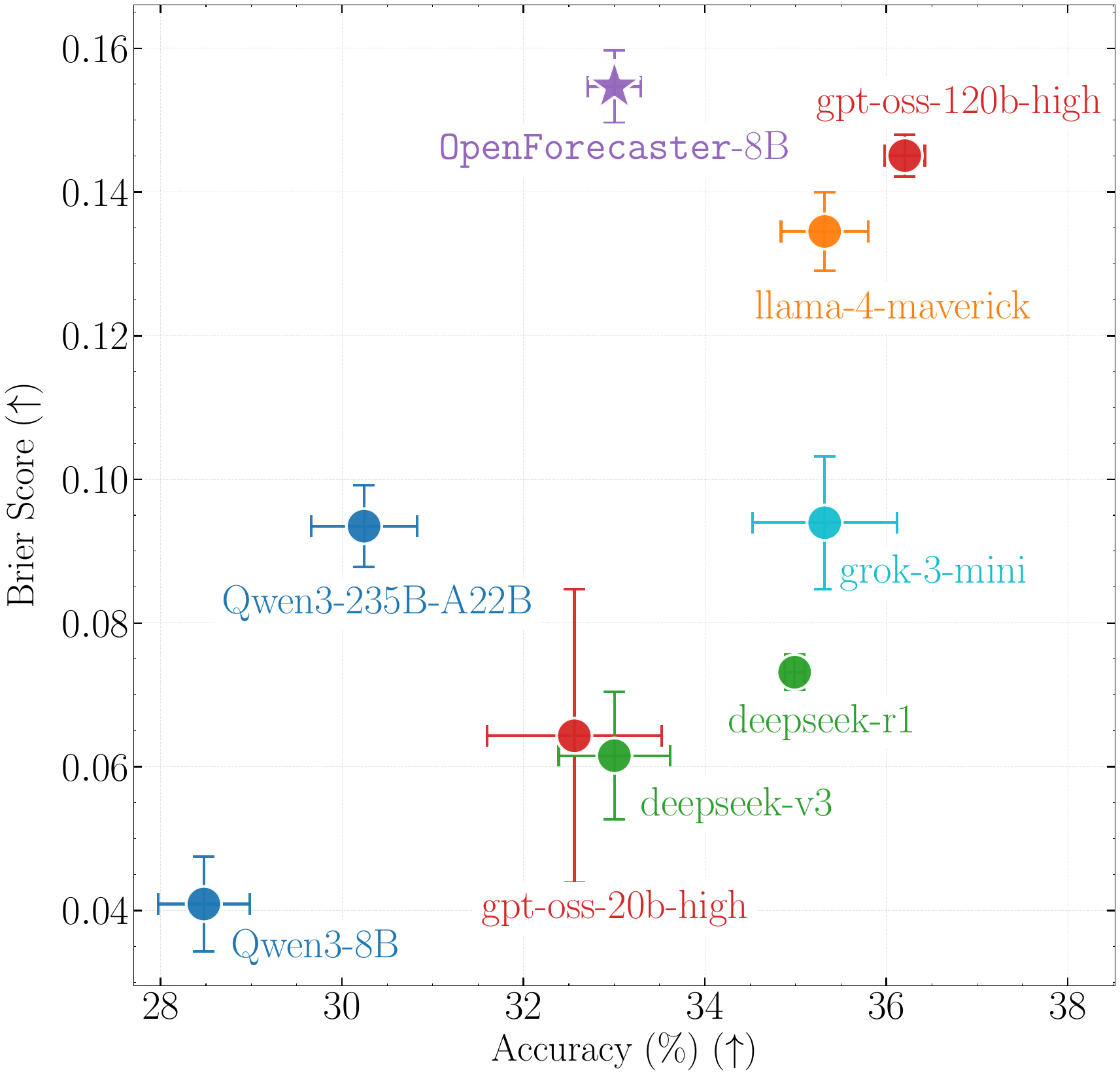}
        \caption{Results on \ourdataset\; May-Aug 2025.}
        \label{fig:test_set_performance}
    \end{subfigure}
    \hfill
    \begin{subfigure}[t]{0.49\textwidth}
        \centering
        \includegraphics[width=0.9\linewidth]{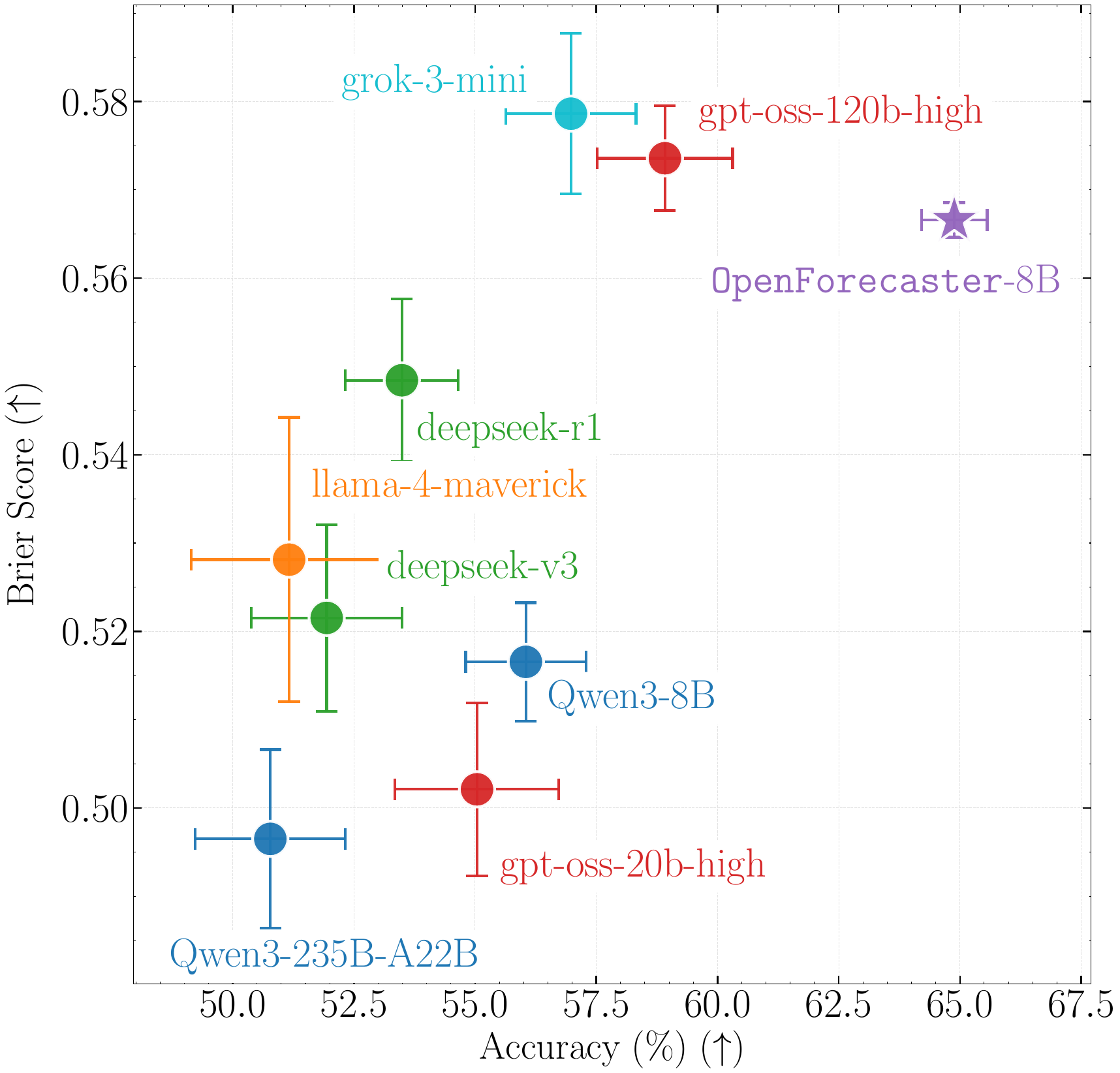}
        \caption{ 
            Results on FutureX July-Aug 2025.
             }
        \label{fig:futurex_performance}
    \end{subfigure}

    \caption{
        \textbf{Our forecasting training improves accuracy and calibration} both on open-ended questions in our test set,  and the external FutureX benchmark. It makes \ourmodel 8B competitive with much larger models that have knowledge cutoffs before May 2025.  
    }
    \vspace{-.2cm}
\end{figure}

\begin{figure}[h]
\vspace{-0.0cm}
    \centering
   
    \begin{subfigure}[t]{0.49\textwidth}
        \centering
        \includegraphics[width=0.95\linewidth]{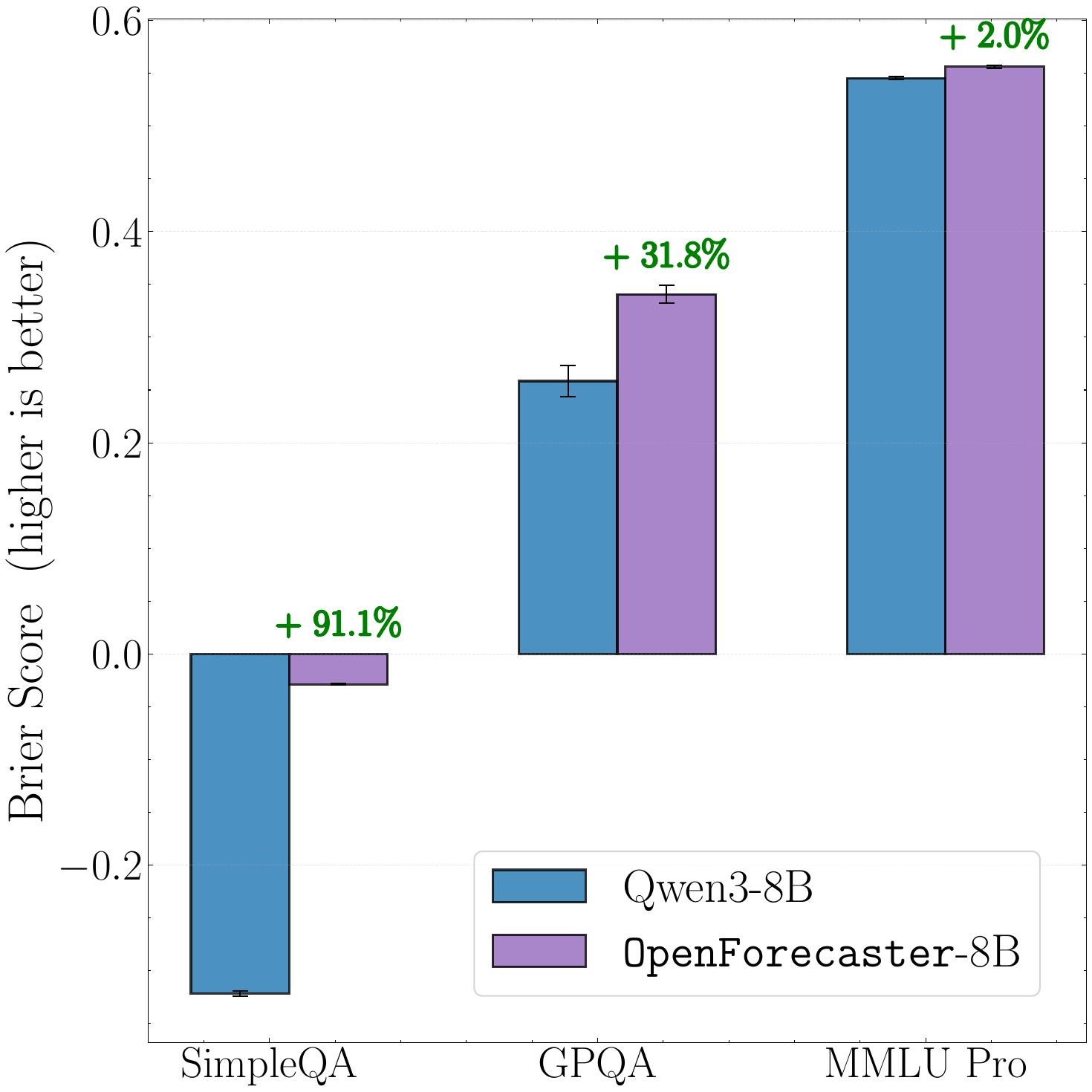}
        \caption{Brier score on general benchmarks.}
        \label{fig:simpleqa_performance}
    \end{subfigure}
    \hfill
    \begin{subfigure}[t]{0.49\textwidth}
        \centering
        \includegraphics[width=0.95\linewidth]{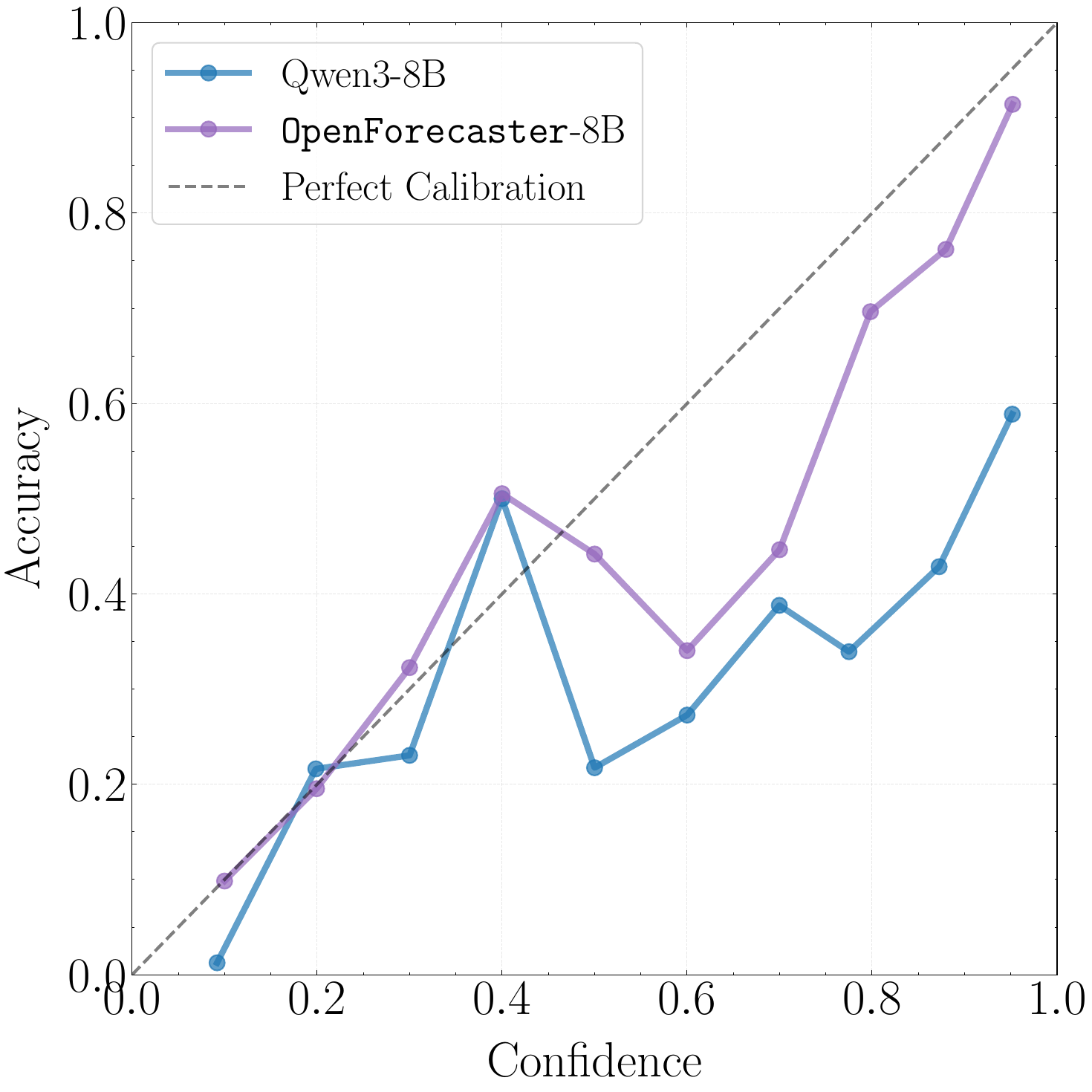}
        \caption{Calibration curve on our test set. }
        \label{fig:calibration_curve}
    \end{subfigure}
\vspace{-.2cm}
    \caption{Calibration of the models improve significantly after training on \ourdataset \ both on (a) out of distribution benchmarks and (b) on \ourdataset \ test set.}
    
    \vspace{-.25cm}
\end{figure}

\contribution{Result 4: Training on our dataset leads to large improvements in forecasting.}
\Cref{fig:test_set_performance} shows performance of models on our held-out test set of open-ended forecasting questions. On the Brier score (Y axis), the primary metric recommended for forecasting~\citep{tetlock2016superforecasting}, as it measures both accuracy and calibration, \ourmodel 8B outperforms even GPT OSS 120B. Our improvements are not merely from calibration, the predictions also become more accurate (X axis), beating Qwen3 235B, but are a bit behind others. Training on \ourdataset \ also improves models from other families like Llama and Gemma as we show in \Cref{sec:extra_forecasting_results}. We saw a particularly large (+25\% absolute improvement in accuracy) improvement for Llama 3.1 8B Instruct, surpassing the much larger Qwen3-235-A22B. We also show model accuracy by month in ~\Cref{fig:monthly_accuracy}. \vspace{0.15cm}

\looseness -1 On FutureX, our model has the strongest accuracy by a large margin, even compared to much larger proprietary counterparts. It is close to the best for Brier score as well. Finally, our training leads to more consistent long-term predictions, improving 44\% on arbitrage metrics, and 19\% on frequentist metrics, with detailed results in \Cref{sec:consistency}. Our model also improves on questions from Metaculus prediction market albeit staying behind few larger models which we show in \Cref{sec:metaculus_results}.


\contribution{Result 5: Calibration training for forecasting generalizes to other domains.} \Cref{fig:simpleqa_performance} shows downstream improvements in calibration across SimpleQA, GPQA-Diamond and MMLU-Pro (green text highlighting the relative improvement).
This calibration can then be used to reduce hallucinations, for example abstaining on questions the model is not confident about, using simple rules like \texttt{if probability < 0.1, replace prediction with ``I do not know''}\vspace{0.15cm}


%% file: sections/9_conclusion.tex
\section{Conclusion}
\label{sec:conclusion}

\looseness -1 In this paper, we show how to curate data for \textit{scalable training} of \textit{open-ended forecasting}. The results are promising, an 8B model finetuned on our data becomes competitive with proprietary models like GPT-OSS-120B, DeepSeek-R1, and Grok-3-Mini. Calibration improvements from forecasting training generalize out of distribution. A few limitations remain. For example, we only use news to create forecasting questions, which leads to a distributional bias. The news also reports some events late, such as scientific breakthroughs, and this can make such questions easier to ``predict'' than others in our dataset. This should not affect relative performance comparisons between models though. We also do not consider long-form forecasts, as it is unclear how to grade these. Overall, open-ended forecasting, being a challenging and highly valuable task, offers exciting directions to pursue across research communities. A strong forecaster needs to reason about uncertainty, efficiently seek new information, and make optimal Bayesian updates to its world model, long-standing challenges in the quest for general intelligence. Scaling up end-to-end training of open-ended forecasting systems may lead to emergent improvements in such capabilities. By open-sourcing all our artefacts, we hope to spark more research on this important direction.

%% file: sections/11_ack.tex
\section*{Acknowledgments}
\label{sec:acknowledgments}

\looseness -1 We thank Douwe Kiela, Alexander Panfilov, Tim Rockt{\"a}schel, and Guanhua Zhang for valuable discussions. We thank Maksym Andriushchenko, Arvindh Arun, Alessandro Bifulco, Paras Chopra, and Daniel Paleka for helpful feedback on our draft. We thank CCNews and TheGuardian for providing free access to news articles, Thinking Machines for providing Tinker API research credits, and Contextual AI for letting us test their retrieval system.

%% file: sections/16_appendix.tex
\newpage
\appendix
\part{Appendix}

\localtableofcontents
\clearpage
\input{sections/12_brier}

\input{sections/17_extra_results}

\input{sections/22_train_details}
\input{sections/13_extra_exps}
\input{sections/19_qualitative}
\input{sections/21_failure_modes}

\input{sections/14_prompts}

%% file: sections/12_brier.tex
\section{Adapting Brier Score to free-form responses}
\label{sec:freeform_brier}

Let $\mathcal{X}$ be the set of open-ended forecasting questions; and $\mathcal{Y}$ the set of short textual answers. Let  $x \in \mathcal{X}$ be a resolved forecasting question and $y^\star$ be the ground truth answer (as the question has already resolved). We ask the forecaster to respond with its best guess answer $y$, and the probability $q$ they assign to that being the true outcome. We evaluate this prediction tuple <$y, q$> using the Brier score \citep{mucsanyi2023trustworthy} but adapt it to our setting. For a $K$-class outcome space $\mathcal{Y}$ with reported distribution $q$ and true class $y^*$, the (multi-class) Brier score is
\[
S(q,k) = -\sum_{y\in\mathcal{Y}} (q_y - k_y)^2
= -(q_{y^*}-1)^2 - \sum_{y\neq y^*} q_y^2,
\]
where $k$ is the one-hot encoding with $k_{y^*}=1$. In our open-ended setting, $\mathcal{Y}$ is not predefined but rather its instances are provided by the forecaster. For simplicity, we elicit only a \textbf{single guess} $y$ with probability $q\in[0,1]$ and assume the forecaster's probability is $0$ for all other (semantically different) answers $y^\prime \neq y$.\footnote{This is technically incorrect to assume as the forecaster may have non-zero probability for guesses other than $y$. Ideally the forecaster should report all its guesses which have non-zero probability (with the multi-class Brier scoring rule still being applicable) but we leave exploring this direction for future work.} Applying the multi-class Brier scoring rule in such a case induces a simplified score:
\[
S(q,y,y^*) =
\begin{cases}
-(q-1)^2 - 0= -1 + 2q - q^2, & \text{if } y \equiv y^*,\\
- (0 - 1)^2 - q^2 = -1 - q^
2, & \text{if } y \neq y^*.
\end{cases}
\]
Dropping the constant $-1$ yields
\[
S'\!(q,y,y^*) =
\begin{cases}
1-(q-1)^2, & \text{if } y \equiv y^*,\\
-\,q^2, & \text{if } y \neq y^*,
\end{cases}
\qquad
\label{eq:rlcr2}
\]
\begin{wrapfigure}[9]{r}{0.5\textwidth} 
    
    \centering
    \includegraphics[width=0.9\linewidth]{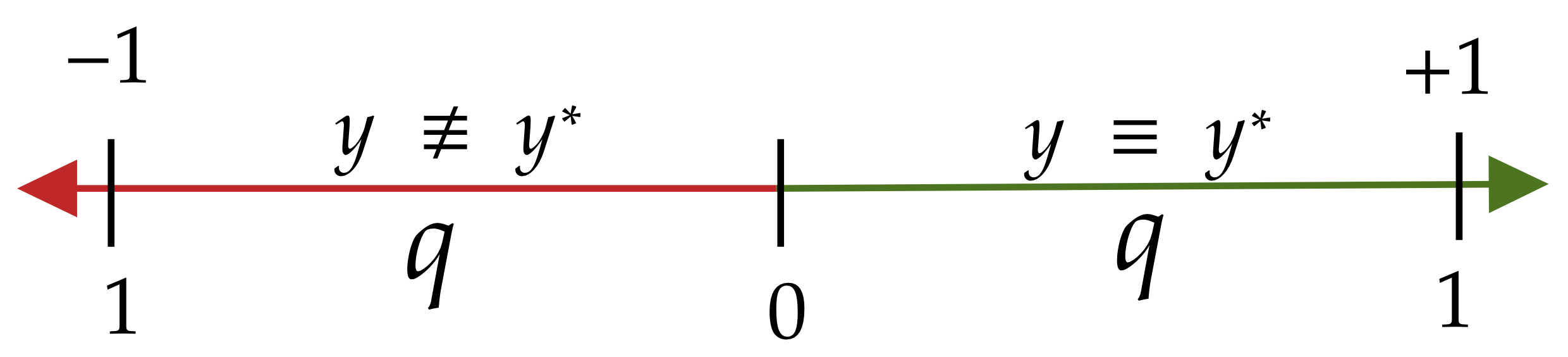}
    \caption{Illustration of the Brier score when adapted to free-form response with answer $y$ and probability $q$.}
    \label{fig:brier_score_visualization}
   
\end{wrapfigure}

which shifts the range from $[-2,0]$ to $[-1,1]$ while providing a more natural interpretation: predicting $q=0$ gives a baseline $0$ regardless of $y$; correct answers receive positive scores, incorrect answers negative scores; and magnitude scales quadratically with confidence. We report $S'$score (visualized in ~\Cref{fig:brier_score_visualization}) as the \emph{Brier score} in this paper. 

Recent work by ~\citet{damani2025beyond} shows that this metric is a proper scoring rule, incentivizing both high accuracy and truthful reporting of probability on the answer that seems most likely. However, note that what we call the Brier score here is distinct from the Brier score considered by \citet{damani2025beyond}. Their Brier score is the one traditionally used for evaluating binary outcomes while ours is for free-form responses. Yet, we can show that our Brier score is \textit{same} as the \textbf{training reward} considered by them.
\clearpage

%% file: sections/17_extra_results.tex
\section{Additional Results}
\label{sec:extra_results}

\subsection{Ablation: Using Prediction Market Binary Data}
\label{sec:binary_ablations}

\begin{figure}[h]
  \centering
  \begin{subfigure}[t]{0.98\textwidth}
    \centering
    \includegraphics[width=\linewidth,keepaspectratio]{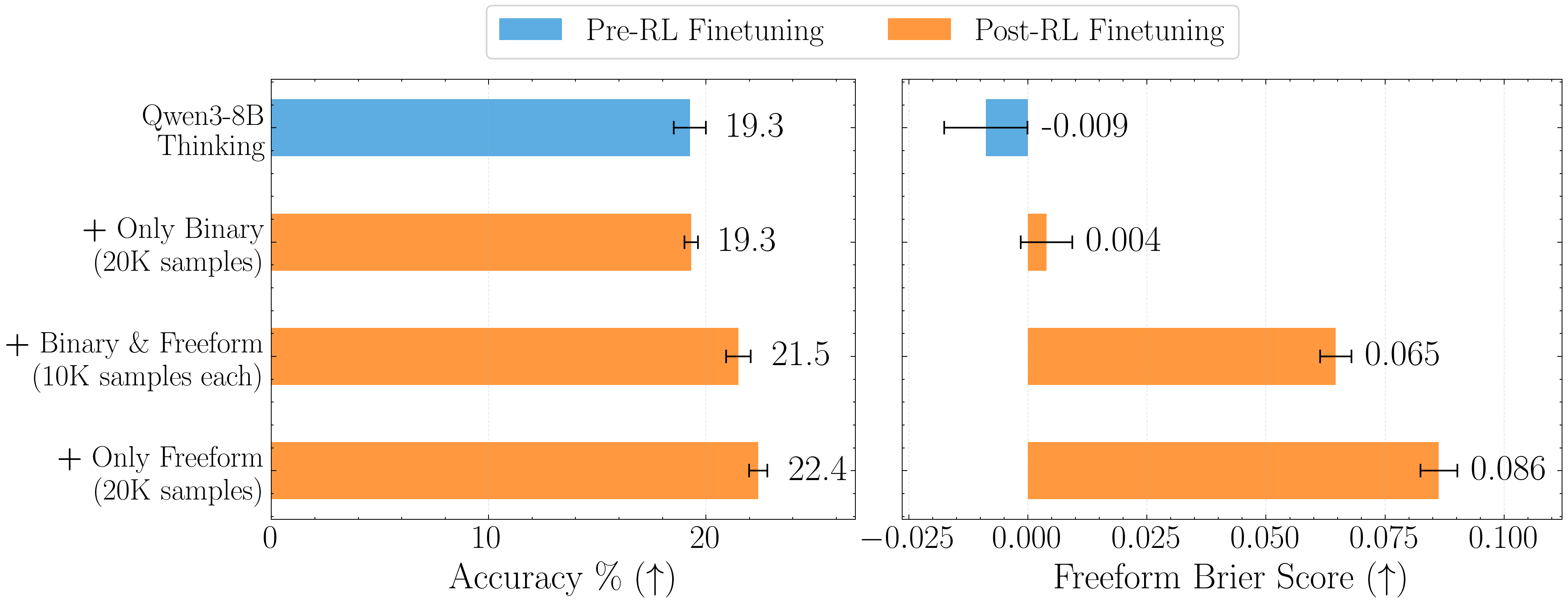}
    \caption{Performance on our \textbf{Validation Set} composed of question from TheGuardian news source from July 2025.}
    \vspace{0.3cm}
    \label{fig:binary_ablations_guardian}
  \end{subfigure}
  \hfill
  \begin{subfigure}[t]{0.98\textwidth}
    \centering
    \includegraphics[width=\linewidth,keepaspectratio]{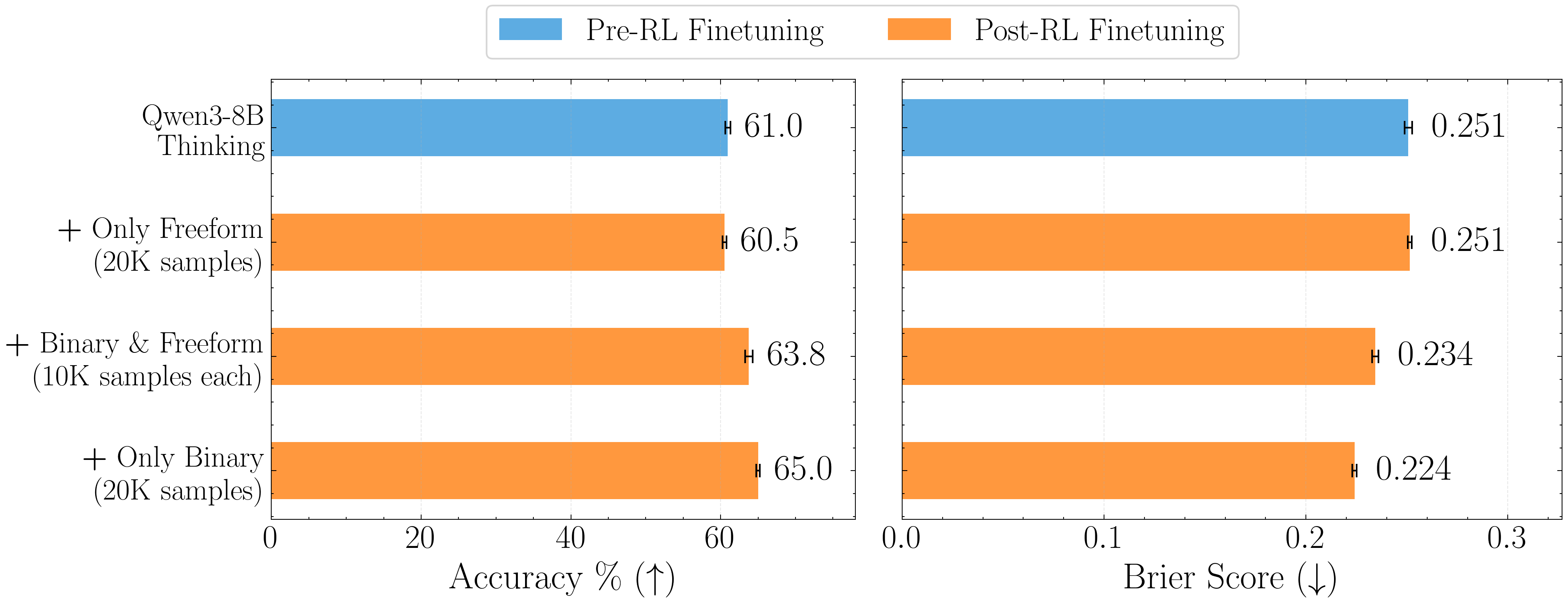}
    \caption{ Performance on \textbf{Metaculus binary} questions resolved in May–July 2025.}
    \label{fig:binary_ablations_metaculus}
  \end{subfigure}

  \caption{\looseness -1 \textbf{Performance of different data ablations.} 
  We evaluate performance after training on 3 different supervision signals: 
  (i) only binary data (20K samples), 
  (ii) only freeform data (20K samples), and 
  (iii) both binary and freeform data (10K samples each) for data-matched comparison. 
  (a) Accuracy and freeform Brier score of the initial and post-RL model on our Validation Set from July 2025. 
  (b) Accuracy and binary Brier score of initial and post-RL model on volume-filtered binary questions resolved between May and July 2025 on Metaculus. 
  \textit{We find training on binary questions hurts performance on open-ended forecasting, but is necessary to retain performance on binary prediction market questions.}}
  \label{fig:binary_ablations}
\end{figure}

We ablate supervision type with \texttt{Qwen3-8B} using three size-matched settings (Figure~\ref{fig:binary_ablations}). For \emph{binary-only}, we curate \textbf{20K} resolved markets from Manifold, volume-filtered to ensure engagement; because many markets resolve slowly, this set spans the past five years. For \emph{free-form only}, we use \textbf{20K} pipeline-generated, usable questions from Forbes articles. For the \emph{binary+free-form mix}, we take \textbf{10K} Manifold + \textbf{10K} Forbes questions to keep total examples constant. The goal is to isolate which \emph{learning signal}---binary resolution vs.\ open-ended outcome specification---most effectively trains calibrated forecasters under identical compute and token budgets.

On the free-form test set (Fig.~\ref{fig:binary_ablations} Left), post-RL performance improves most with \emph{free-form only} supervision (Accuracy \(19.3\%\!\rightarrow\!22.4\%\); Free-form Brier \(-0.009\!\rightarrow\!0.086\)). Mixing binary and free-form also helps (Brier \(0.065\)), whereas \emph{binary-only} yields minimal gains on free-form evaluation (Brier \(0.004\)). On Metaculus (binary) (Fig.~\ref{fig:binary_ablations} Right), both \emph{binary-only} and the \emph{mixed} setting improve accuracy and Brier, with the \emph{binary+free-form} mix offering the best overall trade-off across testing formats. Our gains by training on binary-only format are consistent with prior work by \citet{turtel2025outcome, turtel2025llms}. However, we do not arrive at a single unanimous recipe: free-form data is essential for open-ended forecasting, while combining formats appears Pareto-optimal across binary and free-form evaluations. Practically, it seems training on a \emph{mixture} of question styles provides the most robust gains across tasks.


\subsection{Varying models and evaluation months}
\label{sec:extra_forecasting_results}

\begin{wrapfigure}[18]{r}{0.5\textwidth} 
    \centering
    \includegraphics[width=0.95\linewidth]{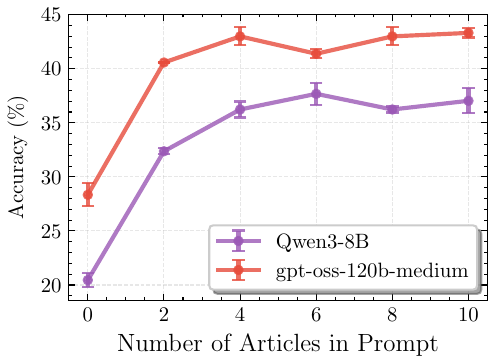}
    \caption{\textbf{Improvements from retrieval plateau at $\sim 5$ chunks.} We show the accuracy of both a large GPT-OSS-120B model and a small Qwen3-8B model which we finetune.}
    \label{fig:articles_retrived}
\end{wrapfigure}

In ~\Cref{fig:articles_retrived} we observe that while the first few article chunks that are retrieved lead to large improvements, at around five articles, improvements plateau, both on our Qwen3-8B and also other large models like GPT-OSS-120B. Thus, unless otherwise specified, we use 5 articles for all evaluations and training in this work.

\begin{figure}[h]
    \centering
    
    \begin{minipage}[t]{0.48\textwidth}
        \centering
        \includegraphics[width=\linewidth]{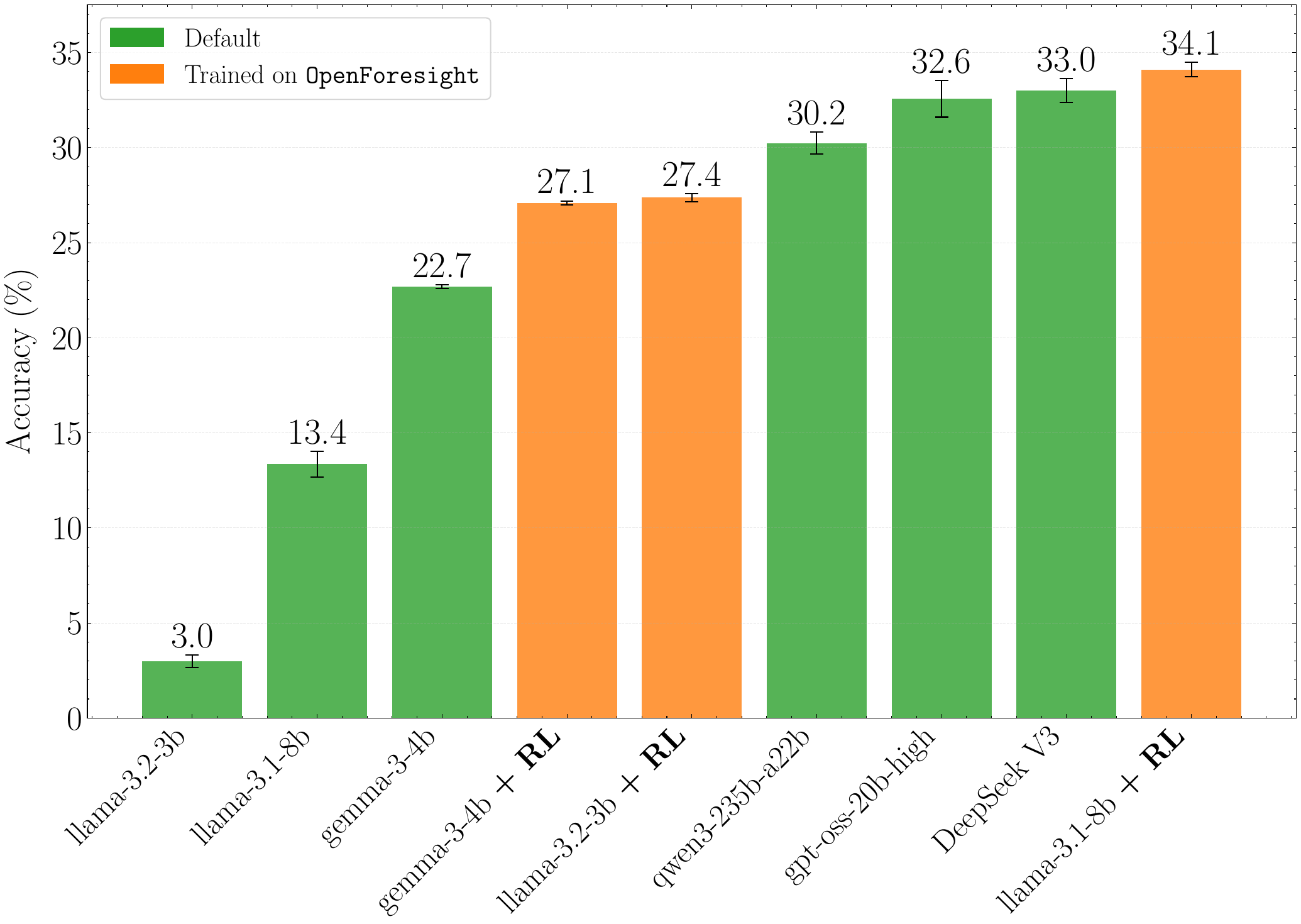}
    \end{minipage}
    \begin{minipage}[t]{0.48\textwidth}
        \centering
        \includegraphics[width=\linewidth]{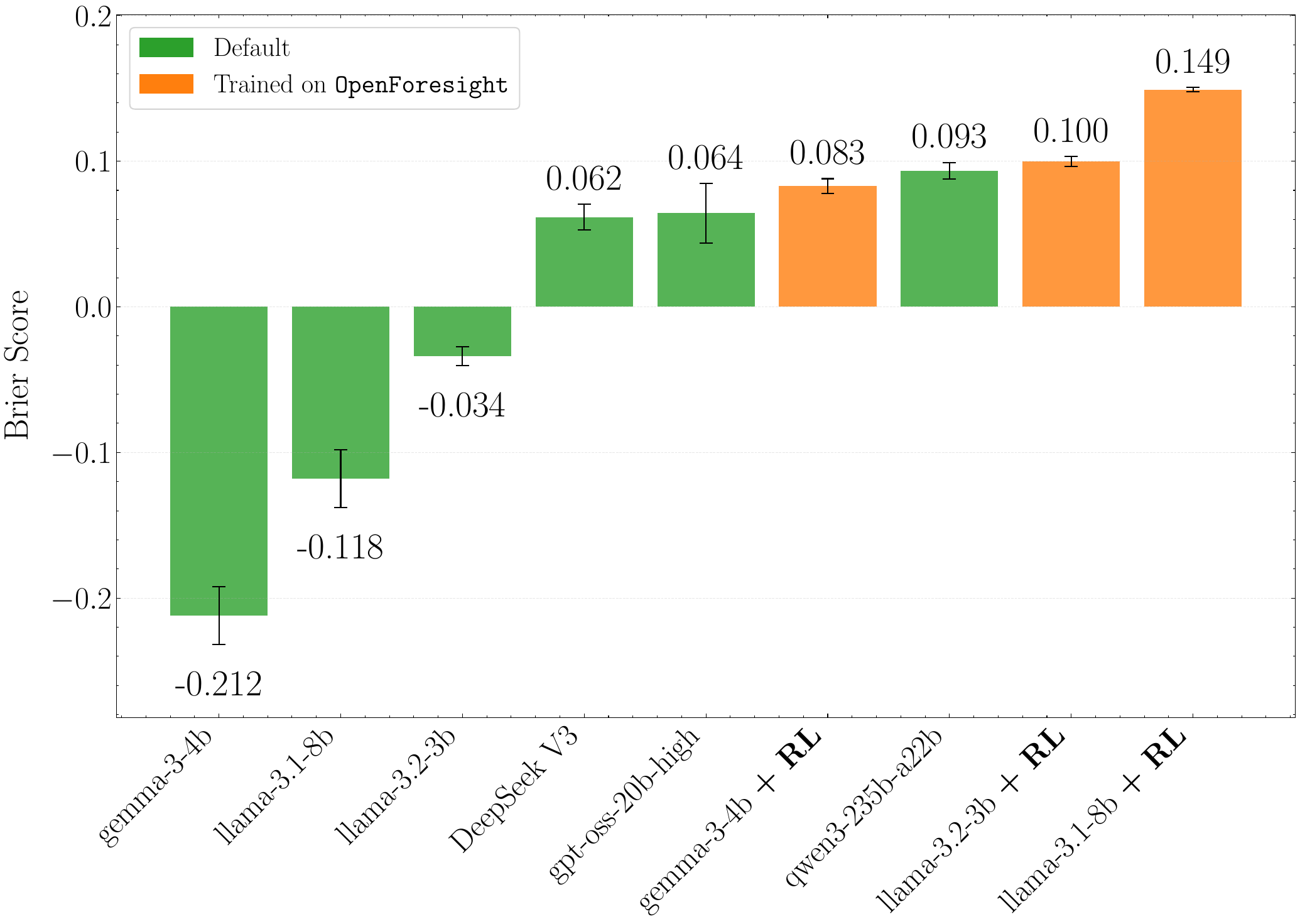}
        \vspace{-.2cm} 
    \end{minipage}
    
    \caption{Performance of models from Llama and Gemma family on our test set.} 
    \label{fig:extramodels_performance}
\end{figure}

\paragraph{Improvement on non-Qwen models.} 
Our training data \ourdataset \ can be used to improve models across different families. In Figure~\ref{fig:extramodels_performance} we show improvements for Llama-3.1-8B-Instruct, Llama-3.2-3B-Instruct and Gemma-3-4B-Instruct. We see particularly large improvements in both accuracy and Brier score for Llama due to both: poor initial performance, but also surprising amenability to RL training with our data as the final performance exceeds much larger models like Qwen3-235B-A22B and DeepSeek-v3. 

\begin{figure}[h]
    \centering\includegraphics[width=0.75\linewidth]{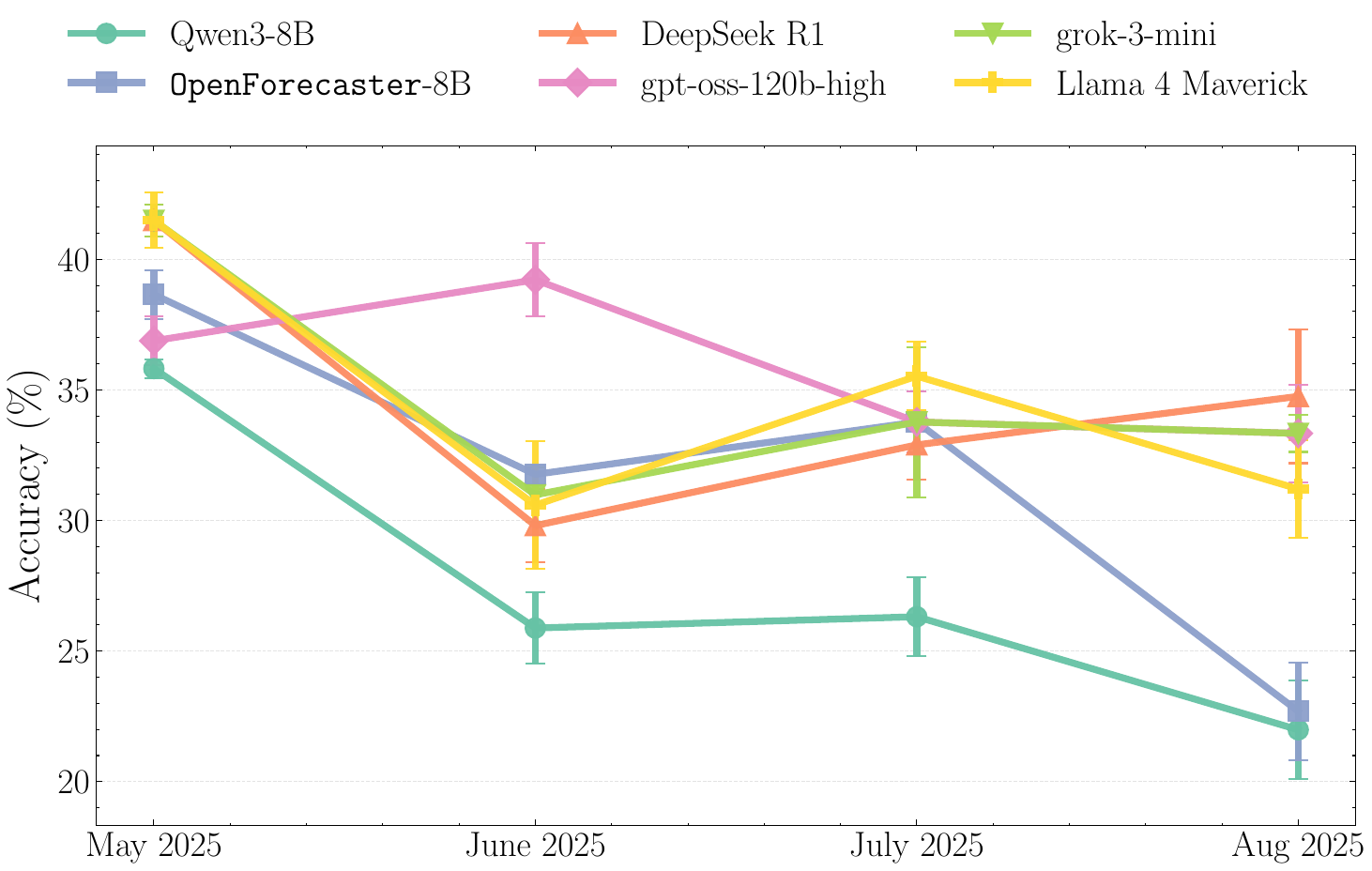}
        \caption{{Monthly accuracy of the models on our test set.} }
        \label{fig:monthly_accuracy}
        \vspace{0.2cm}
    
  \includegraphics[width=0.75\linewidth]{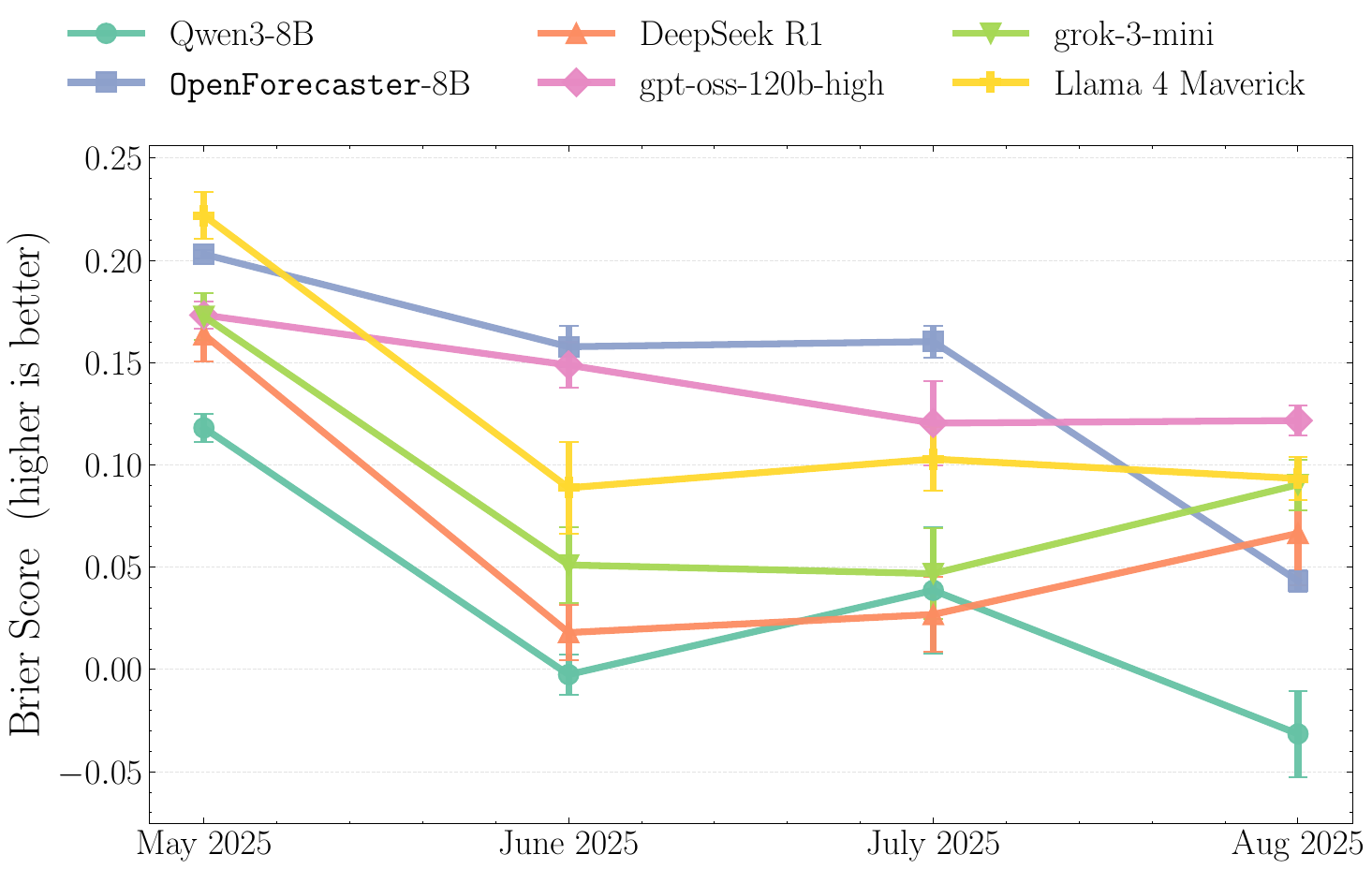}
    \caption{{Monthly Brier score of the models} }
    \label{fig:monthly_brier}
    \vspace{-.2cm} 
\end{figure}

\paragraph{Results over time.} As our test set is derived from articles from May to August 2025, so we split the questions by resolution date to get monthly performance of the models. Breaking down by month, our test has 94 questions resolving in May, 85 in June, 76 resolving in July and 47 resolving in August. We have lower number of questions in later months due to our filtering strategy for addressing late reporting in news. We first generated roughly equal number of questions per month and post-hoc filtered the ones whose true resolution date (found using grok-4.1-fast with search) was before May'25. 

For monthly performance, our hypothesis is that as we go further into the future, forecasting should become more difficult leading to lower performance. In \Cref{fig:monthly_accuracy} and \Cref{fig:monthly_brier}, we find that the accuracy and brier score of the models indeed drops gradually month-by-month consistent with our hypothesis. We also find that our trained models are consistently better than the original versions and also better than all other models in Brier score.

\subsection{Ablation with Supervised Finetuning}
\label{sec:sft_ablation}

\vspace{-.2cm}
Here we study what would be the benefit if we add a supervised finetuning (SFT) stage in our training process? While we start from the RL trained Qwen3 thinking models, they are far behind proprietary models as shown in~\Cref{fig:theguardian_retrieval_acc}. Several frontier model training reports~\citep{guo2025deepseek} mention using an SFT stage as a warm start before RL. We choose Grok-3-Mini to generate forecasting reasoning traces for SFT, as it has high performance, low cost, and provides the full reasoning trace through the API. 
Specifically, we construct a dataset of 10,000 questions from \textit{The Guardian} dated January–March 2025, beyond Grok-3-mini's reported knowledge cutoff of June 2024. Obtaining Grok-3-Mini's reasoning traces on this data costed $15$ USD. For distillation, we randomly choose the number of articles to put in its prompt (0 to 10) so that the student model can reason with any number of articles.  

\begin{figure}[t]
\vspace{-0.25cm}
    \centering
    \includegraphics[width=0.9\linewidth]{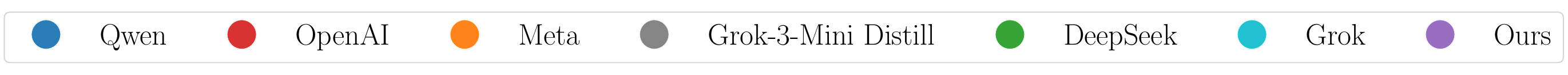}
    \hfill

    \begin{subfigure}[t]{0.49\textwidth}
        \centering
        \includegraphics[width=0.9\linewidth]{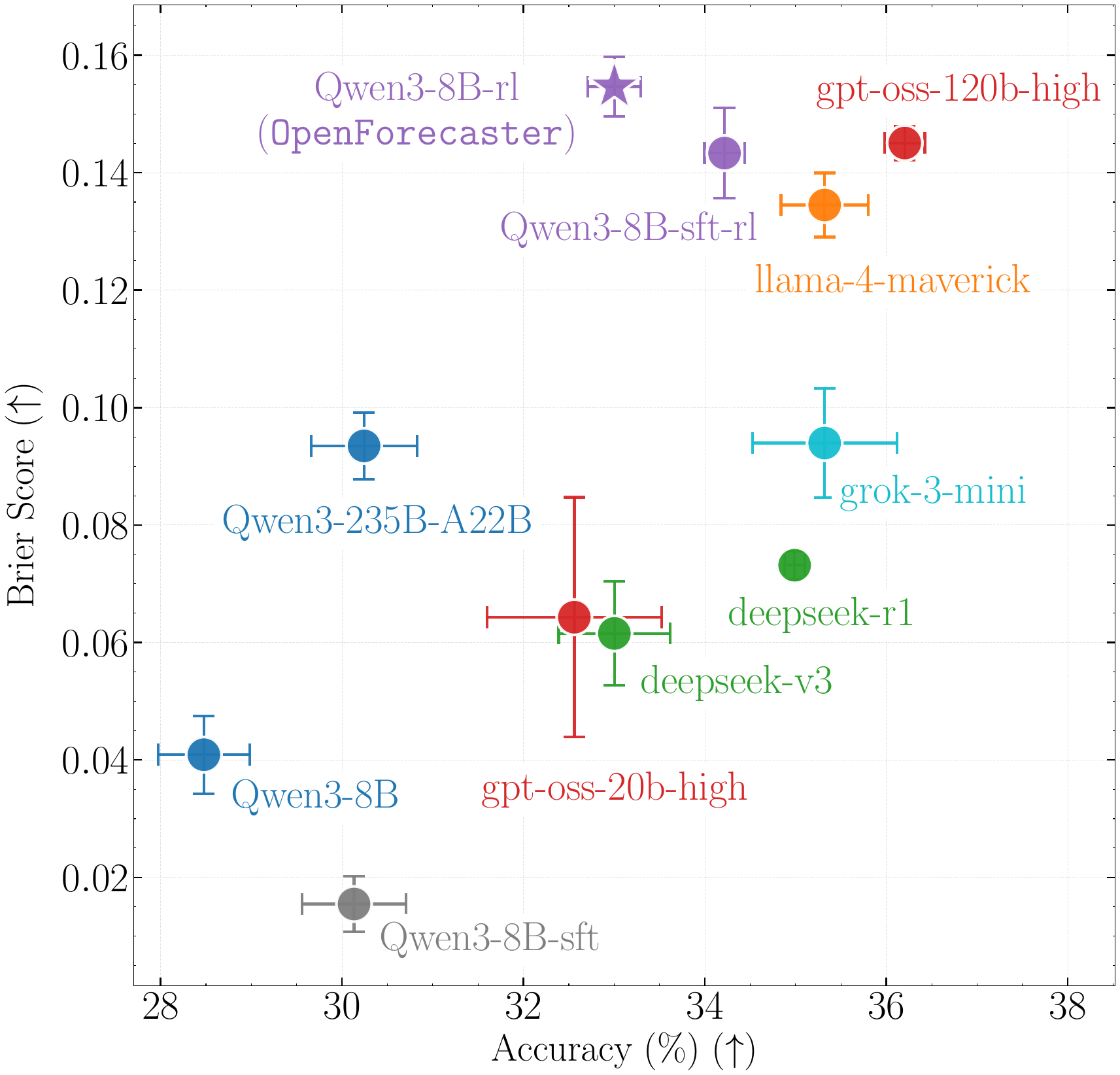}
        \caption{Results on our curated test set.}
        \label{fig:sft_test_set_performance}
    \end{subfigure}
    \hfill
    \begin{subfigure}[t]{0.49\textwidth}
        \centering
        \includegraphics[width=0.9\linewidth]{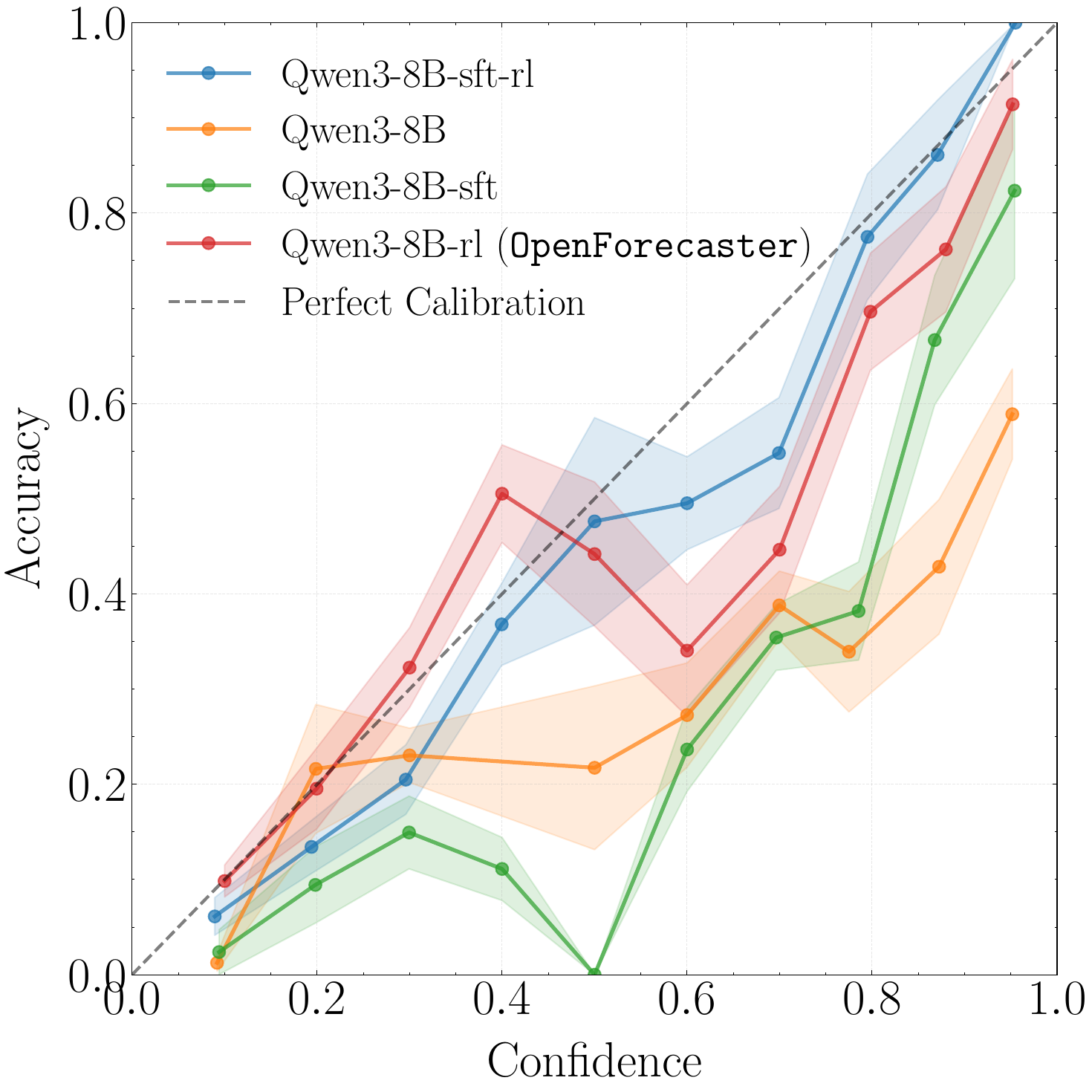}
        \caption{ 
            Calibration curve of the models on our test set.
             }
        \label{fig:sft_calibration}
    \end{subfigure}

    \caption{
        RL training on \ourdataset \  improves the SFT models on accuracy. In particular, the calibration of Qwen-8B-sft-rl model (blue line) is near perfect. 
    }
    \vspace{-.2cm}
\end{figure}

Finally, we train this distilled checkpoint with GRPO using same data mixture, reward design and configurations as we used in training \ourmodel-8B. We report the results for both just distillation (Qwen3-8B-sft) and RL on the SFT checkpoint (as Qwen3-8B-sft-rl) in \Cref{fig:sft_test_set_performance} on our curated test set from May to August 2025. The final model achieves higher accuracy and much better calibration albeit with a slightly lower brier score. 




\subsection{Consistency Evaluation}
\label{sec:consistency}
\begin{table}[htbp]
\centering
\caption{\textbf{Consistency checks before and after RL training}. We report average violation scores and relative changes (negative percentages indicate improvements, positive indicate regressions). The RL-trained model shows improvements in most areas.}
\label{tab:consistency_comparison}
\resizebox{\textwidth}{!}{%
\begin{tabular}{l|ccc|ccc}
\toprule
& \multicolumn{3}{c|}{\textbf{Arbitrage}} & \multicolumn{3}{c}{\textbf{Frequentist}} \\
\cmidrule(lr){2-4} \cmidrule(lr){5-7}
\textbf{Check} & Qwen3-8B & \ourmodel-8B & \textbf{$\Delta$} & Qwen3-8B &  \ourmodel-8B & \textbf{$\Delta$} \\
\midrule
\textsc{Paraphrase} & 0.030 & 0.020 & -33\% & 0.157 & 0.131 & -17\% \\
\textsc{Consequence} & 0.010 & 0.003 & -67\% & 0.048 & 0.029 & -39\% \\
\textsc{AndOr} & 0.033 & 0.025 & -25\% & 0.205 & 0.187 & -9\% \\
\textsc{And} & 0.016 & 0.004 & -75\% & 0.063 & 0.037 & -42\% \\
\textsc{Negation} & 0.043 & 0.063 & +46\% & 0.198 & 0.271 & +37\% \\
\textsc{Or} & 0.022 & 0.019 & -12\% & 0.094 & 0.120 & +28\% \\
\textsc{But} & 0.040 & 0.027 & -31\% & 0.234 & 0.202 & -14\% \\
\textsc{Cond} & 0.039 & 0.033 & -15\% & 0.227 & 0.222 & -2\% \\
\textsc{CondCond} & 0.036 & 0.035 & -3\% & 0.256 & 0.258 & +1\% \\
\textsc{ExpEvidence} & 0.041 & 0.034 & -18\% & 0.240 & 0.195 & -19\% \\
\midrule
\textbf{Aggregated} & \textbf{0.031} & \textbf{0.026} & \textbf{-15\%} & \textbf{0.172} & \textbf{0.165} & \textbf{-4\%} \\
\bottomrule
\end{tabular}%
}
\end{table} 

\citet{paleka2025consistency} release a dataset of long-term forecasting questinos set to resolve up to 2028, showing language models exhibit inconsistencies in their probabilistic predictions. To evaluate consistency, they propose ten consistency checks measuring both arbitrage and frequentist violations. %

We evaluate Qwen3-8B and our trained model on the dataset created by \citet{paleka2025consistency}. We measure performance of the models on all consistency check tuples proposed by them. Table~\ref{tab:consistency_comparison} compares the baseline Qwen3-8B with our RL-trained model. The results show improvements across most consistency checks. We observe strong gains in Boolean logic operations (AND: 75\% reduction in arbitrage violations, 42\% in frequentist) and consequence checks (67\% and 39\% reductions respectively) but also some regressions, particularly in negation consistency. Overall, our training achieves a 15\% reduction in arbitrage violations and 4\% reduction in frequentist violations.

\vspace{1cm}
\begin{figure}[h]
    \centering
    
    \begin{minipage}[t]{0.8\textwidth}
        \centering
        \includegraphics[width=0.95\linewidth]{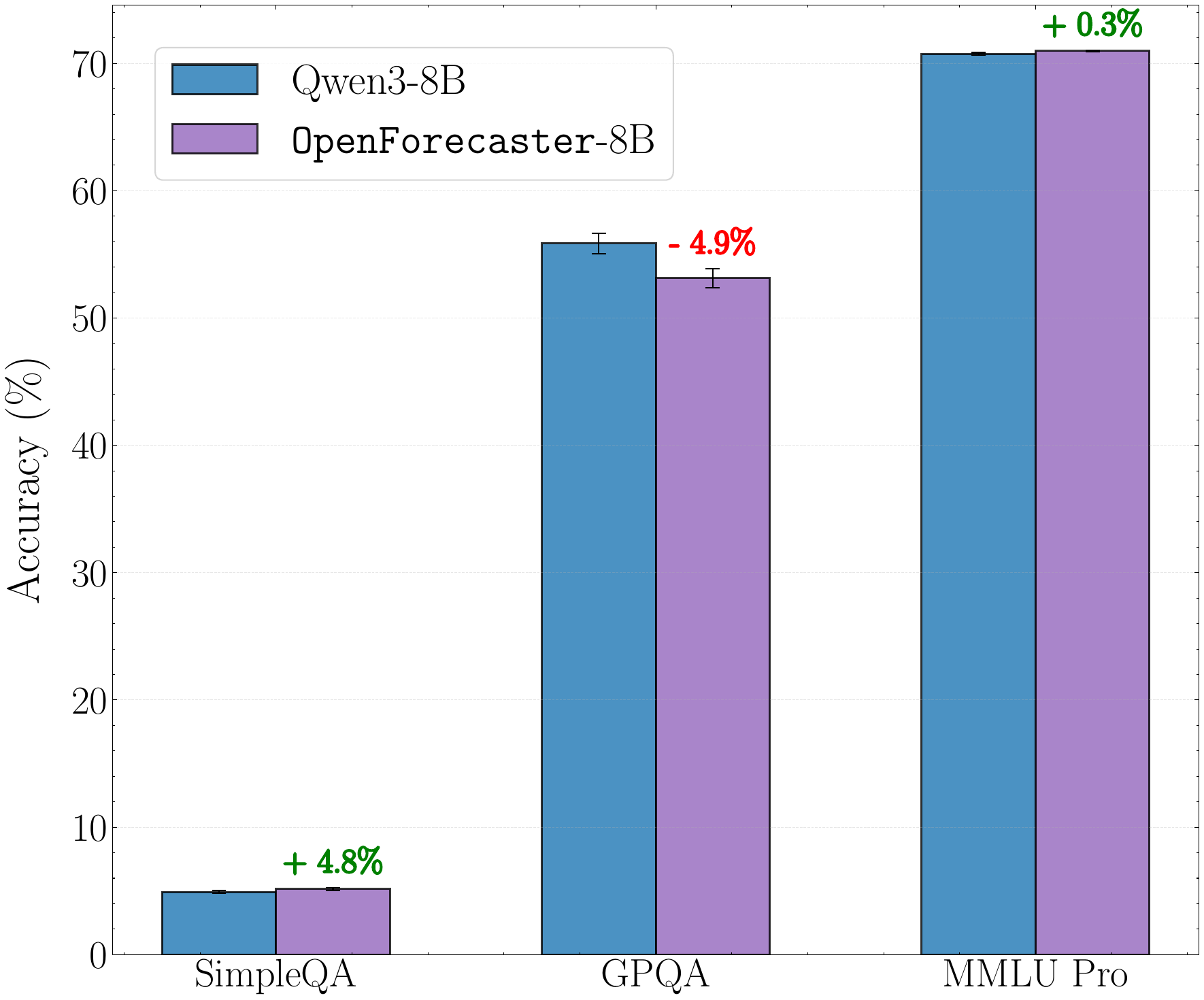}
        \caption{Accuracy comparison on general benchmarks. The number highlighted in green (red) shows \emph{relative} improvement (degradation) over the pre-RL model.}
        
    \end{minipage}
    \label{fig:accuracy_otherbenchmarks}
\end{figure}


\subsection{Evaluation on Metaculus Questions}
\label{sec:metaculus_results}

\begin{wrapfigure}[21]{r}{0.5\textwidth} 
    \centering
    \includegraphics[width=\linewidth]{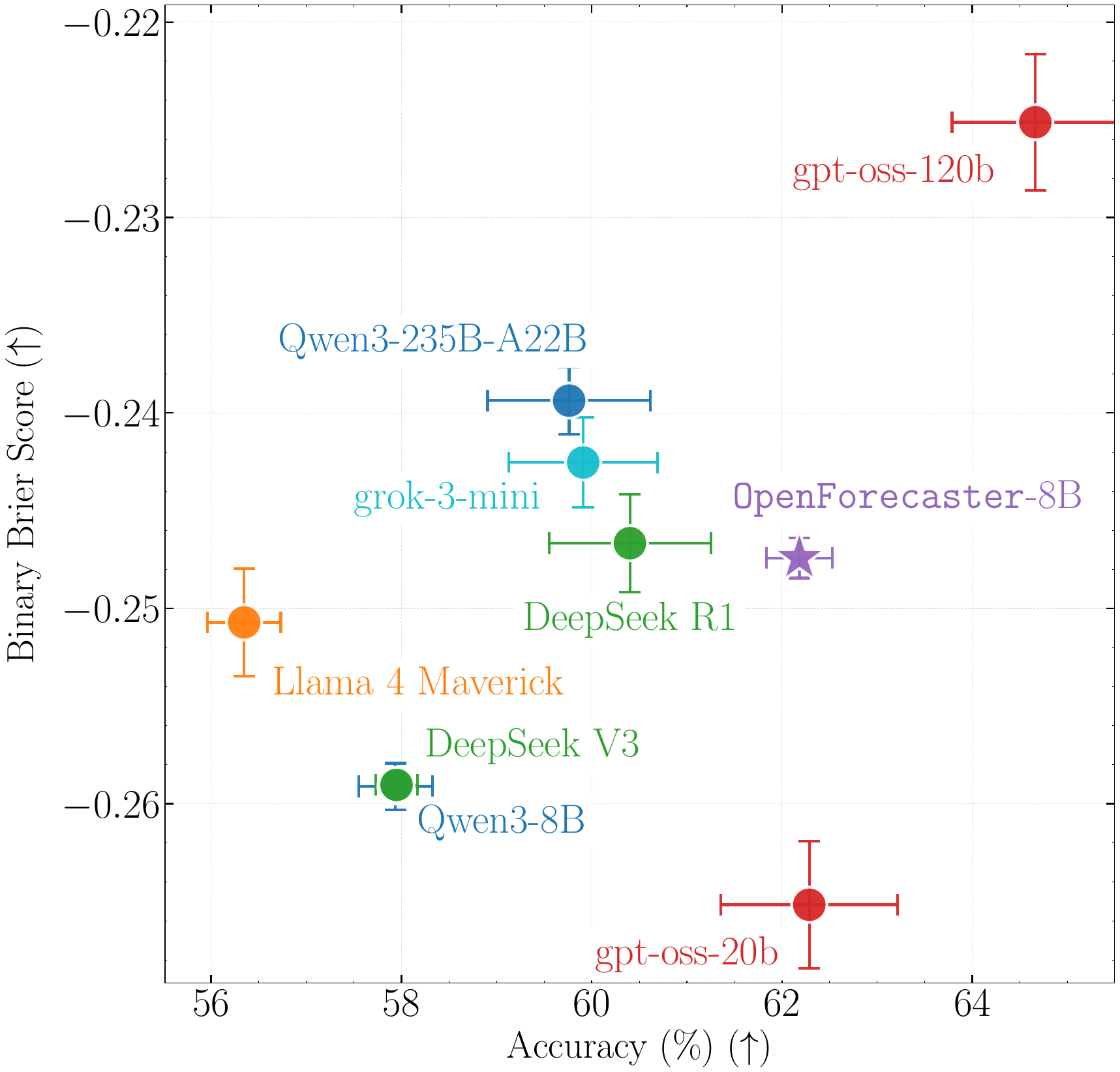}
    \caption{Performance of models on Metaculus questions from May 01 to Nov. 18 2025. 
    }
    \label{fig:metaculus_performance}
    \vspace{-.2cm} 
\end{wrapfigure}

\paragraph{Prediction Market Dataset.} We also source questions from the Metaculus prediction market from May - Nov. 2025. We filter questions which are either meta-prediction questions (prediction about some other prediction question on Metaculus), stock price prediction or below a certain trading volume leaving 449 high-interest questions. Each question is binary so a naive baseline gets 50\% accuracy. We evaluate models on this benchmark by computing both accuracy and brier score.

\paragraph{Results on Real Market Questions.} As shown in \Cref{fig:metaculus_performance}, \ourmodel \ 8B performs better than many larger models like DeepSeek-R1 and Llama-4-Maverick while GPT-OSS-120B is significantly better than rest of the models. \textbf{Note:} The brier score reported here is the \emph{standard brier score} used in prediction markets ranging from $-1$ to $0$ (with -0.25 being the baseline for a constant prediction of $50\%$) and this is different from the brier score we have reported earlier for freeform questions.

%% file: sections/22_train_details.tex
\section{Experimental Setup}

\subsection{Training Details}

\label{sec:setupdetails}

\textbf{Models.} We primarily train and perform all our ablations on Qwen3-8B~\citep{yang2025qwen3technicalreport} thinking model. No official knowledge–cutoff date for Qwen-3 is reported. When queried directly, the models returns inconsistent cutoff dates (most often \emph{October 2023} or \emph{June 2024}). 
Although the family was released in April 2025, the models frequently treat events from 2024 as future,  suggesting a practical cutoff date. This behavior is acceptable for training—even when some prompts refer to events that lie in the model’s past. We also train models for Llama and Gemma family to understand how much post-training on \ourdataset \ helps models which haven't undergone RL training already.  

\textbf{Framework.} We perform RL training using the \texttt{VeRL} package with GRPO algorithm \citep{shao2024deepseekmath} for optimization.

\textbf{Policy/backbone.} Unless noted, the trainable policy is \texttt{Qwen3-8B}. Prompts (with retrieval of up to 5 chunks) are truncated to $4{,}096$ tokens and responses are capped at $8{,}192$ tokens.

\textbf{Sampling.} We generate with a vLLM-based sampler (chunked prefill enabled). Training uses temperature $1.0$ with $K{=}8$ samples per prompt. 

\textbf{Optimization.} We use AdamW \citep{loshchilov2017fixing} with learning rate $5{\times}10^{-6}$, cosine decay, $1\%$ warmup, and a minimum LR ratio of $0.1$. FSDP parameter \emph{and} optimizer offloading are enabled; gradient checkpointing, padding removal, and dynamic batch sizing are used. Global train batch size is $256$ (PPO mini-batch $64$). Training runs are performed a node of $8$ H100 GPUs for $5$ epochs.

\textbf{Data and Training Order.} We train on all 52K samples from \ourdataset \ and also include additional 2K resolved binary questions sourced from Metaculus (essential for performing well on binary questions as we show in \Cref{sec:binary_ablations}). Importantly, we do not shuffle all the 54K samples randomly as that would lead to less than $10$ binary samples in each batch (of size 256) on average. We instead create two separate groups for freeform and binary questions (randomly shuffling within each group). We order our training set such that first 52K questions are free-form and last 2K being binary. 

\emph{{Note:} We found having separate batches of binary and freeform questions is {crucial} to get good performance on binary questions like those found in prediction markets.}

\textbf{Advantages and losses.} GRPO with group-centered advantages (no standard-deviation normalization). PPO clipping uses $\epsilon_{\text{low}}{=}0.20$, $\epsilon_{\text{high}}{=}0.28$, and clip-$c{=}10.0$. We apply a low-variance KL penalty with coefficient $\beta{=}0.005$.

\textbf{Rewards.} We use the \texttt{Qwen3-4B} with greedy decoding (temperature=0) as the judge for assessing answer correctness. We instruct it to enforce strict, reference-guided matching with tolerance for case and common aliases, and prompt it in \texttt{non-thinking} mode. 

For freeform questions, the reward function is the sum of accuracy (0 to 1) and brier score (-1 to 1). For binary questions, the reward is just the (negated) binary brier score ($-1$ to $0$ with $-0.25$ being the baseline for a prediction of $50\%$). We also include a format reward penalty of $-1$ if the answer and/or probability values cannot be extracted or parsed succesfully and $0$ otherwise. Thus, the final (total) reward range on freeform questions is $[-2, 2]$ and on binary questions is $[-2, 0]$.

\subsection{Evaluation Details}

For all our main results, we sample $n=3$ response per prompt for each model with temperature $0.6$ and top-$p=0.95$. 

\textbf{Freeform Evaluation.} To ensure that our model is not exploiting any LLM-as-a-judge biases in training, we employ \texttt{Llama-4-Scout} (instead of \texttt{Qwen3-4B} judge used in training) with greedy decoding (to minimize variance) for checking model's responses against the ground truth using answer matching.

\textbf{Binary Evaluation.} For binary questions, we ask the model to report its best probability estimate $p$ for the event to resolve \texttt{YES}. Let $o$ be the actual outcome of the event (1 if \texttt{YES} and 0 otherwise). The brier score is computed as $-(p - o)^2$ with $0$ being the best and $-1$ being the worst. 

\subsection{Details on Compute}

To improve transparency around data and compute, we report approximate token counts, training steps, and GPU-hours for both SFT and RL. Our curated \ourdataset training set contains 52,183 samples. The average sample (just the question without any retrieval) has about 1000 characters and corresponds to roughly 400 tokens under the Qwen3 tokenizer for the question text, yielding approximately \(2\times 10^{7}\) prompt tokens in total.

For SFT, fine-tuning \texttt{Qwen3-8B} for 3 epochs took 5 hours on 8 \textsc{H100} GPUs, corresponding to roughly 40 \textsc{H100} GPU-hours. RL training is substantially more expensive: Our final run lasted for 5 epochs over the training set, resulting in about 1{,}300 optimization steps, for an estimated total of \(\sim 1{,}000\) \textsc{H100} GPU-hours. Including all ablations, we estimate we used \(\sim 20{,}000\) \textsc{H100} GPU-hours.

\clearpage

%% file: sections/13_extra_exps.tex
\clearpage
\section{Analyzing the \ourdataset \ Dataset}
\label{sec:dataset_details}

\begin{wraptable}[12]{r}{0.5\textwidth}
\vspace{-0.4cm}
\caption{Breakdown of source news articles by news outlet.}
\vspace{-0.25cm}
\resizebox{\linewidth}{!}{
\begin{tabular}{lc}
\toprule
\textbf{Source} & \textbf{Articles (\%Total)} \\
\midrule
Forbes & 110,103 (44.3\%) \\
The Hindustan Times & \ 80,000 \ (32.2\%) \\ 
The Irish Times & \ 29,546 \  (11.9\%) \\ 
Deutsche Welle (DW) & \ 21,317 \ (8.6\%) \\ 
Cable News Net (CNN) & \ \ 7,355 \ \  (3.0\%) \\ \midrule
\textbf{Total} & \ \textbf{248,321 (100\%)} \\
\bottomrule
\end{tabular}}
\vspace{-0.3cm}
\label{tab:dist}
\end{wraptable}
\textbf{News Corpus Details.} We drew from a heterogeneous pool of news articles diversified by geography, time, and topic to ensure broad coverage. We collected a English-language articles from outlets including Forbes, Hindustan Times, The Irish Times, Deutsche Welle, and CNN. They were selected by first collecting a corpus of approximately 250,000 articles spanning from June 2023 to April 2025, encompassing major large-scale events across sports, geopolitics, local news, crime, entertainment, and the arts. Then, performing de-duplication and filtering for language, text availability, and valid dates, we retained approximately 248,000 articles for the question generation phase. Table \ref{tab:dist} details the distribution across news outlets for our corpus.

The overall dataset creation process costed us ~3000\$ with training set costing ~2200\$ (using DeepSeek-v3) while creating the test set costed ~800\$ (using o4-mini-high and grok-4.1-fast with search tool).

\vspace{0.15cm}

\subsection{Analysis of Best Question Selection}

\begin{figure}[h]
\begin{minipage}{0.55\textwidth}
\resizebox{\linewidth}{!}{
\centering
    \includegraphics[]{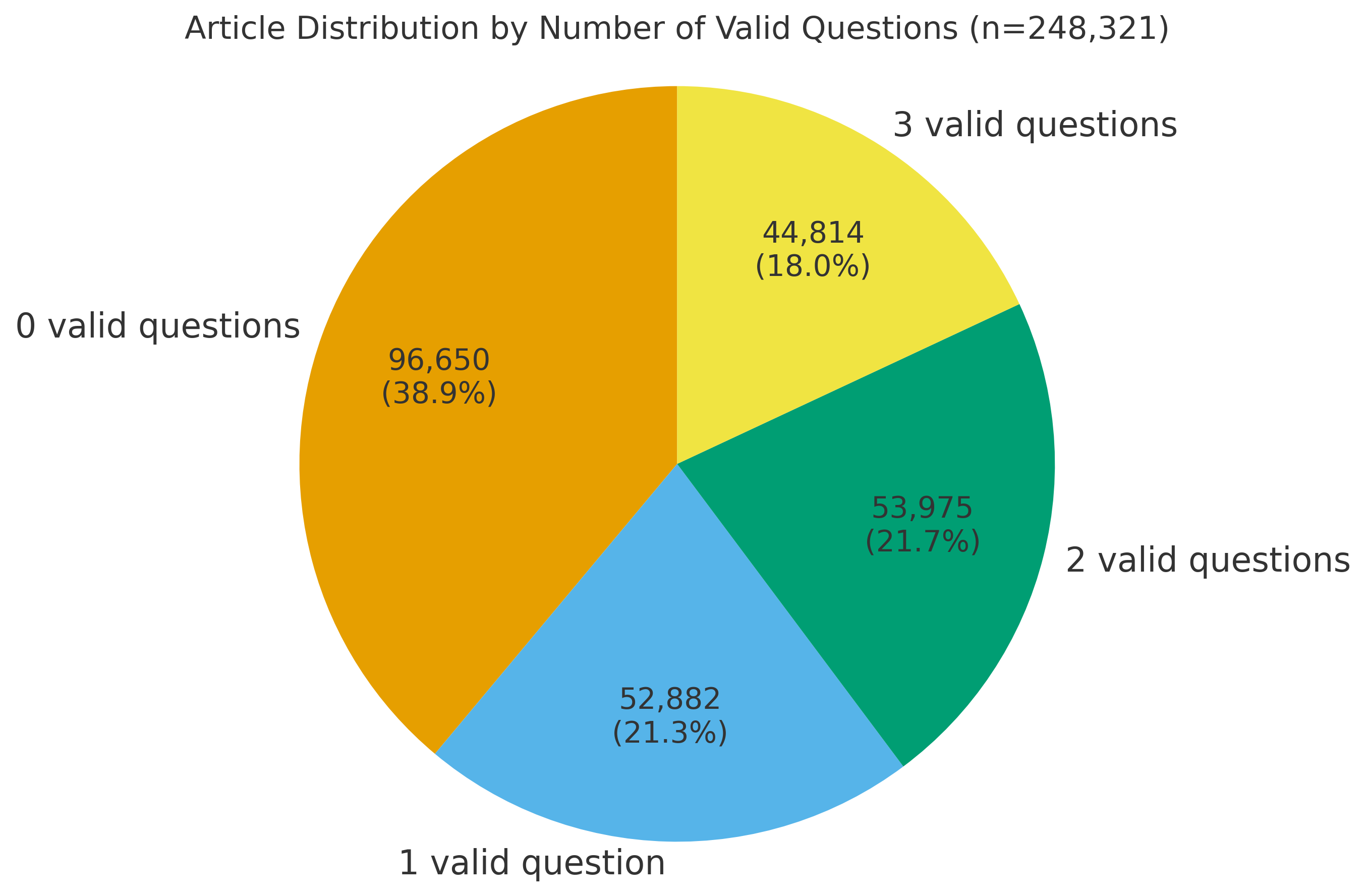}}
    \end{minipage}
    \hfill
\begin{minipage}{0.41\textwidth}
\resizebox{\linewidth}{!}{
\centering
    \includegraphics[]{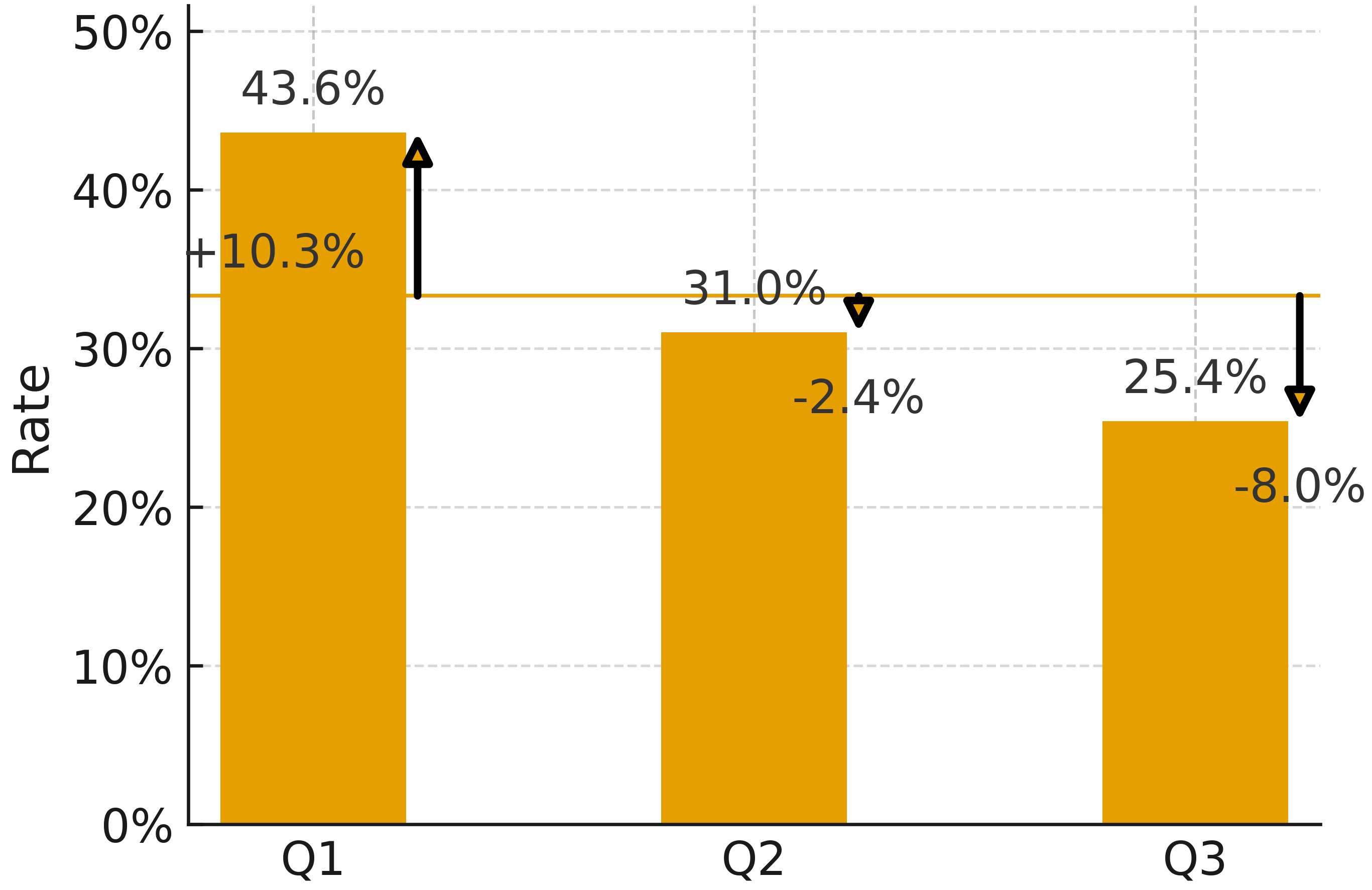}}
    \end{minipage}
 \caption{\textbf{Data Distribution of Questions in} \ourdataset. (Left) Distribution of number of questions selected after filtering from articles (Right) We show the number of questions generated, and the proportion of the first, second and third generated question being picked as the final ``best question''.}
    \label{fig:databreakdown}
\end{figure}

As Figure \ref{fig:databreakdown} (left) illustrates, the generation process yielded mixed results. Post-processing, 39\% of source articles failed to produce any valid questions. Among the surviving articles, 21\% yielded exactly one valid question, which we retained. For the 61\% of articles producing multiple valid questions, we employed another LLM to identify the best candidate based on global relevance, specificity, and unambiguity. Our analysis of selected questions, shown in Figure \ref{fig:databreakdown} (right) reveals a selection bias:
\begin{itemize}
    \item Question 1: Selected 43.6\% of the time (10.3\% above random).
    \item Question 2: Selected 31\% of the time (2.4\% below random).
    \item Question 3: Selected 26.7\% of the time (6.7\% below random).
\end{itemize}
This trend suggests that initial generation attempts (question positioned earlier) frequently produce higher-quality results. Figure \ref{tab:succinct_forecasting_questions_v2} provides qualitative examples of these generated questions.

\begin{table}[h]
\centering
\resizebox{0.9\textwidth}{!}{%
\begin{tabular}{l*{10}{c}}
\toprule
 & \textbf{Name(s)} & \textbf{Location} & \textbf{Country} & \textbf{Title} & \textbf{Team} & \textbf{Color} & \textbf{Organization} & \textbf{Currency} & \textbf{Brand} & \textbf{Month} \\
 & & & & & \textbf{name} & & & & \textbf{name} & \\
\midrule
\textbf{Count} & 32{,}213 & 14{,}337 & 2{,}579 & 2{,}479 & 1{,}445 & 1{,}047 & 1{,}030 & 877 & 779 & 730 \\
\textbf{Share} & 44.8\% & 20.0\% & 3.6\% & 3.5\% & 2.0\% & 1.5\% & 1.4\% & 1.2\% & 1.1\% & 1.0\% \\
\bottomrule
\end{tabular}%
}
\caption{Top ten answer types of the questions in our curated dataset. These ten categories cover 80.1\% of our training dataset.}
\label{tab:answer_type_top10_horizontal}
\end{table}

\subsection{Distribution of Answer Types}
Table \ref{tab:answer_type_top10_horizontal} categorizes the answer types within the training data. Two categories dominate the dataset:
\begin{itemize}
    \item People and Places (65\%): Names of individuals constitute nearly 45\% of the answers, while locations account for 20\%.
    \item Miscellaneous Entities (35\%): The remainder consists of teams, countries, organizations, colors, and similar entities.
\end{itemize}

\begin{table}[h]
\centering
\small
\setlength{\tabcolsep}{4pt}
\renewcommand{\arraystretch}{1.2}
\begin{tabularx}{\textwidth}{L L L L L L}
\toprule
\textbf{Question} & \textbf{Background} & \textbf{Resolution (trigger \& deadline)} & \textbf{Answer Type} & \textbf{Answer} & \textbf{Source} \\
\midrule
Host country of COP30 (Nov 2025)? & UNFCCC COP venue rotates among regions. & Host confirmed by UNFCCC/organizers; no later than COP30 start (Nov 2025). & string (country) & Brazil & DW:\newline \href{https://www.dw.com/en/donald-trump-climate-policy-2025/a-71354256?maca=en-rss-en-all-1573-xml-atom}{link} \\
\midrule
Release month of Marvel's \emph{Fantastic Four} (2025)? & Reboot announced with lead cast; 2025 release slated. & Month confirmed by Marvel/Disney; by Dec 2025. & string (month) & July & Forbes:\newline \href{https://www.forbes.com/sites/caileygleeson/2024/02/16/x-men-fans-push-back-against-anger-over-non-binary-character/}{link} \\
\midrule
First state to require Ten Commandments in public classrooms (by 2025)? & Several U.S. states advance religion-in-school measures. & First state enacts requirement; by Dec 31, 2025. & string (state name) & Louisiana & Forbes:\newline \href{https://www.forbes.com/sites/maryroeloffs/2024/06/20/louisianas-new-ten-commandments-law-is-latest-example-of-states-pushing-to-allow-religion-in-public-schools/}{link} \\
\midrule
African host of G20 Summit (Nov 2025)? & G20 presidency rotates; South Africa presiding from Dec 2024. & G20/host government confirms location; by Nov 2025. & string (country) & South Africa & DW:\newline \href{https://www.dw.com/en/africa-in-2025-strengthening-democracy-key-to-growth/a-70945864}{link} \\
\midrule
Recipient of Lesotho--Botswana Transfer Scheme (by 2025)? & Regional pipeline to pump water from Lesotho via SA. & ORASECOM or governments confirm recipient; by 2025. & string (country name) & Botswana & DW:\newline \href{https://www.dw.com/en/lesotho-urged-to-compensate-villagers-for-water-projects/a-69352942}{link} \\
\bottomrule
\end{tabularx}
\caption{Five succinct forecasting questions spanning climate, entertainment, law, geopolitics, and infrastructure; selected for brevity and diverse sources (DW, Forbes). Each row lists the question (summarized here for conciseness), short background, resolution trigger with deadline, answer type, ground-truth answer, and citation.}
\label{tab:succinct_forecasting_questions_v2}
\end{table}

\subsection{Test Set}
\label{sec:test_set_details}

\begin{figure}[b]
\vspace{-0.05cm}
    \centering

    \begin{subfigure}[t]{0.49\textwidth}
        \centering
        \includegraphics[width=0.95\linewidth]{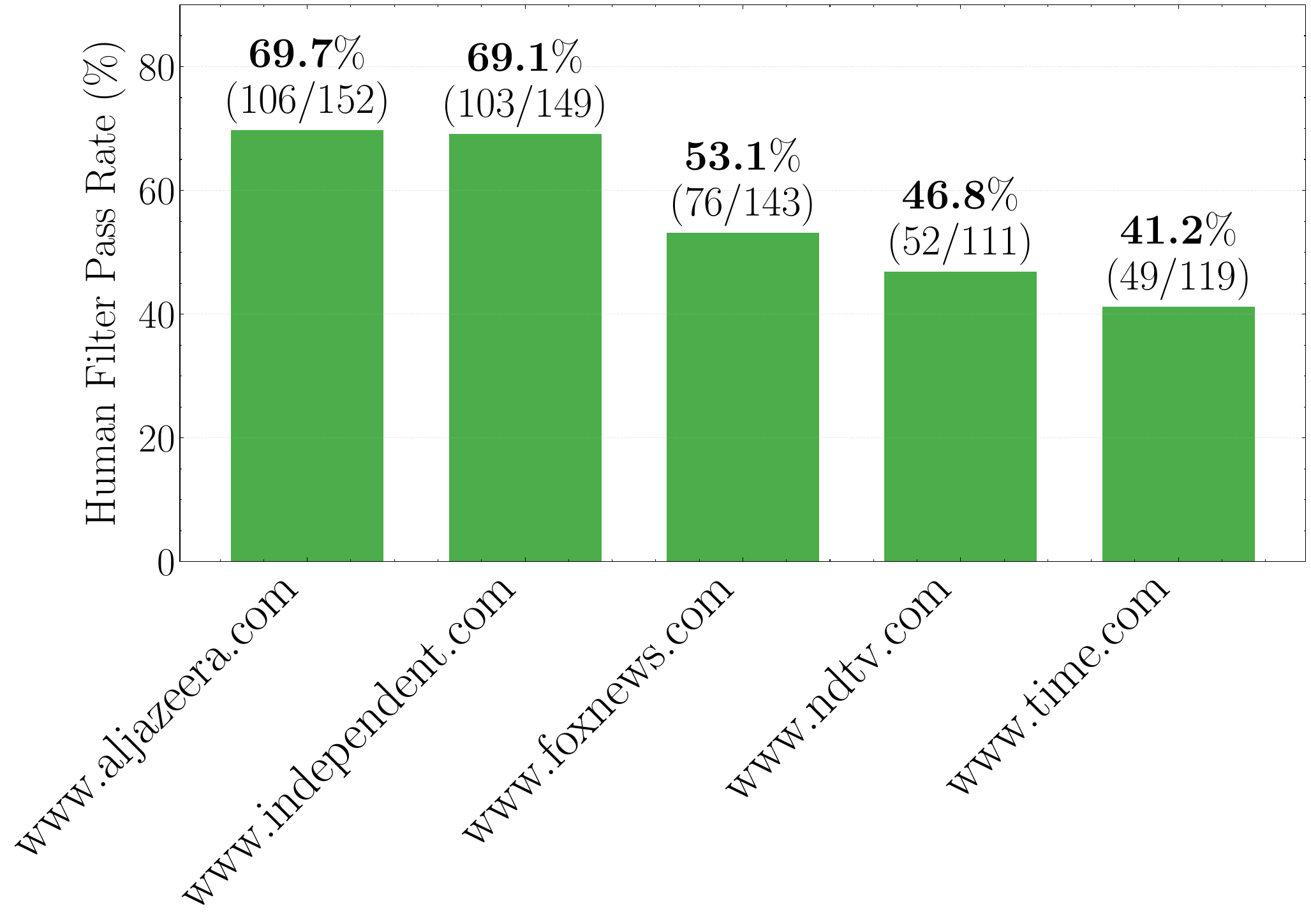}
        \caption{Manual filter pass rate per news source.}
        \label{fig:manual_filter_rate}
    \end{subfigure}
    \hfill
    \begin{subfigure}[t]{0.49\textwidth}
        \centering
        \includegraphics[width=0.95\linewidth]{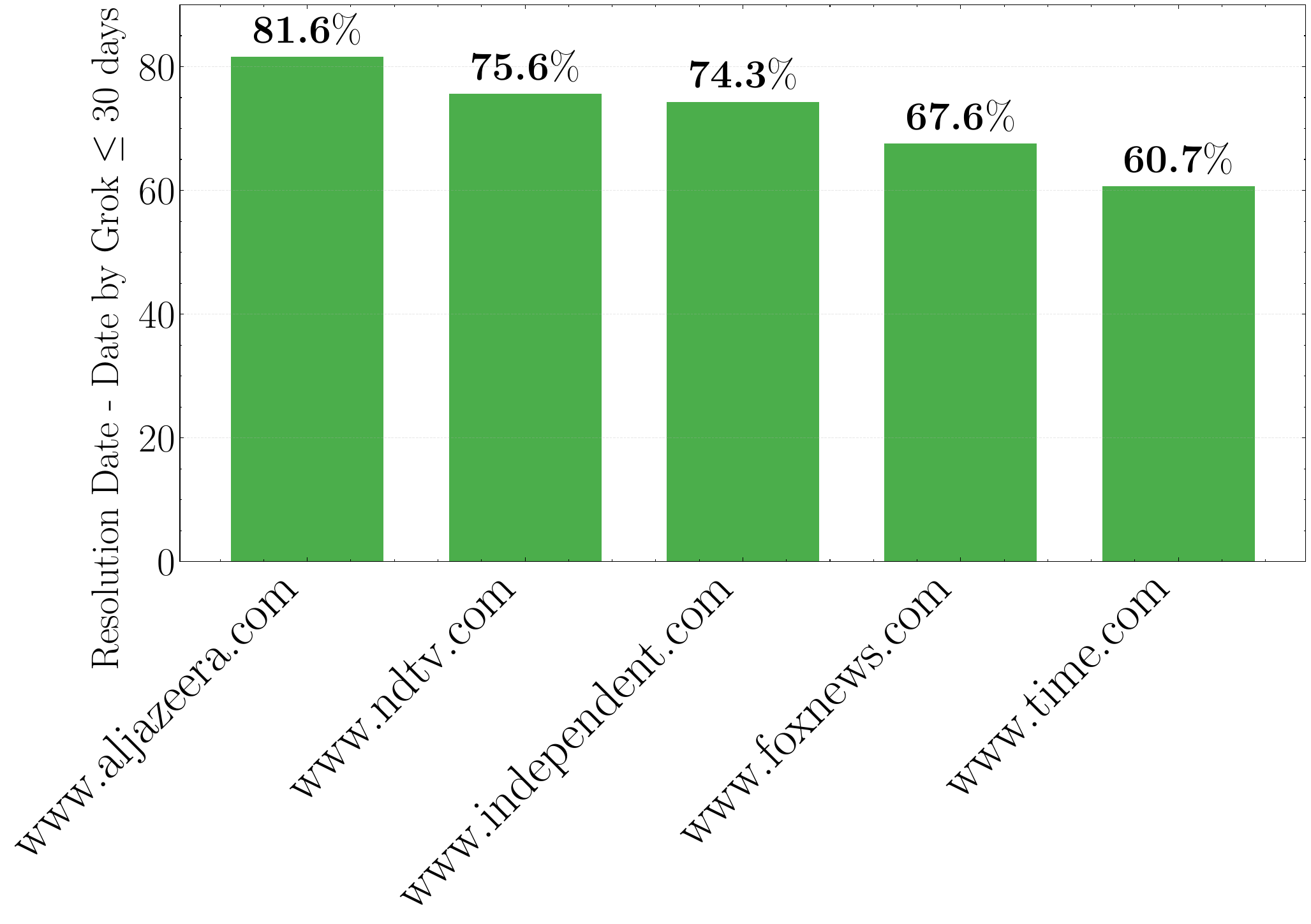}
        \caption{Proportion of questions whose resolution date is within 1 month of the resolution date found by grok-4.1-fast.}
        \label{fig:correct_resolution_date}
    \end{subfigure}

    \caption{
        Filtering pass rate of forecasting questions across news sources.
        }
    
    \vspace{-.4cm}
\end{figure}

For the test set, we first prepare an initial set of 1000 questions generated using \texttt{o4-mini-high} model as described in \Cref{sec:results}. We next performed additional filtering steps to retain high-quality future-facing questions: 
\begin{enumerate}
    \item We removed any potentially unanswerable questions (noise) by keeping only those which \texttt{grok-4.1-fast} could successfully answer majority of the times (run repeatedly, $n=5$) with search tool access. This filtered out $15\%$ of the questions.
    \item To address the issue of late reporting in news outlets, we again use \texttt{grok-4.1-fast} with search tool to find the \textbf{earliest resolution date} for a given question. This is important to prevent leakage from retrieving articles with the true answer. We retain only those questions with resolution date after May 2025. In \Cref{fig:correct_resolution_date}, we report the number of questions per news source for whom the generated resolution date was within 1 month period of the date found by \texttt{grok-4.1-fast}. We notice an average of $70\%$ questions have resolution date even within the 1 month period. 
\end{enumerate} 

Finally, we manually filter the remaining questions to meet our quality checks like: 
\begin{enumerate}
    \item Question may have multiple possible correct answers. 
    \item Question actually resolves in the future (after September 2025) for which the article reports the scheduled/planned place/event/etc. 
    \item Question being irrelevant because it is too niche to a certain place or locality.
    \item Question is about something which is already established (known).
\end{enumerate}

We provide the full guidelines we followed for manual filtering in the box below. This resulted in a final test set of $302$ questions which we use for reporting the results.

\begin{tcblisting}{
  title={\textbf{Manual Filtering Guidelines}},
  colframe=black!70,
  colback=black!2,
  boxrule=0.5pt,
  breakable,
  listing only
}
**Task:** You will be given (i) a news article and (ii) a question that is supposed to be answerable from that article. Your job is to decide whether to KEEP the question in the test set or REJECT it during manual filtering.

**How to review:**
1. Read the full question (Title, Background, Resolution Criteria, and the proposed Answer).
2. Apply the rejection criteria below. If **any** criterion triggers, REJECT the question. Refer to the article text if required.

------------------------------------------------------------
**REJECT the question if it fails ANY of the following criteria:**

1) **Multiple possible correct answers (Ambiguous / not uniquely resolvable)**
   - Reject if the question could reasonably have more than one correct answer, even after reading the article.
   - Examples:
     - "Which company invested in ABC in July 2025?" when multiple companies are mentioned investing in ABC in July 2025.
     - "Who announced a new partnership?" when the article lists several partnerships.

2) **Resolves in the future (after September 2025)**
   - Reject if the question's true resolution depends on an event that happens **after 2025-09-30**, and the article only reports a *plan/schedule/announcement* rather than the event actually occurring.
   - Examples:
     - Article: "The ABC event is scheduled for Jan 2026 in PQR venue."
       Question: "Where will event ABC be held in Jan 2026?" -> Reject (the question resolves only after the event actually takes place as the venue can change last minute).

3) **Too niche / overly local / low general relevance**
   - Reject if the question is narrowly tied to a very specific locality, institution, or small community such that it is not broadly meaningful for a general test set.
   - Examples:
     - A minor municipal policy affecting only one small village.
     - A local event with no wider regional/national/global significance.

4) **Already established (Known / not a meaningful forecast target)**
   - Reject if the question is about something that is already settled or widely known at the time (i.e., not an uncertain outcome).
   - Examples:
     - "Who stepped down from the role of CEO of Company ABC in August 2025?" (requires knowing the current CEO which generally doesn't change frequently and is often stable over multiple years)
     - Long-established facts, definitions, or historical constants.

------------------------------------------------------------
**Decision format (for annotation):**
- If any criterion triggers, then **REJECT**.
- Otherwise, **KEEP**.

**Question:**
{questions_text}

<toggle> **Source Article:** {source_article} </toggle>
\end{tcblisting}

\clearpage

%% file: sections/19_qualitative.tex
\clearpage
\section{Qualitative Analysis}
\label{sec:qualitative_analysis}

\subsection{Qualitative Analysis of Final Answers}
\label{sec:qualia}

We manually annotated responses to 207 questions by both the initial Qwen3-8B thinking model and the trained \ourmodel  8B on the Guardian validation set. Using this set, we found that the agreement between the two models used for grading, Llama 4 Scout and Qwen3 4B is $\sim 97\%$, and we agree with their grading in over 95\% of cases. This confirms the reliability of automatic answer matching based evaluation.

In \Cref{tab:domain}, we analyze the domains (by news section) in which our trained model improves. We find significant improvements in the World, Australian, and US news sections, with no significant change for sports. This suggests our model may not yet perform well on sports-heavy prediction markets like Kalshi.

\begin{table}[h]
\centering
\begin{tabular}{lrrrr}
\hline
Domain & $n$ & Before & After & $\Delta$ \\ 
\hline
world & 20 & 21.7 & 33.3 & +11.6 \\
australia-news & 15 & 35.6 & 42.2 & +6.7 \\
us-news & 21 & 41.3 & 44.4 & +3.2 \\
sport & 37 & 43.2 & 43.2 & +0.0 \\
football & 30 & 34.4 & 33.3 & -1.1 \\
\hline
\end{tabular}
\caption{Avg@3 by domain (n $\ge$ 10).}
\label{tab:domain}
\end{table}

In \Cref{tab:questionform}, we analyze change in performance by question type, finding significant improvements on questions of the form ``what'', ``which'', and ``who'', while a slight regression in performance on location questions (``where'').

\begin{table}[h]
\centering
\begin{tabular}{lrrrr}
\hline
Question form & $n$ & Before & After & $\Delta$ \\ 
\hline
what & 25 & 14.7 & 29.3 & +14.7 \\
which & 98 & 45.2 & 51.4 & +6.1 \\
who & 60 & 27.8 & 33.9 & +6.1 \\
other & 10 & 40.0 & 43.3 & +3.3 \\
where & 14 & 47.6 & 45.2 & -2.4 \\
\hline
\end{tabular}
\caption{Avg@3 by question form (n $\ge$ 10).}
\label{tab:questionform}
\end{table}

Below, we present qualitative examples where our training improves and worsens predictions compared to the original model. 

\subsubsection*{Qualitative examples (improved; first sample)}
\begin{itemize}
\item \textbf{Q:} Who will be wearing the yellow jersey in the general classification at the end of stage eight of the 2025 Tour de France? \\
\textbf{Truth:} Tadej Poga{\v{c}}ar \\
\emph{Before:} Jonas Vingegaard (p=0.10) \\
\emph{After:} Tadej Poga{\v{c}}ar (p=0.60)

\item \textbf{Q:} Who will withhold a resolution from the U.S. House floor to force a vote on releasing the Epstein documents by July 25, 2025? \\
\textbf{Truth:} Mike Johnson \\
\emph{Before:} Pam Bondi (p=0.30) \\
\emph{After:} Mike Johnson (p=0.60)
\end{itemize}

\subsubsection*{Qualitative examples (regressed; first sample)}
\begin{itemize}


\item \textbf{Q:} Who will be the Democratic Party's nominee for New York City mayor in the November 2025 general election? \\
\textbf{Truth:} Zohran Mamdani \\
\emph{Before:} Zohran Mamdani (p=0.60) \\
\emph{After:} Andrew Cuomo (p=0.40)

\item \textbf{Q:} Who will post the lowest first-round score among Rory McIlroy, Scottie Scheffler and Viktor Hovland at the 2025 Scottish Open? \\
\textbf{Truth:} Viktor Hovland \\
\emph{Before:} Viktor Hovland (p=0.60) \\
\emph{After:} Scottie Scheffler (p=0.40)
\end{itemize}

\subsection{Reasoning Evolution During Training}
\label{sec:reasoning_evolution}

We now provide a few qualitative examples of how the model reasoning evolves when training on OpenForesight. We ask the model itself to provide a summary of its reasoning (after the <think></think> tags), and show this summary below for the initial Qwen3-8B model, followed by training checkpoints.

\subsubsection{Example 1: Model stays incorrect but learns to hedge}

\noindent\textbf{Question:} Which country will face a new 25\% US tariff in a trade announcement by 30 July 2025?\\
\noindent\textbf{Background:} The US administration is evaluating other countries' trade barriers and is expected to announce new tariffs by the end of July 2025.\\
\noindent\textbf{Resolution Criteria:}

\textbf{Source of Truth}: Official tariff announcement issued by the White House or the Office of the United States Trade Representative.

\textbf{Resolution Date}: 30 July 2025, the date by which the 25\% tariff announcement is expected.

\textbf{Accepted Answer Format}: The full name of the country as stated in the official announcement.

\noindent\textbf{Resolution answer:} India

\medskip

\noindent\textit{Checkpoint 0: score=-0.722, status=wrong}\\
<answer>South Korea</answer> <probability>0.85</probability> **Reasoning**: - **Article 2** (May 25, 2025) explicitly mentions a **25\% tariff** on **Samsung** (South Korea) and other smartphone manufacturers. This directly ties the 25\% tariff to **South Korea**. - While the EU is mentioned with a **50\% tariff** in the same article, the **25\% tariff** specifically targets South Korea's exports. - Other articles reference tariffs on China, the EU, or automakers, but none specify a **25\% tariff** on a country beyond South Korea. - The resolution date (July 30, 2025) aligns with the timeline of the May 25 announcement, as no newer data is provided. **Confidence**: 85\% (strong evidence from Article 2, though no post-May 25 updates confirm further announcements).

\medskip

\noindent\textit{Checkpoint 400: score=-0.360, status=wrong}\\
The retrieved articles indicate ongoing US tariff announcements by the Trump administration in 2025. Article 2 explicitly mentions a 25\% tariff on South Korea for smartphone manufacturers like Samsung, while Article 4 references 25\% tariffs on steel and aluminum imports from Canada and Mexico. Article 3 also notes a 25\% tariff on foreign automakers, which could target countries like South Korea or others, but South Korea is specifically named in Article 2 for a 25\% rate. Given the question's focus on a "new" 25\% tariff announcement by July 30, 2025, and the patterns in the articles, South Korea appears as a likely candidate, as it has already faced such announcements and fits the criteria. However, the question implies a potential new announcement, and other countries like Canada or Mexico could also be possibilities based on Article 4, introducing uncertainty. My confidence is moderate due to the reliance on past announcements and the possibility of new developments by the resolution date. <answer>South Korea</answer> <probability>0.6</probability>

\medskip

\noindent\textit{Checkpoint 800: score=-0.090, status=wrong}\\
The retrieved articles indicate a pattern of US tariffs targeting major trade partners like China, India, the EU, Canada, Mexico, and South Korea. Article 4 specifically mentions a 25\% tariff on steel and aluminum imports, which historically targets countries such as China and those in the EU. Article 2 highlights a 25\% tariff on South Korean products (e.g., Samsung), while Article 1 and Article 3 reference ongoing tariffs on China and other nations. Given the US administration's focus on trade barriers, China is the most frequently mentioned target, making it a likely candidate for a new 25\% tariff announcement by July 2025. However, the articles do not confirm a specific new announcement, and other countries like India are also potential targets based on Article 1, introducing uncertainty. Due to the lack of direct evidence for a July announcement and the possibility of shifts in US policy, my confidence is low. <answer>China</answer> <probability>0.3</probability>

\subsubsection{Example 2: Model goes from incorrect to correct}

\noindent\textbf{Question:} Who will give Harvard one week to produce extensive records for its exchange visitor program review by July 23, 2025? \\
\noindent\textbf{Background:} The U.S. Secretary of State periodically issues directives to universities regarding their exchange visitor program compliance.\\
\noindent\textbf{Resolution Criteria:}

\textbf{Source of Truth}: The official announcement or government statement naming the individual who issued the one-week deadline.

\textbf{Resolution Date}: When the directive is publicly attributed (expected by July 23, 2025).

\textbf{Accepted Answer Format}: Full name, for example "Jane Doe".

\noindent\textbf{Resolution answer:} Marco Rubio

\medskip

\noindent\textit{Checkpoint 0: score=-0.902, status=wrong}\\
The retrieved articles describe actions by the Trump administration's Department of Homeland Security, specifically Secretary Kristi Noem, toward Harvard. In the text, Noem sends Harvard a letter giving them 72 hours to provide information, not one week, so there is a mismatch between the question's ``one week by July 23, 2025'' and the articles' ``72 hours.'' It seems that the July 23 date comes from the resolution criteria and isn't in the articles themselves, and that the 72-hour deadline might be part of a larger process leading up to that date, or that the question wording is slightly off. Despite this discrepancy, all the evidence points to Kristi Noem as the person issuing the directive to Harvard about providing records. Therefore, the answer should be Kristi Noem, and the confidence is high because all the articles point to her as the one who issued the directive to Harvard regarding the records.<answer>Kristi Noem</answer> <probability>0.95</probability>

\medskip

\noindent\textit{Checkpoint 400: score=-0.490, status=wrong}\\
The retrieved articles indicate that Kristi Noem, as the U.S. Department of Homeland Security (DHS) Secretary, has been directly involved in issuing deadlines related to Harvard's exchange visitor program compliance. For instance, in Article 2, she gave Harvard 72 hours (approximately 3 days) to comply with requests for records. The question specifies a one-week deadline by July 23, 2025, which may be a variation or continuation of similar directives from this administration. Given the pattern in the articles, it is reasonable to infer that Noem or a similar high-level official from the Trump administration is likely to issue such a directive. However, there is uncertainty due to the slight discrepancy between 72 hours and one week, and potential changes in personnel or administration by July 2025. Final Answer: <answer> Kristi Noem </answer> <probability> 0.7 </probability>

\medskip

\noindent\textit{Checkpoint 800: score=0.910, status=correct}\\
The retrieved articles primarily discuss the U.S. Department of Homeland Security (DHS) Secretary, Kristi Noem, in the context of Harvard's visa program issues, such as scrutiny and potential termination due to non-compliance. However, the question specifically pertains to the U.S. Secretary of State issuing a directive, as outlined in the question background. Based on the Trump administration's context in the articles, Marco Rubio was the Secretary of State in 2024 and likely remained in that role through at least 2025, as no other name is explicitly mentioned for that position. This is inferred from the articles' references to Trump administration actions, though the articles confuse DHS Secretary Noem with Secretary of State responsibilities. Given the resolution criteria requiring the official announcement, I am considering the most probable candidate from historical context, but the articles' inconsistencies lower my confidence. Final Answer: <answer>Marco Rubio</answer> <probability>0.7</probability>

\subsubsection{Example 3: Model goes from correct to incorrect, but interestingly reasons about Brier}

\noindent\textbf{Question:} Which Australian state besides New South Wales will be affected by the forecast 'second surge' of rain and wind on Wednesday night?\\
\noindent\textbf{Background:} The Bureau of Meteorology has forecast a 'second surge' of rain and wind for Wednesday night, expected to impact coastal regions.\\
\noindent\textbf{Resolution Criteria:}

\textbf{Source of Truth}: The question will resolve based on the Bureau of Meteorology's severe weather warnings and regional impact statements for the second surge, as published on bom.gov.au.

\textbf{Resolution Date}: 2 July 2025, following the Wednesday night updates.

\textbf{Accepted Answer Format}: The name of the Australian state, in 1–2 words.

\noindent\textbf{Resolution answer:} Victoria

\medskip

\noindent\textit{Checkpoint 0: score=0.978, status=correct}\\
The question's background says the resolution is on 2 July 2025, following Wednesday night updates. So the answer should be based on the Bureau's forecast as of that time. But the available articles are up to February 2025. If the 'second surge' is part of the forecast mentioned in Article 2 (from January 2025), then the answer is Victoria. However, the user might have a different scenario. Alternatively, in Article 4, there's a mention of damaging winds in parts of Illawarra, Sydney, and Hunter coast, but again, that's part of NSW. So, based on the available information, the most likely answer is Victoria. But I need to make sure that the 'second surge' refers to the same event. The article mentions a second strong wind coming from the south-west, which would affect the south of the country, including Victoria. Therefore, the answer is Victoria. Confidence: The articles mention Victoria as part of the affected regions. However, since the exact 'second surge' isn't explicitly named in the articles, there's some uncertainty. But based on the context, Victoria is the most likely state besides NSW. </think> <answer>Victoria</answer> <probability>0.85</probability>

\medskip

\noindent\textit{Checkpoint 400: score=0.840, status=correct}\\
The provided articles indicate patterns of weather impacts from the Bureau of Meteorology, particularly in eastern Australia. Article 2 explicitly mentions a weather system affecting eastern New South Wales, with parts of Victoria and south-east Queensland also likely impacted, including strong winds and rain. This aligns with the question's reference to a 'second surge' of rain and wind on Wednesday night. Article 1 and Article 3 further support influences on Victoria and South Australia from similar weather fronts, but Victoria is the most consistently mentioned state besides New South Wales across multiple articles. Given the resolution criteria relying on BoM's warnings, I infer that Victoria is the most probable state based on these patterns. However, the articles do not directly reference the specific forecast for 2 July 2025, introducing uncertainty. Confidence: The articles don't directly match the question, so I'm not 100\% sure. Maybe 0.6, since it's inferred from patterns. To maximize my score with Brier score: If I'm correct, higher p is better; if wrong, lower p is better. I need to be honest about my confidence. I think 0.6 is reasonable, as it's a guess based on the info. <answer>Victoria</answer> <probability>0.6</probability>

\medskip

\noindent\textit{Checkpoint 800: score=-0.360, status=wrong}\\
The retrieved articles provide information on weather patterns in Australia, particularly from 2023 and early 2025. Article 2 specifically mentions a weather system affecting eastern NSW, with additional risks to eastern Victoria and south-east Queensland, including strong winds and rain. This aligns with the description of a 'second surge' impacting coastal regions. Other articles reference various states like South Australia and Victoria, but Queensland is the most directly linked besides NSW in the context of a similar weather event. Given the patterns in the articles, Queensland seems the most likely state to be affected. However, the articles are from different dates and not explicitly about the 2 July 2025 event, introducing uncertainty. Based on the evidence, I am moderately confident in this guess. I need to maximize my score using Brier scoring. If I'm correct, higher p is better; if wrong, lower p is better. Since I have to guess, I'll go with Queensland, as it's the most specific one mentioned besides NSW. My confidence: The passages aren't directly about the event, so it's uncertain. Maybe 0.4 or 0.5? Let's say 0.6, because it's the most prominent besides NSW in the relevant passages. <answer>Queensland</answer> <probability>0.6</probability>
  \clearpage

%% file: sections/21_failure_modes.tex
\subsection{Systematic Failure Modes in Model Reasoning}

\ourmodel \ 8B achieved an accuracy of 33\% on our test set consisting of 302 questions. Through manual analysis of the failure cases, we identified three systematic failure modes in the model's reasoning process.

\paragraph{1. Missing Information in Retrieved Articles}

The most common failure mode occurs when retrieved articles lack the specific information needed. The model acknowledges this but falls back to general knowledge or educated guesses rather than expressing high uncertainty.

\textbf{Example 1:} ``By what date will the New York City Board of Elections certify the official results of the Democratic mayoral primary?''

\begin{itemize}
\item \textbf{Correct answer:} July 15
\item \textbf{Model answer:} August 15 (probability: 0.2)
\item \textbf{Model's reasoning:} ``The provided news articles discuss the New York City Democratic mayoral primary, including polling data, candidate strategies, and the election date of June 24, 2025. However, they do not specify the certification process or timeline for the New York City Board of Elections. Based on general knowledge of past NYC elections, certification typically occurs 2-6 weeks after the primary to allow for vote counting, ballot challenges, and legal reviews. For instance, in the 2013 primary, certification was completed by July 2. Given the late summer 2025 resolution criteria, a reasonable estimate is mid-August, but without specific historical data from the articles, this is uncertain. I am guessing August 15, 2025, as it aligns with typical procedures, but my confidence is low due to the lack of direct evidence.''
\item \textbf{Analysis:} The model correctly identifies that articles lack the certification date but incorrectly infers from general knowledge. It acknowledges uncertainty (probability 0.2) yet provides a specific date. The inference from ``2-6 weeks after June 24'' to ``mid-August'' is flawed, as July 15 falls within this range.
\end{itemize}

\paragraph{2. Over-reliance on General Knowledge}

When articles lack specific information, the model makes educated guesses based on general knowledge, historical patterns, or analogies rather than acknowledging the answer cannot be determined from the provided context.

\textbf{Example 1:} ``What will be the official name of Donna D'Errico's interactive AI voice service by mid-July 2025?''

\begin{itemize}
\item \textbf{Correct answer:} Call Donna D.
\item \textbf{Model answer:} Donna Voice (probability: 0.1)
\item \textbf{Model's reasoning:} ``The provided articles do not mention Donna D'Errico or her specific AI voice service, making it impossible to draw direct insights. However, patterns from the articles, such as the naming of AI voice services (e.g., Melania Trump's ``Melania: The Audiobook''), suggest that such services often incorporate the individual's name followed by descriptors like ``Voice'' or ``AI.'' Based on this analogy, I am guessing the name might be ``Donna Voice'' or something similar. This is a speculative prediction, as there's no concrete information, leading to low confidence.''
\item \textbf{Analysis:} The model creates a naming pattern from one unrelated example (Melania Trump's service) and applies it incorrectly. It recognizes the speculation (probability 0.1) but the actual name ``Call Donna D.'' follows a different pattern entirely.
\end{itemize}

\textbf{Example 2:} ``What name will the new Pope take upon his election on May 8, 2025?''

\begin{itemize}
\item \textbf{Correct answer:} Leo XIV
\item \textbf{Model answer:} Pius (probability: 0.1)
\item \textbf{Model's reasoning:} ``The provided articles do not specify the name of the next Pope or any potential candidates' chosen names. They focus on Pope Francis's health, the conclave process, and recent appointments of cardinals. Historically, new Popes select names that are traditional, often drawing from saints or biblical figures, but without specific information from the articles or external knowledge, any prediction is highly uncertain. Based on patterns from past Popes (e.g., Francis, Benedict), I am guessing ``Pius'' as a common regnal name, but this is purely speculative and not supported by the articles. Therefore, my confidence in this answer is very low.''
\item \textbf{Analysis:} The model relies entirely on general knowledge about papal naming conventions rather than specific information. It correctly identifies high uncertainty (probability 0.1) though. The reasoning conflates ``common regnal names'' with what the next Pope will actually choose.
\end{itemize}

\paragraph{3. Entity Confusion: Selecting Wrong Person/Place/Organization}

When multiple entities are mentioned, the model frequently selects the most prominent or frequently mentioned one rather than the correct answer, even when both are present in the context.

\textbf{Example 1:} ``Who will Chelsea face in the round of 16 of the 2025 FIFA Club World Cup?''

\begin{itemize}
\item \textbf{Correct answer:} Benfica
\item \textbf{Model answer:} Bayern Munich (probability: 0.4)
\item \textbf{Model's reasoning:} ``The provided articles detail the groups for the 2025 FIFA Club World Cup, with Group C consisting of Bayern Munich, Benfica, Boca Juniors, and Auckland City, and Group D including Chelsea, Flamengo, Esperance Sportive de Tunisie, and a replacement (possibly Los Angeles FC or Club America based on Article 1). The question specifies that the round of 16 matchup is between the runner-up of Group D and the winner of Group C. Without the actual group stage results, I must make an educated guess based on team strengths. Bayern Munich is a strong candidate for the winner of Group C, as they are the defending champions and one of the top teams. Chelsea, in Group D, could plausibly be the runner-up, given their history, but this is uncertain. The replacement team adds further uncertainty, as it could affect group dynamics.''
\item \textbf{Analysis:} The model selects the more prominent team (Bayern Munich) from Group C rather than correctly identifying which team would actually win the group. It uses team prominence (``defending champions,'' ``top teams'') as a proxy for group stage results, ignoring that the question requires specific match outcomes that aren't in the articles.
\end{itemize}

%% file: sections/14_prompts.tex
\clearpage
\section{Prompts}
\label{sec:prompts}

\subsection{Prompt Templates}
\label{sec:prompt_templates}

\subsubsection{Question Creation Pipeline}

\begin{tcblisting}{title={\textbf{Stage 1 --- Question Generation (Requires:  self.num\_questions\_per\_article > 1)}},colframe=black!70,colback=black!2,boxrule=0.5pt,breakable,listing only}
**Task:** Based on the provided news article, generate {self.num_questions_per_article} high-quality, DIVERSE forecasting questions which have a short answer (1 - 3 words), using the XML format specified below.
Each forecasting question should be posed in a way to predict future events. Here, the predictor will have a knowledge cutoff before the article is published and no access to the article, so a forecasting question has to be posed about information explicitly stated in the article. The question should be stated in a forward-looking manner (towards the future).
The correct answer should be a specific, short text response. The answer should be a WELL DEFINED, SPECIFIC term which the answerer can come up with on its own, without access to the news article. 

**Example Format**:
<q1>
<question_id>0</question_id>
<question_title>Who will win the Nobel Prize in Literature in 2016?</question_title>
<background>Question Start Date: 10th January 2016. The Nobel Prize in Literature is awarded annually by the Swedish Academy to authors for their outstanding contributions to literature.</background>
<resolution_criteria> 
<ul>
    <li>
      <b>Source of Truth</b>: The question will resolve when the Swedish Academy publicly announces the official 2016 Nobel Prize in Literature laureate(s)---typically via a press release on NobelPrize.org (expected on or about October 13, 2016).  
    </li>
    <li>
      <b>Resolution Date</b>: The resolution occurs on the calendar date when the 2016 laureate(s) are formally named
      (typically mid-October 2016). 
    </li>
    <li>
      <b>Accepted Answer Format</b>: The full name of the laureate exactly as given in the announcement should be provided. If more than one person shares the prize, all names must be listed in the same order as the official communiqu{\'e}.
    </li>
</ul>
</resolution_criteria>
<answer>Bob Dylan</answer>
<answer_type>String (Name)</answer_type>
</q1>

The question should follow the structured guidelines below.

### **Guidelines for Creating Short Answer Forecasting Questions**

**Title Question Guidelines**
- **Quality**: The question should be of HIGH QUALITY and hard to answer without access to the article. It should not be about any minute details in the article. THE QUESTION SHOULD BE SUCH THAT ITS ANSWER REVEALS A KEY PIECE OF INFORMATION, FROM THE ARTICLE, WHICH HAS MAXIMAL IMPACT.
- **Specific and Answerable**: The question to be created SHOULD BE FREE-FORM and have a unique, specific answer (a single word, or short phrase) without access to the article. The answer to the question should be definite, well-defined and NOT NUMERIC. IT SHOULD ALSO NOT BE UNCERTAIN like "above XYZ" OR A RANGE LIKE "between XYZ and ABC". Avoid creating binary questions (yes/no, either/or) or questions with a list of specific options (multiple choice).
- **Answerable based on article**: Each question must have a CLEAR AND DEFINITE answer based on information stated in the article. Given the question, the content of the article should be able to resolve the answer to the question INDISPUTABLY WITHOUT ANY AMBIGUITY OR UNCERTAINTY. THE ARTICLE SHOULD NOT STATE THAT THE ANSWER IS TENTATIVE OR AN ESTIMATE OR LIKELY. The answer SHOULD HAVE HAPPENED BY NOW.
- **Temporal Information**: The question should not be about recall of (past) facts or events known before the article publish date. Include any temporal information necessary to answer the question (like by which month, year, etc.) in the question. The question should always be posed in a forward-looking manner. 
- **Direct and Precise**: Titles must be straightforward and unambiguous, avoiding vague terms. Use future tense when appropriate.
- **Resolution Criteria**: ALWAYS INCLUDE A BRIEF RESOLUTION CRITERIA in the question title. This is often the date by which the question will be resolved. For example, resolution dates such as "by {{month_name}}, {{year}}?" or "in {{month_name}}, {{year}}?". THE RESOLUTION DATE SHOULD BE BASED ON (AND FAITHFUL TO) THE CONTENT OR PUBLICATION DATE OF THE ARTICLE.
- **No references to article or future information**: DO NOT refer to the specific article, such as by saying "in the article". The forecaster does not have access to the article, its metadata or any information beyond the article publish date.
- **Question Types**: Focus on "Who", "What", "When", "Where" questions that have concrete answers.
- **Understandability**: The question title should have ALL the information to be understandable by a 10 year old. It should be independently understandable without the article.
- **Tense**. ALWAYS POSE THE QUESTION IN A FORWARD-LOOKING MANNER. THE QUESTION SHOULD BE IN FUTURE TENSE. Try to use phrases like "What will", "Who will", "When will", "Where will", "How much/many will" etc. It should appear as a forecasting question and not past prediction. 

**Answer Guidelines**
- **Faithfulness to Article**: The answer should be based on information explicitly stated in the article, and not implications or your own knowledge. IT SHOULD BE STATED VERBATIM IN THE ARTICLE.
- **Non-Numeric**: The answer should not be a number or a percentage. It can be a word, phrase, date, location, etc BUT NOT MORE THAN 3 WORDS.
- **Definite** - Given the question and the article, the answer should be CLEAR, CONCRETE, CERTAIN AND DERIVABLE from the article. It should be short, WELL-DEFINED TERM and not uncertain or vague. It SHOULD NOT BE A RANGE like "between XYZ and ABC" or "above XYZ" or "below PQR".
- **Resolved** - The answer MUST be something that has already happened or is happening now. It should be resolved given today's date and not be something that will happen in the future.
- **Specificity**: The answer should be specific enough to be unambiguous. Avoid overly general answers.
- **Conciseness**: Keep answers short - typically 1-3 words, occasionally a short phrase if necessary.
- **Exactness**: For names, use the exact names mentioned (full name, if possible).
- **Uniqueness**: The answer should be unique and THE ONLY CORRECT ANSWER to the question. 
- **No Ambiguity**: The answer should be indisputable and not be open to multiple interpretations. IT SHOULD BE PRECISE AND NOT A RANGE OR UNCERTAIN ESTIMATE.

**Background Guidelines**
- **Mention Question Opening Date**: ALWAYS INCLUDE THE START DATE OF THE QUESTION IN THE BACKGROUND. IT SHOULD BE AT LEAST A FEW DAYS (OR WEEKS IF THE QUESTION IS ABOUT A LONG-TERM EVENT) BEFORE THE ARTICLE'S PUBLISH DATE AND ALSO BEFORE THE RESOLUTION DATE OF THE QUESTION. CONSEQUENTLY, THE BACKGROUND SHOULD NOT CONTAIN ANY INFORMATION WHICH HAS HAPPENED AFTER THE START DATE OF THE QUESTION.
- **Necessary Context**: The answerer does not have access to the article, so include MINIMAL CONTEXT required to understand the question keeping in mind the question opening date. Do not give (extra) details of the event from the article as background. If required, EITHER pose the event as a hypothetical scenario as if it were to happen in the future OR describe it as happening (unfolding) in real time. Describe any unfamiliar terms or concepts in the question title. 
- **SHOULD NOT HELP ANSWER**: WHILE PROVIDING THE CONTEXT, DO NOT REFER OR MENTION OR LEAK THE ACTUAL ANSWER. The background must not help answer the forecasting question. DO NOT INCLUDE ANY INFORMATION from the article or elsewhere that either directly or indirectly (even partially) reveals the answer.
- **No Additional Knowledge**: Do not add any knowledge beyond what is required to understand the question. Only include information necessary to understand the question and its context. 
- **Tense**. ALWAYS POSE THE BACKGROUND INFORMATION IN CURRENT TENSE. Only provide minimal information which is known until the question opening date.

**Resolution Criteria**
- **Necessary Criteria**: State the EXACT conditions by which the outcome will be judged. Include the criteria which determines how the question will be resolved. state the conditions by which the outcome will be judged. 
- **Date and Source of Resolution**: Always state the date and the source by which the question will be resolved. For example, resolution dates such as "by {{month_name}}, {{year}}?" or "in {{month_name}}, {{year}}?", and potential source(s) of resolution such as "based on {{news source}}", "reports from {{official name}}", etc. THE RESOLUTION DATE SHOULD BE CHOSEN THOUGHTFULLY AS THE ANSWER'S VALIDITY AND SOUNDNESS DEPENDS ON IT. THE RESOLUTION DATE SHOULD BE SUCH THAT THE ANSWER CAN BE RESOLVED DEFINITELY AND INDISPUTABLY FROM THE CONTENT OR PUBLICATION DATE OF THE ARTICLE. IT SHOULD MENTION BY WHEN IS THE OUTCOME OF THE QUESTION EXPECTED TO HAPPEN. HOWEVER, IT SHOULD NOT LEAK OR MENTION ANYTHING ABOUT THE ARTICLE. 
- **Details**: Be as detailed as possible in creating the resolution criteria for resolving the question as cleanly as possible. There should be no ambiguity in the resolution criteria.
- **Expectation and Format of Answer**: Based on the actual answer, the resolution criteria should state how precise the expected answer should be and in what format it should be. For example, if the actual answer is a date, the resolution criteria should specify how detailed the expected date should be -- only year, or both month and year, or day, month, and year all together. DO NOT GIVE THE ACTUAL DATE (ANSWER). If the actual answer is a percentage, then the criteria should state the expected answer should be a percentage. DO NOT GIVE THE ACTUAL PERCENTAGE.  If the actual answer is in certain unit, then the criteria should specify that. THE RESOLUTION CRITERIA SHOULD MAKE IT EXACTLY CLEAR AND PRECISE WHAT IS EXPECTED FROM THE ANSWERER AND IN WHAT FORMAT AND HOW IT WILL BE CHECKED LATER. IF GIVING AN EXAMPLE, IT SHOULD BE VERY GENERIC AND AS FAR AWAY FROM THE ACTUAL ANSWER AS POSSIBLE.
- **SHOULD NOT HELP ANSWER**: The resolution criteria must not directly help answer the forecasting question. DO NOT INCLUDE ANY INFORMATION from the article or elsewhere that either directly or indirectly (even partially) reveals the answer. DO NOT REFER OR MENTION OR LEAK THE ACTUAL ANSWER HERE.

**Answer Type Guidelines**
- **Expected Format**: The answer type should be either "numeric (XYZ)" if the answer is a number (of any kind) or "string (XYZ)" in all other cases. In numeric cases, XYZ should be the exact type of number expected. For example, "numeric (integer)", "numeric (decimal)", "numeric (percentage)", "numeric (whole number)", etc. In string cases, XYZ should broadly be the category of string expected. For example, "string (name)", "string (date)", "string (location)", etc. If the category is not clear, use "string (any)". HOWEVER, ALWAYS TRY TO CREATE QUESTIONS WHERE THE ANSWER CATEGORY IS CLEAR AND PRECISE.

**Question Quality Criteria**
- **Forecastable**: The question should be something that could reasonably be predicted or forecasted before the article's publication.
- **Towards the future**: THE QUESTION SHOULD BE POSED IN A FORWARD-LOOKING MANNER.
- **Interesting**: The question should be about a meaningful event or outcome, not trivial details.
- **Impactful**: The question should be such that if its answer is forecasted ahead of time, it should have significant (downstream) impact (relevant to high number of people).
- **Difficulty**: While the question should be hard to answer without access to the article, it should also not be unreasonably difficult.
- **Verifiable**: The answer should be something that can be EXACTLY verified from the article itself.
- **Time-bound**: Include clear timeframes or deadlines when relevant.
- **Free-form**: If possible, avoid creating binary questions (yes/no, either/or) or questions with a list of specific options (multiple choice).

Generate {self.num_questions_per_article} high-quality, DIVERSE short answer forecasting questions based on the provided article. Use the XML format with question_id value "0", "1", "2", etc. DO NOT INCLUDE ANY ANALYSIS, RANKING, OR ADDITIONAL COMMENTARY.

Article:
{source_article}

**Required Output Format**:
<q1>
<question_id>0</question_id>
<question_title>[Question 1]</question_title>
<background>[Background 1]</background>
<resolution_criteria>[Resolution Criteria 1]</resolution_criteria>
<answer>[Answer 1]</answer>
<answer_type>[Answer Type 1]</answer_type>
</q1>
..
<q{self.num_questions_per_article}>
<question_id>{self.num_questions_per_article - 1}</question_id>
<question_title>[Question {self.num_questions_per_article}]</question_title>
<background>[Background {self.num_questions_per_article}]</background>
<resolution_criteria>[Resolution Criteria {self.num_questions_per_article}]</resolution_criteria>
<answer>[Answer {self.num_questions_per_article}]</answer>
<answer_type>[Answer Type {self.num_questions_per_article}]</answer_type>
</q{self.num_questions_per_article}>
\end{tcblisting}

\begin{tcblisting}{title={\textbf{Stage 2 --- Individual Validation}},colframe=black!70,colback=black!2,boxrule=0.5pt,breakable,listing only}
**Task:** You will be provided with a news article and a question WHOSE ANSWER IS SUPPOSED TO BE BASED ON THE ARTICLE. Your job is to validate whether the answer to the question is valid by being faithful to the article (content, title, or description).

GO THROUGH EACH SEGMENT OF THE QUESTION ONE BY ONE (TITLE, BACKGROUND, RESOLUTION CRITERIA, ANSWER) TO UNDERSTAND THE WHOLE QUESTION. THEN CHECK EACH OF THE FOLLOWING CRITERIA: 

1. **Tense and Details**: FIRST CHECK WHETHER THE QUESTION IS NOT UNDER SPECIFIED OR STATED IN PAST TENSE. IT IS FINE IF THE QUESTION IS STATED IN CURRENT OR FUTURE TENSE. 
2. **Definite resolution of the answer by the article**: CHECK WHETHER THE ANSWER TO THE QUESTION IS SOUND, CLEAR AND PRESENT IN OR CAN BE DERIVED FROM THE ARTICLE. THE ARTICLE SHOULD RESOLVE THE ANSWER DEFINITELY AND IN AN INDISPUTABLE MANNER (WITHOUT ANY AMBIGUITY). THIS IS THE MOST IMPORTANT CRITERIA.
3. **Well-defined Answer**: The answer to the question should be short (NOT MORE THAN 3 WORDS). IT SHOULD NOT BE A PHRASE AND SHOULD BE SOMETHING WHICH IS CONCRETE, SPECIFIC AND WELL-DEFINED.
4. **Non-Numeric**: THE *ANSWER TYPE* SHOULD NOT BE NUMERIC LIKE A PERCENTAGE, INTEGER, DECIMAL, OR A RANGE.
5. **Single Correct Answer**: ANALYZE WHETHER THE QUESTION CAN HAVE MULTIPLE OUTCOMES OR RIGHT ANSWERS. IF SO, THE QUESTION FAILS THIS CRITERIA. OTHERWISE, ENSURE THAT THE PROVIDED ANSWER IS THE SOLE CORRECT ANSWER TO THE QUESTION. IT SHOULD NOT BE THE CASE THAT THE QUESTION CAN HAVE MULTIPLE (DISTINCT) CORRECT ANSWERS.

If ALL the above criteria pass (question is stated as required, answer to the whole question is valid, well-defined, and it is the only correct answer to the question), ONLY THEN return <answer>1</answer>. Otherwise, return <answer>0</answer>. ALWAYS END YOUR RESPONSE IN <answer> </answer> tags.

**Article:**
{source_article}

**Question:**
{questions_text}

**Output Format:**
<answer>0/1</answer>
\end{tcblisting}

\begin{tcblisting}{title={\textbf{Stage 3 --- Choose Best}},colframe=black!70,colback=black!2,boxrule=0.5pt,breakable,listing only}
**Task:** You will be provided with a list of questions (possibly with size 1). Your job is to choose the best question from the list based on the following criteria or end your response with "NO GOOD QUESTION" if none of the questions meet the criteria.

**Instructions:**
GO THROUGH EACH QUESTION ONE BY ONE AND ANALYZE IT FOR THE FOLLOWING:
1. **Valid for forecasting**: Check if the WHOLE QUESTION is stated in a forward-looking manner. FROM THE PERSPECTIVE OF THE START DATE TO THE RESOLUTION DATE MENTIONED IN THE QUESTION, CHECK IF IT IS A VALID FORECASTING QUESTION. IF THE TIME HORIZON (START DATE TO RESOLUTION DATE) IN THE QUESTION IS AT LEAST A SINGLE DAY, THEN THE QUESTION SHOULD BE CONSIDERED VALID FOR FORECASTING. Go through each segment of the question (question title, background, resolution criteria) and check if each of them is valid and forward-looking.
2. **Tense**: The question SHOULD NOT BE STATED IN PAST TENSE. If the question covers an event, it should not imply as if the outcome of the event has already happened or occurred.
3. **Single Correct Answer**: ANALYZE WHETHER THE QUESTION CAN HAVE MULTIPLE OUTCOMES OR RIGHT ANSWERS. IF SO, THE QUESTION FAILS THIS CRITERIA. OTHERWISE, ENSURE THAT THE PROVIDED ANSWER IS THE SOLE CORRECT ANSWER TO THE QUESTION. IT SHOULD NOT BE THE CASE THAT THE QUESTION CAN HAVE MULTIPLE (DISTINCT) CORRECT ANSWERS.
4. **Impact**: How many people will the outcome of the question be relevant or interesting to? Consider on the basis of significant downstream impact or enabling meaningful action.
5. **Not Binary/Multiple Choice**: Question SHOULD NOT BE BINARY (yes/no, either ABC or XYZ, etc.) OR MULTIPLE CHOICE (SELECT FROM A LIST OF OPTIONS). It should be free-form (string -- name, date, place, etc.) or numerical (number, percentage, etc.). 
6. **Understandable**: The question as a whole (title, background, resolution criteria) should have sufficient details to understand the premise of the question. Every detail should be crystal clear and the question should not be under or over specified. 
7. **Definite Answer**: EXTRACT THE ACTUAL ANSWER TO THE QUESTION PROVIDED IN ITS <answer> </answer> TAG. The extracted answer should be short, definite, well-defined and not uncertain or vague. It SHOULD NOT BE A PHRASE OR A RANGE like "between XYZ and ABC" or "above XYZ" or "below PQR".

ANALYZE EACH QUESTION BASED ON THE ABOVE CRITERIA ONE BY ONE AND CHOOSE THE ONE WHICH PASSES ALL THE ABOVE CRITERIA. IF MULTIPLE QUESTIONS SATISFY THE CRITERIA, CHOOSE THE ONE WHICH WILL HAVE THE HIGHEST IMPACT (AFFECTS OR IS RELEVANT TO THE MOST NUMBER OF PEOPLE). IF NO QUESTION MEETS THE CRITERIA, RETURN "NO GOOD QUESTION FOUND". OTHERWISE, RETURN THE BEST QUESTION IN THE SAME FORMAT AS THE INPUT. 

**Generated Questions:**
{questions_text}

**Output Format:**
<q1>
<question_id>0</question_id>
<question_title>[ORIGINAL Title of the best question]</question_title>
<background>[ORIGINAL Background of the best question]</background>
<resolution_criteria>
<ul>
    <li> <b>Source of Truth</b>: [ORIGINAL Source of Truth of the best question] </li>
    <li> <b>Resolution Date</b>: [ORIGINAL Date of the best question] </li>
    <li> <b>Accepted Answer Format</b>: [ORIGINAL Accepted Answer Format of the best question] </li>
</ul>
</resolution_criteria>
<answer>[ORIGINAL Answer of the best question]</answer>
<answer_type>[ORIGINAL Answer Type of the best question]</answer_type>
</q1>
\end{tcblisting}

\begin{tcblisting}{title={\textbf{Stage 4 --- Leakage Removal}},colframe=black!70,colback=black!2,boxrule=0.5pt,breakable,listing only}
**Task:** You will be provided with a forecasting question. Your job is to ANALYZE whether the question's answer has obviously leaked in the content of the question. The question will have multiple segments -- question title, background, resolution criteria. EXCEPT THE QUESTION TITLE, GO THROUGH EACH SEGMENT STEP BY STEP and check if any part DIRECTLY leaks the actual answer. If leakage is found, ONLY THEN rephrase the problematic parts appropriately to remove the answer while maintaining the question's integrity and focus. DO NOT CHANGE ANY PART OF THE QUESTION UNNECESSARILY. 

USE THE SAME XML FORMAT IN YOUR RESPONSE AS IS IN THE INPUT.

**Generated Question:**
{questions_text}

**Instructions:**
1. **Keep the title unchanged**: DO NOT MAKE ANY CHANGE TO THE QUESTION TITLE.
2. **Keep the start date in the background unchanged**: DO NOT MAKE ANY CHANGE TO THE QUESTION'S START DATE IN THE BACKGROUND.
3. **Identify the answer**: First, extract the actual answer from the XML tags for the current question being processed.
4. **Identify Leakage**: Keeping the extracted answer in mind, check if the  background, or resolution criteria (each of them -- source of truth, resolution date, accepted answer format) contain information that reveals the answer.
5. **Types of leakage which can be ignored**: The following types of leakage are fine and don't need to be rephrased:
   - If the outcome (actual answer) of the question is binary (yes/no, either ABC or XYZ, etc.), then NO NEED TO CHANGE ANYTHING ANYWHERE.
   - If the resolution criteria is based on a list of specific options, then NO NEED TO CHANGE ANYTHING IN ANY SEGMENT (BACKGROUND, RESOLUTION CRITERIA, etc.). For example, if the accepted answer format states "answer must be either .." OR "answer must be one of the following terms..", then NO NEED TO CHANGE ANYTHING ANYWHERE.
6. **Types of Leakage to Check:** ONLY CONSIDER THE FOLLOWING KIND OF LEAKAGE:
   - DIRECT MENTIONS of the answer (either in word or number form) or part of the answer in the question/background/resolution
   - References to specific outcomes that ARE CLOSE TO (OR REVEAL)THE ACTUAL ANSWER
7. **Rephrase Strategy**: If leakage is found, rephrase the problematic part while:
   - Keeping the question's core intent
   - Maintaining forecasting nature
   - Preserving necessary context
   - Making the answer UNOBVIOUS by replacing with a FAKE ANSWER (FAKE NAME, DATE, NUMBER, PERCENTAGE, etc.) WHICH IS GENERIC AND NOT CLOSE TO THE ACTUAL ANSWER.
   - The rephrased part should not contain any information that is part of the actual answer. Neither should it indirectly hint or reveal the answer. 
8. **Check Accepted Answer Format**: IF THERE IS ANY EXAMPLE MENTIONED IN ACCEPTED ANSWER FORMAT ("e.g..."), MAKE SURE THE EXAMPLE IS GENERIC AND AS FAR AWAY FROM THE ACTUAL ANSWER AS POSSIBLE. DO NOT INCLUDE AN EXAMPLE IF NOT MENTIONED ALREADY. 
9. **Do not change the answer**: Do not change the actual answer to the question. 
10. **Do not change the answer_type**: DO NOT MAKE ANY CHANGE TO the answer_type.
11. **Each segment should be checked independently**: Go through each segment of the whole question one by one. Everything from the title of the question to the background information to the resolution criteria should be checked independently with reference to the answer of the question. In the resolution criteria, go through each <li> step by step. Do not change the other segments when rephrasing a problematic segment.
12. **Do not change anything unless leakage is found**: DO NOT UNNECESSARILY CHANGE ANY PART OF THE QUESTION UNLESS LEAKAGE IS FOUND.

IT IS ALSO POSSIBLE THAT MULTIPLE PARTS OF THE QUESTION HAVE LEAKAGE. YOU SHOULD CHECK EACH OF THEM INDEPENDENTLY AND ONLY IF LEAKAGE IS FOUND, REPHRASE THE PROBLEMATIC PARTS. DO NOT OVER-ANALYZE.

During your analysis, you should:
- Go through EACH SEGMENT OF THE QUESTION STEP BY STEP INDEPENDENTLY. First <background> and then inside <resolution_criteria>. Under the resolution criteria, go through the source of truth, resolution date, accepted answer format (each of them is a <li> tag) one by one. For each such segment, do the following:
    - Compare the content in the current segment with the actual answer. If ANY PART OF THE ANSWER is mentioned in the current segment, then consider that as a leakage UNLESS THE ACCEPTED ANSWER FORMAT IS BINARY (yes/no, either ABC or XYZ, etc.) OR A LIST OF SPECIFIC OPTIONS.
    - IF THE CURRENT SEGMENT IS BACKGROUND, DO NOT CHANGE THE QUESTION START DATE.
    - If the current segment is accepted answer format and there is a SPECIFIC EXAMPLE MENTIONED in it ("e.g. XYZ") which is close to the actual answer, then consider that as a leakage.
    - If leakage is found in the current segment, mention "Leakage found -- {{reason for leakage}}". Form the segment with the problematic parts rephrased and mention it as "Replacement -- {{rephrased_text}}." THE REPHRASED TEXT SHOULD BE AS FAR AWAY FROM THE ACTUAL ANSWER AS POSSIBLE. It should now be present in the final output (instead of the original text).
    - Otherwise, mention "No leakage found". In your final output after you finish the analysis, return this segment UNCHANGED.
    - These outputs should be in the same format as the original input. 
- Return the actual answer unchanged in the <answer> tag in your final output.
- Skip any other segments (question title, answer_type, etc.) in your analysis and output them unchanged (verbatim) in the final output.

Output your analysis step by step, and then end your response with the CORRECTED question in THE SAME XML FORMAT AS THE ORIGINAL. 

**Output Format**:
{{ analysis }}

<q1>
<question_id>0</question_id>
<question_title>[UNCHANGED Question Title]</question_title>
<background>[Corrected Background]</background>
<resolution_criteria>
<ul>
    <li> [UNCHANGED Question Start Date] [Corrected Source of Truth] </li>
    <li> [UNCHANGED Resolution Date] </li>
    <li> [Corrected Accepted Answer Format] </li>
</ul>
</resolution_criteria>
<answer>[UNCHANGED Answer]</answer>
<answer_type>[UNCHANGED Answer Type]</answer_type>
</q1>
\end{tcblisting}

\subsubsection{Evaluation Prompts}

\begin{tcblisting}{title={\textbf{Model Evaluation Prompt (Uses retrieved news summaries)}},colframe=black!70,colback=black!2,boxrule=0.5pt,breakable,listing only}
You will be asked a forecasting question (which might be from the past). You have to come up with the best guess for the final answer.
You will also be provided with a list of retrieved news articles summaries which you may refer to when coming up with your answer.

Please provide your reasoning before stating your final answer and also express how likely you think your answer is to be correct (your confidence in your answer).

Question Title: {question_title}

Question Background:{question_background}

Resolution Criteria: {resolution_criteria}

Expected Answer Type: {expected_answer_type}

Relevant passages from retrieved news articles:
{retrieved_news_articles_text}

Think step by step about the information provided, reason about uncertainty and put your final answer (in the format asked) in <answer> </answer> tags.
You should also specify your confidence in your answer in <probability> </probability> tags.
The probability should be a number between 0 and 1.

You will be rewarded based on the probability (p) you assign to your answer.
Your answer will be evaluated using the BRIER SCORING RULE which is basically (- (1 - p)\^{2}) if your answer is correct and (- 1 - p\^{2}) if your answer is incorrect.

For example, if p = 0.5, and your answer is incorrect, then your score will be (-1 - 0.5\^{}2) = (-1 - 0.25) = -1.25
whereas if the answer was correct, then your score would be (- (1 - 0.5)\^{2}) = (- (0.5)\^{2}) = -0.25.

Thus, the range of the score is [-2, 0] where your score lies between [-2, -1] if the answer is incorrect and [-1, 0] if your answer is correct.

If your answer is correct, you will be REWARDED more if your probability is higher whereas if your answer is incorrect, you will be PENALIZED more if your probability is higher.
YOU HAVE TO MAXIMIZE YOUR SCORE.

Your final answer should be concise (NOT MORE THAN A FEW WORDS LONG) and your response SHOULD STRICTLY END with <answer> </answer> tags and <probability> </probability> tags.
\end{tcblisting}

\begin{tcblisting}{title={\textbf{Example Prompt from Test Set (Without Retrieval)}},colframe=black!70,colback=black!2,boxrule=0.5pt,breakable,listing only}

You will be asked a forecasting question (which might be from the past).
You have to come up with the best guess for the final answer.
Please provide your reasoning before stating your final answer and also express how likely you think your answer is to be correct (your confidence in your answer).

Question Title:
Which country in the Americas will report the highest number of chikungunya cases by July 15, 2025?

Question Background:
Public health agencies in the Americas are compiling chikungunya case counts for individual countries as the outbreak spreads in the region.

Resolution Criteria:
<ul>
  <li><b>Source of Truth</b>: Cumulative case figures published by the Pan American Health Organization or the European Centre for Disease Prevention and Control.</li>
  <li><b>Resolution Date</b>: July 15, 2025, when the mid-July regional report is issued.</li>
  <li><b>Accepted Answer Format</b>: The name of the country in the Americas with the highest total reported chikungunya cases.</li>
</ul>

Expected Answer Type:
string (location)

Think step by step about the information provided, reason about uncertainty and put your final answer (in the format asked) in <answer> </answer> tags.
You should also specify your confidence in your answer in <probability> </probability> tags.
The probability should be a number between 0 and 1.

You will be rewarded based on the probability (p) you assign to your answer.
Your answer will be evaluated using the BRIER SCORING RULE which is basically (- (1 - p)^2) if your answer is correct and (- 1 - p^2) if your answer is incorrect.

For example, if p = 0.5, and your answer is incorrect, then your score will be (-1 - 0.5^2) = (-1 - 0.25) = -1.25
whereas if the answer was correct, then your score would be (- (1 - 0.5)^2) = (- (0.5)^2) = -0.25.

Thus, the range of the score is [-2, 0] where your score lies between [-2, -1] if the answer is incorrect and [-1, 0] if the answer is correct.

If your answer is correct, you will be REWARDED more if your probability is higher whereas if your answer is incorrect, you will be PENALIZED more if your probability is higher.
YOU HAVE TO MAXIMIZE YOUR SCORE.

Your final answer should be concise (NOT MORE THAN A FEW WORDS LONG) and your response SHOULD STRICTLY END with <answer> </answer> tags and <probability> </probability> tags.

\end{tcblisting}

\begin{tcblisting}{title={\textbf{Example Prompt from Test Set (With Retrieval)}},colframe=black!70,colback=black!2,boxrule=0.5pt,breakable,listing only}

You will be asked a forecasting question (which might be from the past).
You have to come up with the best guess for the final answer.
You will also be provided with a list of retrieved news articles summaries which you may refer to when coming up with your answer.
Please provide your reasoning before stating your final answer and also express how likely you think your answer is to be correct (your confidence in your answer).

Question Title:
Which country in the Americas will report the highest number of chikungunya cases by July 15, 2025?

Question Background:
Public health agencies in the Americas are compiling chikungunya case counts for individual countries as the outbreak spreads in the region.

Resolution Criteria:
<ul>
  <li><b>Source of Truth</b>: Cumulative case figures published by the Pan American Health Organization or the European Centre for Disease Prevention and Control.</li>
  <li><b>Resolution Date</b>: July 15, 2025, when the mid-July regional report is issued.</li>
  <li><b>Accepted Answer Format</b>: The name of the country in the Americas with the highest total reported chikungunya cases.</li>
</ul>

Expected Answer Type:
string (location)

Relevant passages from retrieved news articles:

Article 1:
Title: CDC warns US travellers of growing Dengue threat. Here's what you need to know
Source: www.hindustantimes.com
Article Publish Date: March 21, 2025
Relevant Passage: CDC warns of rising dengue fever cases among U.S. travellers, reporting 3,484 cases in 2024, an 84% increase from last year. CDC cited a "record number" of cases reported among travellers in 2024, totalling 3,484 cases. This marked an 84 percent surge compared to the previous year. "This trend is expected to continue with increased dengue activity in endemic areas in 2025," the CDC stated in its warning. Dengue remains prevalent in certain regions of the States. The virus is being actively transmitted in U.S. territories, including Puerto Rico and the U.S. Virgin Islands, where outbreaks have been declared. Warmer temperatures during the spring and summer months create favourable conditions for the spread of the disease. Over the past five years, dengue cases have surged worldwide, with the Americas being notably affected. Data from the World Health Organization (WHO) reveals that in 2024 alone, there were 7.6 million reported cases. Among these, 3.4 million were confirmed, over 16,000 were classified as severe, and more than 3,000 resulted in fatalities. Puerto Rico has been grappling with a sustained dengue outbreak since early 2024. The island surpassed the outbreak threshold in February of that year, leading to the declaration of a public health emergency in March 2024, which remains in effect. Puerto Rico recorded 6,291 dengue cases in 2024, with more than half of the patients requiring hospitalization. Thirteen individuals lost their lives to the virus, according to the CDC data. Similarly, the U.S. Virgin Islands declared an outbreak in August 2024, which also remains ongoing. Health authorities reported 208 cases in 2024 and an additional 30 cases in early 2025. The highest numbers of travel-related dengue infections in 2024 were reported in Florida, California, and New York.

Article 2:
Title: Vaccine Against Chikungunya Approved By The FDA. Should You Get It?
Source: www.forbes.com
Article Publish Date: November 20, 2023
Relevant Passage: efficient vector was Aedes aegypti; however, an interesting phenomenon occurred in 2005. There was a slight alteration of the virus genome, which allowed it to spread more efficiently with a more common mosquito, Aedes albopictus. That facilitated a massive pandemic in 2005 on La Reunion island and neighboring areas around the Indian Ocean. In 2013, chikungunya arrived in the Americas for the first time and it subsequently tore through the Caribbean islands. Is The United States At Risk? Yes. Before 2006, chikungunya rarely occurred in U.S. travelers. Then between 2006-2013, we had about 28 cases per year, but those cases were infected outside the U.S. The situation changed after the Caribbean outbreak and chikungunya arrived at our shores in 2014, with affected areas in Florida, the U.S. Virgin Islands and more severely in Puerto Rico (over 30,000 suspected cases). The continental U.S. dodged a bullet, though, with fewer cases than feared. It probably helps that we have widespread air conditioning and window screens in our southern states, which reduces contact with mosquitoes. We remain vulnerable, though, since we still have the mosquito vectors, primarily in the central and southeast parts of the United States. How Can I Reduce My Risk? Minimizing risk focuses on avoiding mosquito bites when living in or visiting an area with active spread of chikungunya, including staying indoors in screened areas during the daytime, using bed nets, using insect repellant and wearing long, loose-fitting clothing. By avoiding Aedes mosquitoes, you reduce your risk of infection from chikungunya as well as dengue and zika viruses. Do I Need The Vaccine?

Article 3:
Title: First Chikungunya Vaccine Now FDA Approved - What To Know About The `Emerging Global Health Threat'
Source: www.forbes.com
Article Publish Date: November 10, 2023
Relevant Passage: severe chikungunya-like adverse reactions following administration of Ixchiq." Big Number 5 million. There have been at least that many chikungunya cases reported globally over the past 15 years, the FDA said. The agency described the virus as an ``emerging global health threat" that ``has spread to new geographical areas causing a rise in global prevalence of the disease." What We Don't Know Health officials and agencies like the World Health Organization warn official infection counts are likely to significantly underestimate the true prevalence of chikungunya. Accurate diagnosis, disease surveillance and reporting can be tricky in some parts of the world on account of funding and capacity within healthcare systems and chikungunya is also ``easy to misdiagnose" on account of causing similar symptoms to other mosquito-borne illnesses like dengue and Zika. What To Watch For Valneva said the vaccine will initially address the ``potential needs" of some 60 million Americans who it says travel to countries where mosquito-borne diseases are endemic each year. This fits in well with its other shots for cholera and Japanese encephalitis aimed at travelers, the company said. Valneva said it will work towards commercializing the shot in the U.S. early next year and work towards securing a vote of approval endorsing the shot from the Centers for Disease Control and Prevention's vaccine advisory committee at the end of February. The FDA is an influential regulator and its go-ahead will likely speed Ixchiq's passage through other regulatory processes globally, particularly in areas where chikungunya is a more pressing concern. Key Background Chikungunya is regularly identified as an emerging threat to global health on account of the debilitating and prolonged disease it can cause. Chikungunya was first identified in Tanzania in 1952 and sporadic outbreaks were later recorded in parts of Africa and Asia. The virus has since spread globally and has been identified on all continents except Antarctica. The economic and social impact of the disease can be devastating -- the Coalition for Epidemic Preparedness Innovations estimates the cost to the Americas alone to be around $185 billion -- and the warming climate, a boon to the mosquitoes that spread the disease, is likely to widen areas at risk. While the virus is more commonly reported among travelers in the U.S. and parts of Europe, local transmission has been documented, suggesting future outbreaks may be possible or that the virus could gain a permanent foothold. Further Reading First Vaccine For Chikungunya

Article 4:
Title: Dengue fever cases rising in popular spring break locations, CDC alerts
Source: www.foxnews.com
Article Publish Date: March 24, 2025
Relevant Passage: is common in the Americas, Africa, the Middle East, Asia and the Pacific Islands, among other countries, according to the CDC. TRAVEL HOT SPOT SEEKS EMERGENCY DECLARATION OVER MASSIVE BUG INFESTATION In 2024, more than 13 million cases were reported in North, Central and South America, as well as in the Caribbean. Local transmission of these outbreaks was reported in California, Texas and Florida last year. Typical symptoms include aches and pains (in the eyes, muscles, joints, or bones), nausea, vomiting and rash  --  usually experienced within two weeks of being bitten. Most people experience symptoms for two to seven days before recovering. CLICK HERE TO SIGN UP FOR OUR HEALTH NEWSLETTER "It's typically a more mild illness, but can be severe, causing headaches, joint pain, fever, abdominal pain and even death," Dr. Mark Fischer, regional medical director of International SOS, a leading medical and security services company, previously told Fox News Digital. There is not currently any medication to treat dengue, according to the CDC. For more Health articles, visit www.foxnews.com/health Infected people are advised to rest, take acetaminophen for pain and fever, stay hydrated and see a doctor. There is a vaccine available for U.S. children between 9 and 16 years of age who have previously tested positive for dengue and are living in areas where the infection is common. CLICK HERE TO GET THE FOX NEWS APP

Article 5:
Title: Latin America: Key Themes To Watch In 2025
Source: seekingalpha.com
Article Publish Date: January 16, 2025
Relevant Passage: Latin America: Key Themes To Watch In 2025 Latin America's aggregate growth will slightly accelerate in 2025, but this overshadows slower growth across most countries. Read more here. Markit 3.25K Follower s ( 6min ) Summary Latin America's aggregate growth will slightly accelerate in 2025, but this overshadows slower growth across most countries. The potential application of tariffs by the incoming US administration would negatively impact trade and weaken many of the region's currencies, while forced repatriation of illegal workers in the US implies a reduction of remittance flows. We expect lower price pressures based on Market Intelligence's global assumption that prices for agriculture-related commodities and oil prices will fall in 2025. In response to US President Donald Trump's proposals on tariffs and deportations, Mexico and most countries in Central America are likely to align with the requests from the new US administration. Here is how we see our key themes for 2025 shaping Latin America's operational and investment environment. Economic angst Latin America's aggregate growth will slightly accelerate in 2025, but this overshadows slower growth across most countries. We project only five This article was written by 3.25K Follower s IHS Markit (Nasdaq: INFO) is a world leader in critical information, analytics and solutions for the major industries and markets that drive economies worldwide. The company delivers next-generation information, analytics and solutions to customers in business, finance and government, improving their operational efficiency and providing deep insights that lead to well-informed, confident decisions. IHS Markit has more than 50,000 key business and government customers, including 80 percent of the Fortune Global 500 and the world's leading financial institutions. Headquartered in London, IHS Markit is committed to sustainable, profitable growth. Comments Recommended For You Related Stocks SymbolLast Price% ChgEWZ--iShares MSCI Brazil ETFILF--iShares Latin America 40 ETFBRF--VanEck Brazil Small-Cap ETFFLN--First Trust Latin America AlphaDEX Fund ETFFBZ--First Trust Brazil AlphaDEX Fund ETF Related Analysis Trending Analysis Trending News

Think step by step about the information provided, reason about uncertainty and put your final answer (in the format asked) in <answer> </answer> tags.
You should also specify your confidence in your answer in <probability> </probability> tags.
The probability should be a number between 0 and 1.

You will be rewarded based on the probability (p) you assign to your answer.
Your answer will be evaluated using the BRIER SCORING RULE which is basically (- (1 - p)^2) if your answer is correct and (- 1 - p^2) if your answer is incorrect.

For example, if p = 0.5, and your answer is incorrect, then your score will be (-1 - 0.5^2) = (-1 - 0.25) = -1.25
whereas if the answer was correct, then your score would be (- (1 - p)^2) = (- (0.5)^2) = -0.25.

Thus, the range of the score is [-2, 0] where your score lies between [-2, -1] if the answer is incorrect and [-1, 0] if your answer is correct.

If your answer is correct, you will be REWARDED more if your probability is higher whereas if your answer is incorrect, you will be PENALIZED more if your probability is higher.
YOU HAVE TO MAXIMIZE YOUR SCORE.

Your final answer should be concise (NOT MORE THAN A FEW WORDS LONG) and your response SHOULD STRICTLY END with <answer> </answer> tags and <probability> </probability> tags.

\end{tcblisting}